\documentclass[11pt,english,authoryear]{article}
\usepackage[a4paper]{geometry}
\usepackage[export]{adjustbox}
\usepackage[latin9]{inputenc}
\usepackage[T1]{fontenc}
\usepackage{microtype}
\usepackage{mathptmx}
\usepackage{stmaryrd}
\usepackage{stackrel}
\usepackage{booktabs}
\usepackage{makecell}
\usepackage{graphicx}
\usepackage{setspace}
\usepackage{amsmath}
\usepackage{amssymb}
\usepackage{accents}
\usepackage{dcolumn}
\usepackage{diagbox}
\usepackage{amsthm}
\usepackage{subfig}
\usepackage{float}
\usepackage{babel}
\usepackage{tikz}
\usepackage{natbib}

\makeatletter
\def\ps@pprintTitle{
  \let\@oddhead\@empty
  \let\@evenhead\@empty
  \let\@oddfoot\@empty
  \let\@evenfoot\@oddfoot
}
\makeatother

\theoremstyle{definition}
\newtheorem{defn}{Definition}\theoremstyle{plain}
\newtheorem{lem}{Lemma}\theoremstyle{plain}
\newtheorem{thm}{Theorem}\theoremstyle{plain}
\newtheorem{cor}{Corollary}\theoremstyle{plain}
\newtheorem{prop}{Proposition}\theoremstyle{plain}

\floatstyle{ruled}
\newfloat{algorithm}{tbp}{loa}
\providecommand{\algorithmname}{Algorithm}
\floatname{algorithm}{\protect\algorithmname}

\DeclareMathOperator*{\argmin}{argmin}

\newcolumntype{.}{D{.}{.}{-1}}

\newcommand\thickbar[1]{\accentset{\rule{.4em}{.8pt}}{#1}}

\begin{document}

\title
{Finite-sample and asymptotic analysis of generalization ability with an application to penalized regression\footnote{The authors would like to thank Mike Bain, Colin Cameron, Peter Hall and Tsui Shengshang for valuable comments on an earlier draft. We would also like to acknowledge participants at the 12th International Symposium on Econometric Theory and Applications and the 26th New Zealand Econometric Study Group as well as seminar participants at Utah, UNSW, and University of Melbourne for useful questions and comments. Fisher would like to acknowledge the financial support of the Australian Research Council, grant DP0663477.}}

\author
{Ning Xu \\ School of Economics, University of Sydney \\  n.xu@sydney.edu.au
\and
Jian Hong \\ School of Economics, University of Sydney \\  jian.hong@sydney.edu.au
\and
Timothy C.G. Fisher \\ School of Economics, University of Sydney \\  tim.fisher@sydney.edu.au
}

\maketitle

\begin{abstract}

In this paper, we study the performance of extremum estimators from the perspective of generalization ability (GA): the ability of a model to predict outcomes in new samples from the same population. By adapting the classical concentration inequalities, we derive upper bounds on the empirical out-of-sample prediction errors as a function of the in-sample errors, in-sample data size, heaviness in the tails of the error distribution, and model complexity. We show that the error bounds may be used for tuning key estimation hyper-parameters, such as the number of folds $K$ in cross-validation. We also show how $K$ affects the bias-variance trade-off for cross-validation. We demonstrate that the $\mathcal{L}_2$-norm difference between penalized and the corresponding un-penalized regression estimates is directly explained by the GA of the estimates and the GA of empirical moment conditions. Lastly, we prove that all penalized regression estimates are $L_2$-consistent for both the $n \geqslant p$ and the $n < p$ cases. Simulations are used to demonstrate key results.


\end{abstract}

\newpage
\renewcommand{\baselinestretch}{1.2}
\setcounter{page}{1}


\section{Introduction}

Traditionally in econometrics, an estimation method is implemented on sample data in order to infer patterns in a population. Put another way, inference centers on generalizing to the population the pattern learned from the sample and evaluating how well the sample pattern fits the population. An alternative perspective is to consider how well a sample pattern fits another sample. In this paper, we study the ability of a model estimated from a given sample to fit new samples from the same population, referred to as the \textbf{generalization ability} (GA) of the model. As a way of evaluating the external validity of sample estimates, the concept of GA has been implemented in recent empirical research. For example, in the policy evaluation literature \citep{ belloni2013program, gechter2015generalizing, dolton2006econometric, blundell2004evaluating}, the central question is whether any treatment effect estimated from a pilot program can be generalized to out-of-sample individuals. Similarly, for economic forecasting, \citet{stock2012generalized} used GA as a criterion to pick optimal weight coefficients for model averaging predictors. Generally speaking, a model with higher GA will be more appealing for policy analysis or prediction.

With a new sample at hand, GA is easily measured using validation or cross-validation to measure the goodness of fit of an estimated model on out-of-sample data. Without a new sample, however, it can be difficult to measure GA \emph{ex ante}. In this paper, we demonstrate how to quantify the GA of an in-sample estimate when only a single sample is available by deriving upper bounds on the empirical out-of-sample errors. The upper bounds on the out-of-sample errors depend on the sample size, an index of the complexity of the model, a loss function, and the distribution of the underlying population. As it turns out, the bounds serve not only as a measurement of GA, but also illustrate the trade-off between in-sample fit and out-of-sample fit. By modifying and adapting the bounds, we are also able to analyze the performance of $K$-fold cross-validation and penalized regression. Thus, the GA approach yields insight into the finite-sample and asymptotic properties of penalized regression as well as cross-validation.

As well as being an out-of-sample performance indicator, GA may also be used for model selection. Arguably, model selection is coming to the forefront in empirical work given the increasing prevalence of high-dimensional data in economics and finance. We often desire a smaller set of predictors in order to gain insight into the most relevant relationships between outcomes and covariates. Model selection based on GA not only offers improved interpretability of an estimated model, but, critically, it also improves the bias-variance trade-off relative to the traditional extremum estimation approach.

\subsection{Traditional approach to the bias-variance trade-off}

Without explicitly introducing the concept of GA, the classical econometrics approach to model selection focusses on the bias-variance trade-off, yielding methods such as the information criteria (IC), cross-validation, and penalized regression. For example, an IC may be applied to linear regression
\[
	Y = X \beta + u
\]
where $Y\in\mathbb{R}^{n}$ is a vector of outcome variables, $X\in\mathbb{R}^{n\times p}$
is a matrix of covariates and $u\in\mathbb{R}^{n}$ is a vector of i.i.d.\ random errors. The parameter vector $\beta\in\mathbb{R}^{p}$ may be sparse in the sense that many of its elements are zero. Model selection typically involves using a score or penalty function that depends on the data \citep{heckerman95}, such as the Akaike information criterion \citep{akaike73}, Bayesian information criterion \citep{schwarz78}, cross-validation errors \citep{stone74,stone77} or the mutual information score among variables \citep{friedman97,friedman00}.

An alternative approach to model selection is penalized regression, implemented through the objective function:
\begin{equation}
	\underset{b_{\lambda}}{\min}~~
	\frac{1}{n}\left(\left\Vert Y-Xb_{\lambda}\right\Vert _{2}	\right)^{2}
	+\lambda\Vert b_{\lambda}\Vert_{\gamma}
\label{penreg}
\end{equation}
where $\Vert\cdot\Vert_{\gamma}$ is the $\mathcal{L}_{\gamma}$ norm and $\lambda\geqslant0$ is a penalty parameter. One way to derive the penalized regression estimates $b_{\lambda}$ is through validation, summarized in Algorithm~1.

\smallskip
\begin{table}[h]
\caption*{Algorithm 1: Penalized regression estimation under validation}

\begin{tabular}{lp{137mm}}
\toprule
    1. & Set $\lambda = 0$. \\
    2. & Partition the sample into a training set $T$ and a test set $S$. Standardize all variables (to ensure the penalized regression residual $e$ satisfies $\mathbb{E}(e)=0$ in $T$ and $S$). \\
    3. & Compute the penalized regression estimate $b_{\lambda}$ on $T$. Use $b_{\lambda}$ to calculate the prediction error on $S$. \\
    4. & Increase $\lambda$ by a preset step size. Repeat 2 and 3 until	$b_{\lambda} = \mathbf{0}$. \\
    5. & Select $b_{pen}$ to be the $b_{\lambda}$ that minimizes the prediction error on $S$. \\
\bottomrule
\end{tabular}
\end{table}

\noindent
As shown in Algorithm~1, validation works by solving the constrained minimization problem in eq.~(\ref{penreg}) for each value of the penalty parameter $\lambda$ to derive a $b_{\lambda}$. When the feasible range of $\lambda$ is exhausted, the estimate that produces the smallest out-of-sample error among all the estimated $\{b_{\lambda}\}$ is chosen as the penalized regression estimate, $b_{pen}$.

Note in eq.~(\ref{penreg}) that if $\lambda=0$, the usual OLS estimator is obtained. The IC can be viewed as special cases with $\lambda=1$ and $\gamma=0$. The lasso \citep{tibshirani96} corresponds to the case with $\gamma=1$ (an $\mathcal{L}_{1}$ penalty). When $\gamma=2$ (an $\mathcal{L}_{2}$ penalty), we have the familiar ridge estimator \citep{hoerlkennard70}. For any $\gamma>1$, we have the bridge estimator  \citep{frankfriedman93}, proposed as a generalization of the ridge.

A range of consistency properties have been established for the IC and penalized regression. \cite{shao97} proves that various IC and cross-validation are consistent in model selection. \cite{breiman95, chickering04} show that the IC have drawbacks: they tend to select more variables than necessary and are sensitive to small changes in the data. \cite{zhanghuang08, knightfu00, meinshausenbuhlmann06, zhaoyu06} show that $\mathcal{L}_1$-penalized regression is consistent in different settings. \citet{ huangall08, hoerlkennard70} show the consistency of penalized regression with $\gamma > 1$. \cite{zou06, caner09, friedman10} propose variants of penalized regression in different scenarios and \cite{fu98} compares different penalized regressions using a simulation study. Alternative approaches to model selection, such as combinatorial search algorithms may be computationally challenging to implement, especially with high-dimensional data.\footnote{\cite{chickering04} point out that the best subset selection method is unable to deal with a large number of variables, heuristically 30 at most.}

\subsection{Major results and contribution}

A central idea in this paper is that the analysis of GA is closely connected to the bias-variance trade-off. We show below that, loosely speaking, a model with superior GA typically achieves a better balance between bias and variance. Put another way, GA can be though of as a way to understand the properties of model selection methods. By the same token, model selection can be thought of as a tool for GA: if the goal is to improve the GA of a model, model selection is necessary. From the joint perspective of GA and model selection, we unify the class of penalized regressions with $\gamma>0$, and show that the finite-sample and asymptotic properties of penalized regression are closely related to the concept of GA.

The \textbf{first contribution} of this paper is to derive an upper bound for the prediction error on out-of-sample data based on the in-sample prediction error of the extremum estimator and to characterize the trade-off between in-sample fit and out-of-sample fit. As shown in \citet{vc71, vc71b, shalizi2011, smale2009, hu2009}, the classical concentration inequalities underlying GA analysis focus on the relation between the population error and the empirical in-sample error. In contrast, we quantify a bound for the prediction error of the extremum estimate from in-sample data on any out-of-sample data. The bound also highlights that the finite-sample and asymptotic properties of many penalized estimators can be framed in terms of GA. Classical methods to improve GA involve computing discrete measures of model complexity, such as the VC dimension, Radamacher dimension or Gaussian complexity. Discrete complexity measures are hard to compute and often need to be estimated. In contrast, we show that finite-sample GA analysis is easy to implement via validation or cross-validation and possesses desirable finite-sample and asymptotic properties for model selection.

A \textbf{second contribution} of the paper is to show that GA analysis may be used to choose the tuning hyper-parameter for validation (i.e., the ratio of training sample size to test sample size) or cross-validation (i.e., the number of folds $K$). Existing research has studied cross-validation for parametric and nonparametric model estimation \citep{hall1991, hall2011, stone74, stone77}. In contrast, by adapting the classical error bound inequalities that follow from GA analysis, we derive the optimal tuning parameters for validation and cross-validation in a model-free setting. We also show how $K$ affects the bias-variance trade-off for cross-validation: a higher $K$ increases the variance and lowers the bias.

A \textbf{third contribution} of the paper is use GA analysis to derive the finite-sample and asymptotic properties, in particular that of $\mathcal{L}_2$-consistency, for any penalized regression estimate. Various properties for penalized regression estimators have previously been established, such as probabilistic consistency or the oracle property \citep{knightfu00, zhaoyu06, candestao07, meinshausenyu09, bickeletal09}. GA analysis reveals that similar properties can be established more generally for a wider class of estimates from penalized regression. We also show that the $\mathcal{L}_2$-difference between the OLS estimate and any penalized regression estimate can be quantified by their respective GAs.

Lastly, a \textbf{fourth contribution} of the paper is that our results provide a platform to extend GA analysis to time series, panel data and other non-i.i.d.\ data. The literature has demonstrated that the major tools of GA analysis can be extended to non-i.i.d.\ data: many researchers have generalized the VC inequality \citep{vc71, vc71b}---one of the major tools in this paper to analyze i.i.d.\ data---to panel data and times series. Other studies show a number of ways to control for heterogeneity, which guarantees the validity of GA analysis. In addition, other tools used in this paper, such as the Glivenko-Cantelli theorem, the Hoeffding and von Bahr-Esseen bounds, have been shown to apply to non-i.i.d.\ data.\footnote{See, for example, \citet{yu1993glivenko, wellner1981glivenko, tang2007hoeffding}.} Hence, by implementing our framework with the techniques listed above, we can extend the results in this paper to a rich set of data types and scenarios.

The paper is organized as follows. In Section~2 we review the concept of GA, its connection to validation and cross-validation and derive upper bounds for the finite-sample GA of extremum estimates. In Section~3, we implement the results in the case of penalized regression and show that properties of penalized regression estimates can be explained and quantified by their GA. We also prove the $\mathcal{L}_{2}$-consistency of penalized regression estimates for both $p\leqslant n$ and $p>n$ cases. Further, we establish the finite-sample upper bound for the $\mathcal{L}_{2}$-difference between penalized and unpenalized estimates based on their respective GAs. In Section~4, we use simulations to demonstrate the ability of penalized regression to control for overfitting. Section~5 concludes with a brief discussion of our results. Proofs are contained in Appendix~1 and graphs of the simulations are in Appendix~2.


\section{Generalization ability and the upper bound for finite-sample generalization errors}

\subsection{Generalization ability, generalization error and overfitting}

In econometrics, choosing the best approximation to data often involves measuring a loss function, $Q(b\vert y_i,\mathbf{x}_i)$, defined as a functional that depends on the estimate $b$ and the sample points $(y_i,\mathbf{x}_i)$. The population error functional is defined as
\[
	\mathcal{R}(b\vert Y,X)=\int Q(b\vert y,\mathbf{x})\mathrm{d}F(y,\mathbf{x})
\]
where $F(y,\mathbf{x})$ is the joint distribution of $y$ and $\mathbf{x}$. Without knowing the distribution $F(y,\mathbf{x})$ a priori, we define the empirical error functional as follows
\[
	\mathcal{R}_{n}(b\vert Y,X) = \frac{1}{n}\;\sum_{i=1}^n\;Q(b\vert y_i,\mathbf{x}_i).
\]
For example, in the regression case, $b$ is the estimated parameter vector and $\mathcal{R}_{n}(b\vert Y,X)=\frac{1}{n}\sum_{i=1}^n(y_i - \hat{y}_i)^{2}$.

When estimation involves minimizing the in-sample empirical error, we have the extremum estimator \citep{amemiya1985advanced}. In many settings, however, minimizing the in-sample empirical error does not guarantee a reliable model. In regression, for example, often the $R^2$ is used to measure goodness-of-fit for in-sample data.\footnote{For regression, $R^{2}=1-\mathcal{R}_{n}(b\vert Y,X)/(\mbox{TSS}/n)$ where $\mathcal{R}_{n}(b\vert Y,X)=\frac{1}{n}\sum_{i=1}^n(y_i - \hat{y}_i)^{2}$ and $\mbox{TSS}= \sum_{i=1}^n (y-\bar{y})^{2}$.} However, an estimate with a high in-sample $R^{2}$ may fit out-of-sample data poorly, a feature commonly referred to as \textbf{overfitting}: the in-sample estimate is too tailored for the sample data, compromising its out-of-sample performance. As a result, in-sample fit may not be a reliable indicator of the general applicability of the model.

Thus, \citet{vc71} refer to the \textbf{generalization ability} (GA) of a model; a measure of how an extremum estimator performs on out-of-sample data. GA can be measured several different ways. In the case where $X$ and $Y$ are directly observed, GA is a function of the difference between the actual and predicted $Y$ for out-of-sample data. In this paper, GA is measured by the out-of-sample empirical error functional.
\begin{defn}[Subsamples, empirical training error and empirical generalization error]\end{defn}

	\begin{enumerate}

    \item   Let $(y, \mathbf{x})$ denote a sample point from $F(y, \mathbf{x})$, where $F(y,\mathbf{x})$ is the joint distribution of $(y,\mathbf{x})$. Given a sample $(Y,X)$, the \textbf{training set} $(Y_{t},\,X_{t})\in\mathbb{R}^{n_t \times p}$ refers to data used for the estimation of $b$ and the \textbf{test set} $(Y_{s},\,X_{s})\in\mathbb{R}^{n_s \times p}$ refers to data \emph{not} used for the estimation of $b$. Let $\widetilde{n}=\min\{n_s,n_t\}$. The \textbf{effective sample size} for the training set, test set and the total sample, respectively, is $n_t/p$, $n_s/p$ and $n/p$.

    \item   Let $\Lambda$ denote the space of all models. The \textbf{loss function} for a model $b\in\Lambda$ is $Q(b\vert y_i,\mathbf{x}_i),\,i=1,\ldots,n$. The \textbf{population error functional} for $b\in\Lambda$ is $\mathcal{R}(b\vert Y,X)=\int Q(b\vert y,\mathbf{x})\mathrm{d}F(y,\mathbf{x})$. The \textbf{empirical error functional} is $\mathcal{R}_{n}(b\vert Y,X) =\frac{1}{n}\;\sum_{i=1}^n\;Q(b\vert y_i,\mathbf{x}_i)$.

    \item   Let $b_{train}\in\Lambda$ denote an extremum estimator. The \textbf{empirical training error (eTE)} for $b_{train}$ is $\min_b\mathcal{R}_{n_t}(b|Y_{t},X_{t}) = \mathcal{R}_{n_t}(b_{train}|Y_{t},X_{t})$, where $b_{train}$ minimizes $\mathcal{R}_{n_t}(b|Y_{t},X_{t})$. The \textbf{empirical generalization error (eGE)} for $b_{train}$ is $\mathcal{R}_{n_s}(b_{train}|Y_{s},X_{s})$. The population error for $b_{train}$ is $\mathcal{R}(b_{train}|Y,X)$.
		
    \item	For $K$-fold cross-validation, denote the training set and test set in the $q$th round, respectively, as $(X_{t}^q,Y_{t}^q)$ and $(X_s^q,Y_s^q)$. In each round, the sample size for the training set is $n_t = n(K-1)/K$ and the sample size for the test set is $n_s = n/K$.

	\end{enumerate}

The most important assumptions for the analysis in this section of the paper are as follows.

\bigskip
\noindent
\textbf{Assumptions}

\begin{enumerate}

    \item[\textbf{A1.}]   In the probability space $(\Omega,\mathcal{F},P)$, we assume $\mathcal{F}$-measurability of the loss $Q(b\vert y, \mathbf{x})$, the population error $\mathcal{R}(b|Y,X)$ and the empirical error $\mathcal{R}_{n}(b\vert Y,X)$, for any $b\in\Lambda$ and any sample point $(y, \mathbf{x})$. All loss distributions have a closed-form, first-order moment.

    \item[\textbf{A2.}]   The sample $(Y,X)$ is independently distributed and randomly chosen from the population. In cases with multiple random samples, both the training set and the test set are randomly sampled from the population. In cases with a single random sample, both the training set and the test set are randomly partitioned from the sample.

    \item[\textbf{A3.}]   For any sample, the extremum estimator $b_{train}\in\Lambda$ exists. The in-sample error for $b_{train}$ converges in probability to the minimal population error as $n\rightarrow\infty$.

\end{enumerate}

A few comments are in order for assumptions A1--A3. The loss distribution assumption A1 is merely to simplify the analysis. The existence and convergence assumption A3 is standard (see, for example, \citet{neweymcfadden94}). The independence assumption in A2 is not essential because GA analysis is valid for both i.i.d.\ and non-i.i.d.\ data. While the original research in \citet{vc74, vc74b} imposes the i.i.d.\ restriction on GA, subsequent work has generalized their results to cases where the data are dependent or not identically distributed.\footnote{See, for example, \citet{yu1994rates, cesa2004generalization, shalizi2011, smale2009, mohri2009rademacher, kakade2009generalization}.} Others have shown that if heterogeneity is due to an observed random variable, the variable may be added to the model to control for the heterogeneity while if the heterogeneity is related to a latent variable, various approaches---such as the hidden Markov model, mixture modelling or factor modelling---are available for heterogeneity control.\footnote{See \citet{yashin1986, skrondal2004generalized, wang2005learning, yu2009learning, pearl2015detecting}.} Either way, GA analysis is valid owing to the controls for heterogeneity. In this paper, due to the different measure-theory setting for dependent data, we focus on the independent case as a first step. In a companion paper \citep{xuall16b}, we specify the time series mixing type and the types of heterogeneity across individuals to generalize the results in this paper to time series and panel data.  Lastly, given A1--A3, both the eGE and eTE converge to the population error:
\[
\mathrm{lim}_{\widetilde{n}\rightarrow\infty}\mathcal{R}_{n_t}(b_{train}|Y_{t},X_{t}) = \mathrm{lim}_{\widetilde{n}\rightarrow\infty}\mathcal{R}_{n_s}(b_{train}|Y_{s},X_{s}) = \mathcal{R}(b_{train}|Y,X)
\]

Typically two methods are implemented to compute the eGE of an estimate: validation and cross-validation. For validation when only one sample is available, the sample is randomly partitioned into a training set and a test set; if multiple samples are available, some are chosen as test sets and others as training sets. Either way, we use training set(s) for estimation and test set(s) to compute the eGE for the estimated model, yielding the validated eGE.

$K$-fold cross-validation may be thought of as `averaged multiple-round validation'. For cross-validation, the full sample is randomly partitioned into $K$ subsamples or folds.\footnote{Typically, $K=5$, 10, 20, $40\mbox{ or }N$.} One fold is chosen to be the test set and the remaining $K-1$ folds comprise the training set. Following extremum estimation on the training set, the fitted model is applied to the test set to compute the eGE. The process is repeated $K$ times, with each of the $K$ folds getting the chance to play the role of the test set while the remaining $K-1$ folds are used as the training set. In this way, we obtain $K$ different estimates of the eGE for the fitted model. The $K$ estimates of the eGE are averaged, yielding the cross-validated eGE.

Cross-validation uses each data point in both the training and test sets. Cross-validation also reduces resampling error by running the validation $K$ times over different training and test sets. Intuitively this suggests that cross-validation is more robust to resampling error and should perform at least as well as validation. In Section~3, we study the generalization ability of penalized extremum estimators in both the validation and cross-validation cases and discuss the difference between them in more detail.

\subsection{The upper bound for the empirical generalization error}

The traditional approach to model selection in econometrics is to use the AIC, BIC or HQIC, which involves minimizing the eTE and applying a penalty term to choose among alternative models. Based on a broadly similar approach, \citet{vc71,vc71b} consider model selection from the perspective of generalization ability. \citet{vc71,vc71b} posit there are essentially two reasons why a model estimated on one sample may have a weak generalization ability on another: the two samples may have different sampling errors, or the complexity of the model estimated from the original sample may have been chosen inappropriately.

To improve the generalization ability of a model, \citet{vc71,vc71b} propose minimizing the upper bound of the population error of the estimate as opposed to minimizing the eTE. The balance between in-sample fit and out-of-sample fit is formulated by \citet{vc74b} using the Glivenko-Cantelli theorem and Donsker's theorem for empirical processes. Specifically, the relation between $\mathcal{R}_{n}(b|Y,X)$ and $\mathcal{R}(b|Y,X)$ is summarized by the so-called \textbf{VC inequality} \citep{vc74b} as follows.
%
%
\begin{lem}[The upper bound of the population error (the VC inequality)]
    Under \textbf{A1} to \textbf{A3}, the following inequality holds with probability $1-\eta$, $\forall b \in \Lambda$, and $\forall n\in\mathbb{N}^{+}$,
	\begin{equation}
		\mathcal{R}(b|Y,X) \leqslant\mathcal{R}_{n_t}(b|Y_{t},X_{t})
		+\frac{\sqrt{\epsilon}}{1-\sqrt{\epsilon}}\mathcal{R}_{n_t}(b|Y_{t},X_{t})
		\label{eq:lem2.1}
	\end{equation}
    where $\mathcal{R}(b|Y,X)$ is the population error, $\mathcal{R}_{n_t}(b|Y_t,X_t)$ is the training error from the model $b$, $\epsilon=(1/n_t)[h\ln(n_t/h)+h-\ln\left(\eta\right)]$,
    and $h$ is the \textbf{VC dimension}.
\label{lem2.1}
\end{lem}

A few comments are in order for the VC inequality, eq.~(\ref{eq:lem2.1}).

\begin{figure}
	\centering
	\subfloat[\label{fig:VCineqov}overfitting]
	{\includegraphics[width=0.35\paperwidth]{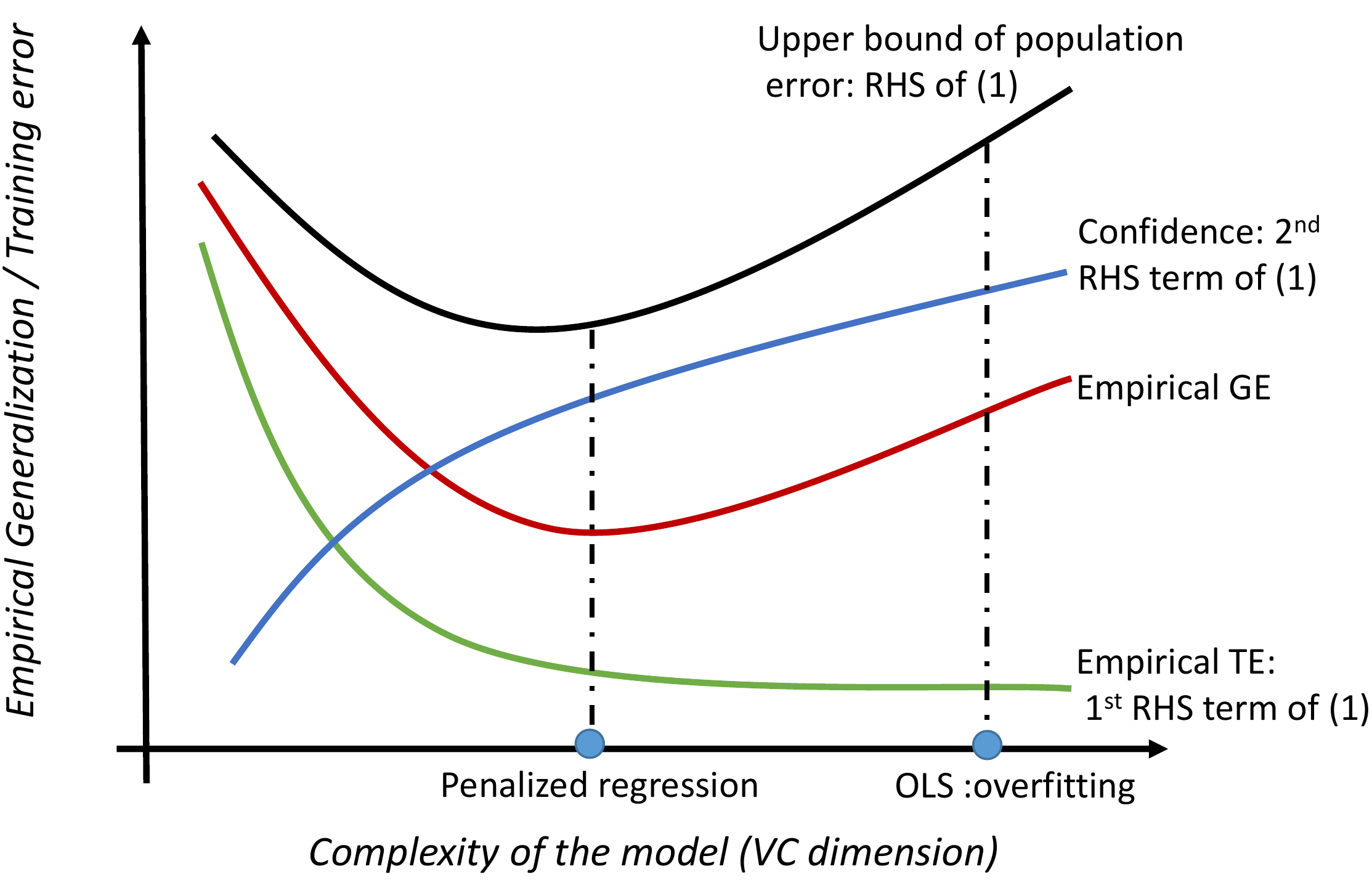}}
	\subfloat[\label{fig:VCinequn}underfitting]
	{\includegraphics[width=0.35\paperwidth]{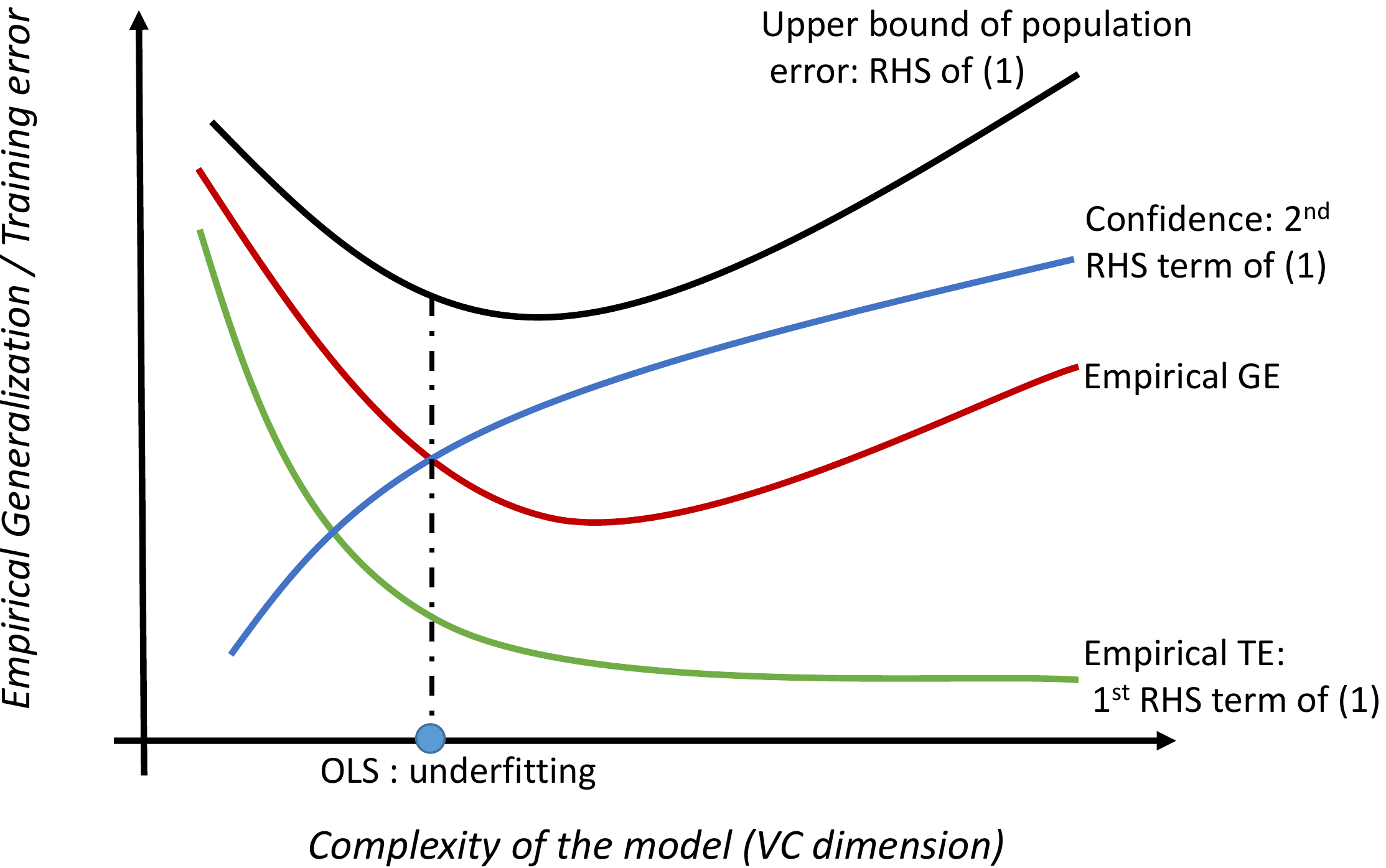}}

	\caption{\label{fig:VCineq}The VC inequality and eGE}

\end{figure}

\begin{enumerate}

    \item   As shown in Figure~\ref{fig:VCineq}, the RHS of eq.~(\ref{eq:lem2.1}) establishes an upper bound for the population error based on the eTE and the VC dimension $h$. When the effective sample size for the training set ($n_t/h$) is very large, $\epsilon$ is very small, the second term on the RHS of (\ref{eq:lem2.1}) becomes small, and the eTE is close to the population error. In this case the extremum estimator has a good GA. However, if the effective sample size $n_t/h$ is small (i.e., the model is very complicated), the second term on the RHS of (\ref{eq:lem2.1}) becomes larger. In such situations a small eTE does not guarantee a good GA, and overfitting becomes more likely.

    \item   The VC dimension $h$ is a more general measure of model complexity than the number of parameters, $p$, which does not readily extend to nonlinear or non-nested models. While $h$ reduces to $p$ directly for generalized linear models, $h$ can also be used to partially order the complexity of nonlinear or non-nested models by summarizing their geometric complexity.\footnote{In empirical processes, several other geometric complexity measures are connected to or derived from the VC dimension, such as the minimum description length (MDL) score, the Rademacher dimension (or complexity), Pollard's pseudo-dimension and the Natarajan dimension. Most of these measures, like the VC dimension, are derived and generalized from the Glivenko-Cantelli class of empirical processes.} As a result, eq.~(\ref{eq:lem2.1}) can be implemented as a tool for both nonlinear and non-nested model selection.

    \item   Eq.~(\ref{eq:lem2.1}) can be generalized to non-i.i.d.\ cases. While the VC inequality focuses on the relation between the population error and the eTE in the i.i.d.\ case, \citet{shalizi2011} generalizes the VC inequality for $\alpha$- and $\beta$-mixing stationery time series while \citet{smale2009} generalizes the VC inequality for panel data. Moreover, \citet{yashin1986, skrondal2004generalized, wang2005learning, yu2009learning, pearl2015detecting} show that heterogeneity can be controlled by implementing the latent variable model or by adding the variable causing heterogeneity into the model, implying eq.~(\ref{eq:lem2.1}) is valid.

\end{enumerate}

Based on the VC inequality, \citet{vc71} propose that minimizing the RHS of (\ref{eq:lem2.1}), the upper bound of the population error, reduces overfitting and improves the GA of the extremum estimator. However, this may be hard to implement because it can be difficult to calculate the VC dimension for anything other than linear models. In practice, statisticians have implemented GA analysis by minimizing the eGE using validation or cross-validation. For example, cross-validation is used to implement many penalty methods, such as the lasso-type estimators, ridge regression or bridge estimators. Clearly, however, the eGE and the population error are not the same thing. Thus, the properties of the minimum eGE, such as its variance, consistency and convergence rate are of particular interest in the present context. By adapting and modifying eq.~(\ref{eq:lem2.1}), we propose the following inequalities that analyze the relation between the eGE and the eTE in finite samples.
%
%
\begin{thm}[The upper bound of the finite-sample eGE for the extremum estimator]
    Under \textbf{A1} to \textbf{A3}, the following upper bound for the eGE holds with probability at least $\varpi(1-1/n_t)$, $\forall\varpi\in\left(0,1\right)$.
	\begin{equation}
		\mathcal{R}_{n_s}(b_{train}|Y_{s},X_{s}) \leqslant
		\frac{\mathcal{R}_{n_t}(b_{train}|Y_{t},X_{t})}{(1-\sqrt{\epsilon})} + \varsigma,
		\label{eq:thm2.1}
	\end{equation}
    where $\mathcal{R}_{n_s}(b_{train}|Y_{s},X_{s})$ is the eGE and $\mathcal{R}_{n_t}(b_{train}|Y_t,X_t)$ the eTE for the extremum estimator $b_{train}$, $\epsilon$ is defined in Lemma~\ref{lem2.1},
	\[
		\varsigma =
			\left\{
				\begin{array}{ll}
                 \sqrt[\nu]{2} \tau
                \left(\mathbb{E}\left[Q(b_{train}|Y_{s},X_{s})\right]\right)/
                (\sqrt[\nu]{1-\varpi}\cdot n_s^{1-1/{\nu}})
					& \mbox{if } \nu\in(1,2] \\[12pt]
				\frac{B}{n_s}\ln\sqrt{2/(1-\varpi)}
					& \mbox{if } Q(\cdot) \in(0,B] \mbox{ and B is bounded} \\[12pt]
				\mathrm{var}[Q(b_{train}\vert y,\mathbf{x})] / (n(1-\varpi))
					& \mbox{if } \nu\in(2,\infty)
               \end{array}
            \right.
	\]
    and
	\[
		\tau\geqslant
		\sup \frac{[\int\left(Q(b|y, \mathbf{x})\right)^{\nu}
		\mathrm{d}F(y, \mathbf{x})]^{1/{\nu}}}
		{\int Q(b|y, \mathbf{x})\mathrm{d}F(y, \mathbf{x})}.
	\]

\label{thm2.1}
\end{thm}

\bigskip
A few comments follow from Theorem~\ref{thm2.1}.

\begin{figure}

    \centering
	\subfloat[\label{fig:TEGEfinite}the eTE-eGE trade-off in finite-sample]
	{\includegraphics[width=0.35\paperwidth]{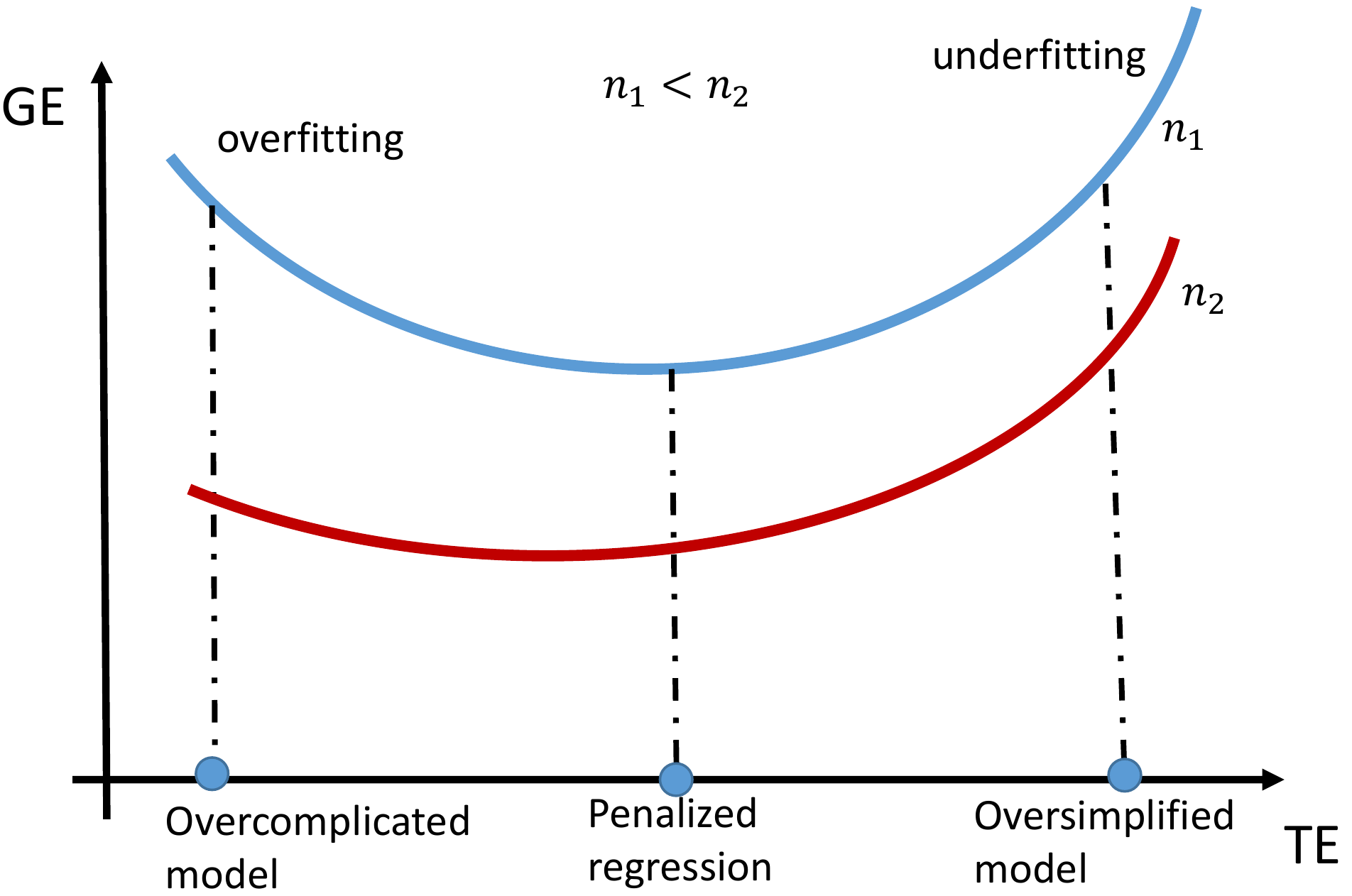}}
	\subfloat[\label{fig:TEGEasym}the eTE-eGE trade-off in asymptotoics]
	{\includegraphics[width=0.35\paperwidth]{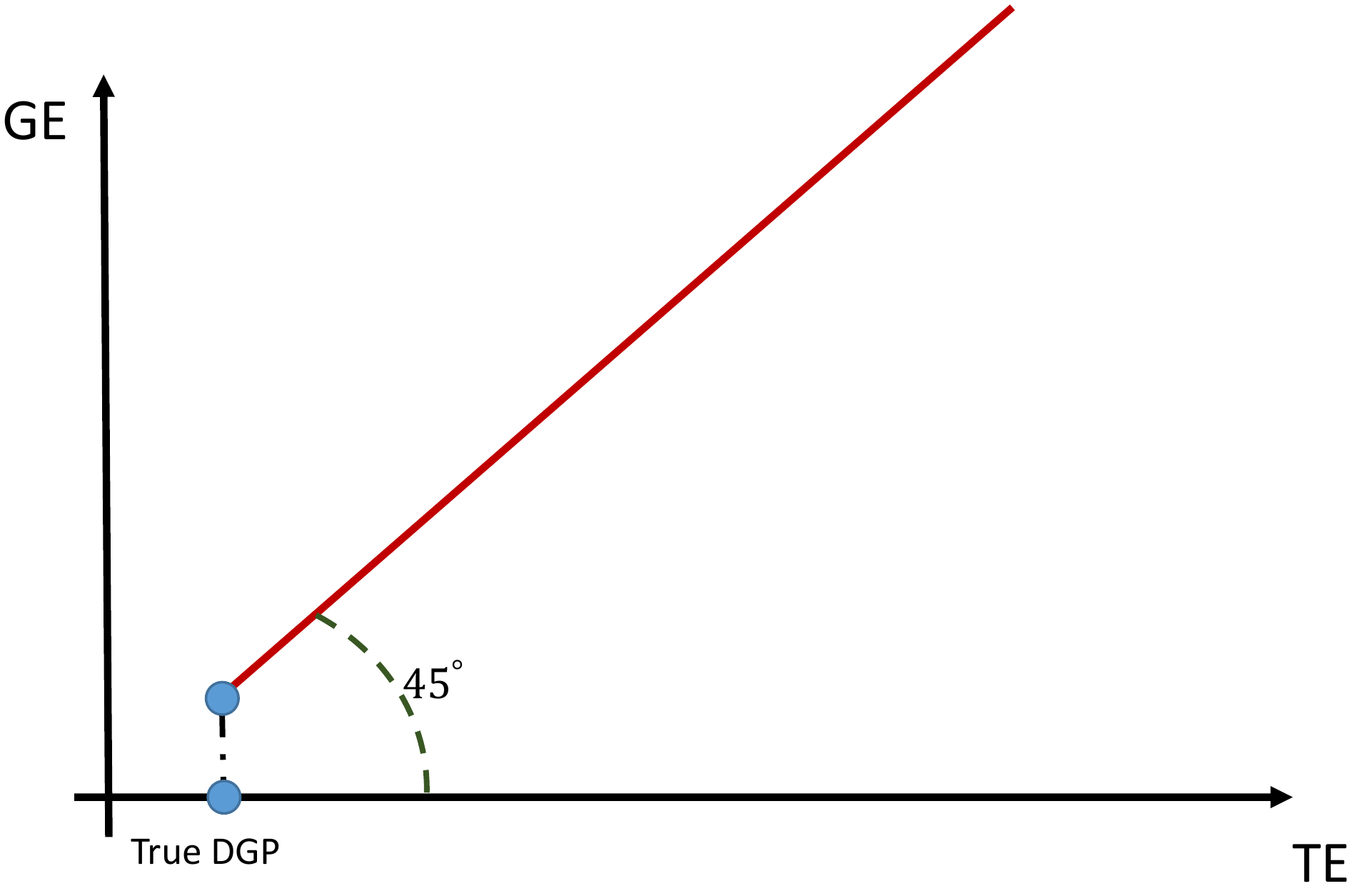}}

	\caption{\label{fig:TEGE}Schematic diagram of the trade-off between eGE and eTE}

\end{figure}

\begin{itemize}

    \item   (\textit{Upper bound of the finite-sample GA}) eq.~(\ref{eq:thm2.1}) establishes the upper bound of the eGE from any out-of-sample data of size $n_s$ based on the eTE from any in-sample data of size $n_t$. Unlike the classical bound in Lemma~\ref{lem2.1}, which captures the relation between the population error and the eTE, eq.~(\ref{eq:thm2.1}) establishes inequalities to quantify the upper bound of the finite sample eGE. Usually, we need to use validation or cross-validation to measure the eGE of a model with new data. However, because the RHS of eq.~(\ref{eq:thm2.1}) is directly computable it may be used as a measure of finite-sample eGE, avoiding the need for validation.

    \item   (\textit{The eGE-eTE trade-off in model selection}) eq.~(\ref{eq:thm2.1}) also characterizes the trade-off between eGE and eTE for model selection in both the finite sample and asymptotic cases. In Figure~\ref{fig:TEGEasym}, the population eGE, population eTE and population error are expected to be identical in asymptotic case. Hence, minimizing eTE can directly lead to the true DGP in the population. In contrast, as illustrated in Figure~\ref{fig:TEGEfinite}, in finite samples, an overcomplicated model with low $n_t/h$ would have a small eTE for the data whereas eq.~(\ref{eq:thm2.1}) show that the upper bound of the eGE on new data will be large. Hence, the overcomplicated model will \textit{overfit} the in-sample data and typically have a poor GA. In contrast, an oversimplified model with high $n_t/h$, typically cannot adequately recover the DGP and the upper bound of the eGE will also be large. As a result, the oversimplified model will \textit{underfit}, fitting both the in-sample and out-of-sample data poorly. Thus, the complexity of a model introduces a trade-off between the eTE and eGE in model selection.

    \item   (\textit{GA and distribution tails}) eq.~(\ref{eq:thm2.1}) also shows how the tail of the error distribution affects the upper bound of the eGE. If the loss distribution $Q(\cdot)$ is bounded or light-tailed, the second term of eq.~(\ref{eq:thm2.1}), $\varsigma$, is mathematically simple and converges to zero at the rate $1/n_s$. If the loss function is heavy-tailed and $\mathcal{F}$-measurable, $\nu$, the highest order of the population moment that is a closed-form for the loss distribution,\footnote{It is closed-form because owing to \textbf{A1}, which guarantees closed-form, first-order moments for all loss distribution in the paper.} can be used to measure the heaviness of the loss distribution tail, a smaller $\nu$ implying a heavier tail. In the case of a heavy tail, the second term of  eq.~(\ref{eq:thm2.1}), $\varsigma$, becomes mathematically complicated and its convergence rate decreases to $1/n_{s}^{1-1/\nu}$. Hence, eq.~(\ref{eq:thm2.1}) shows that the heavier the tail of the loss distribution, the higher the upper bound of the eGE and the harder it is to control GA in finite samples. In the extreme case with $\nu = 1$, there is no way to adapt eq.~(\ref{eq:thm2.1}).

\end{itemize}

Essentially, validation randomly partitions the data into a training set and a test set, yielding an estimate on the training set that is used to compute the eGE of the test set. Eq.~(\ref{eq:thm2.1}) measures the upper bound the eGE on the test set from the model estimated on the training set with a given eTE and $h$. In other words, eq.~(\ref{eq:thm2.1}) directly measure GA using validation. Furthermore, a similar bound to eq.~(\ref{eq:thm2.1}) can be established for $K$-fold cross-validation.

%
%
\begin{thm}[The upper bound of the finite-sample eGE for the extremum estimator under cross-validation]
Under \textbf{A1} to \textbf{A3}, the following upper bound for the eGE holds with probability at least $\varpi(1-1/K)$, $\forall\varpi\in\left(0,1\right)$.

		\begin{equation}
			\frac{1}{K}\sum_{j=1}^{K}\mathcal{R}_{n_s}(b_{train}|Y_{s}^j,X_{s}^j)
			\leqslant
			\frac{\frac{1}{K}\sum_{q=1}^{K}\mathcal{R}_{n_t}(b_{train}|Y_{t}^q,X_{t}^q)}
			{(1-\sqrt{\epsilon})} + \varsigma_{cv},
			\label{eq:thm2.2}
		\end{equation}
        where $\mathcal{R}_{n_s}(b_{train}|Y_{s}^j,X_{s}^j)$ is the eGE of $b_{train}$ in $j$th round of		validations, $\mathcal{R}_{n_t}(b_{train}|Y_{t}^q,X_{t}^q)$ is the eTE of
		$b_{train}$ in $q$th round of validations, and
		\[
			\varsigma_{cv} =
				\left\{
					\begin{array}{ll}
						\sqrt[\nu]{2} \tau
                        \;\mathcal{R}(b_{train}|Y,X)
						/ (\sqrt[\nu]{1-\varpi}\cdot n_s^{1-1/{\nu}})
						    & \mbox{if } \nu \in(1,2] \\[12pt]
						B \ln\sqrt{2/(1-\varpi)} / n_s
						    & \mbox{if } Q(\cdot) \in(0,B] \mbox{ and B is bounded} \\[12pt]
						\mathrm{var}[Q(b_{train}\vert y, \mathbf{x})] / (n_s^2 (1-\varpi))
                            &	\mbox{if } \nu \in (2,\infty)
					\end{array}
                \right.
		\]

\label{thm2.2}
\end{thm}

The errors generated by cross-validation are affected both by sampling randomness from the population and by sub-sampling randomness that arises from partitioning the sample into folds. Thus, the errors from cross-validation are potentially more volatile than the usual errors from estimation. Theorem~\ref{thm2.2} provides an upper bound for the average eGE under cross-validation, which offers a way to characterize the effect of sub-sampling randomness and suggests a method to approximate the GA from cross-validation. The following comments summarize the implications of eq.~(\ref{eq:thm2.2}).

\begin{figure}

    \centering
	\subfloat[\label{fig:Ktradeoff1}eGE in each round of cross-validation]
	{\includegraphics[width=0.37\paperwidth]{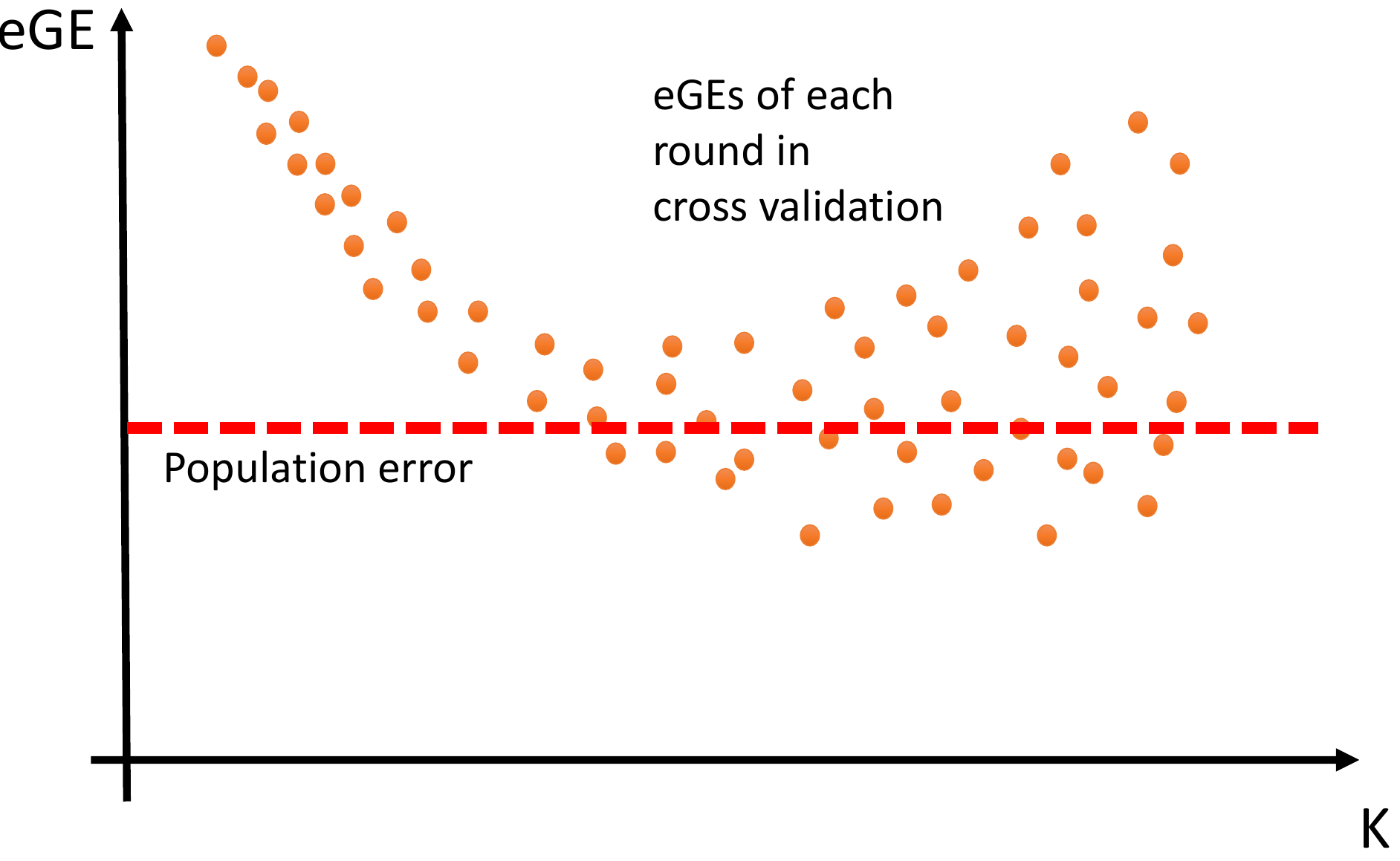}}
	\centering	
	\subfloat[\label{fig:Ktradeoff2}average eGE from $K$ rounds of cross-validation]
	{\includegraphics[width=0.37\paperwidth]{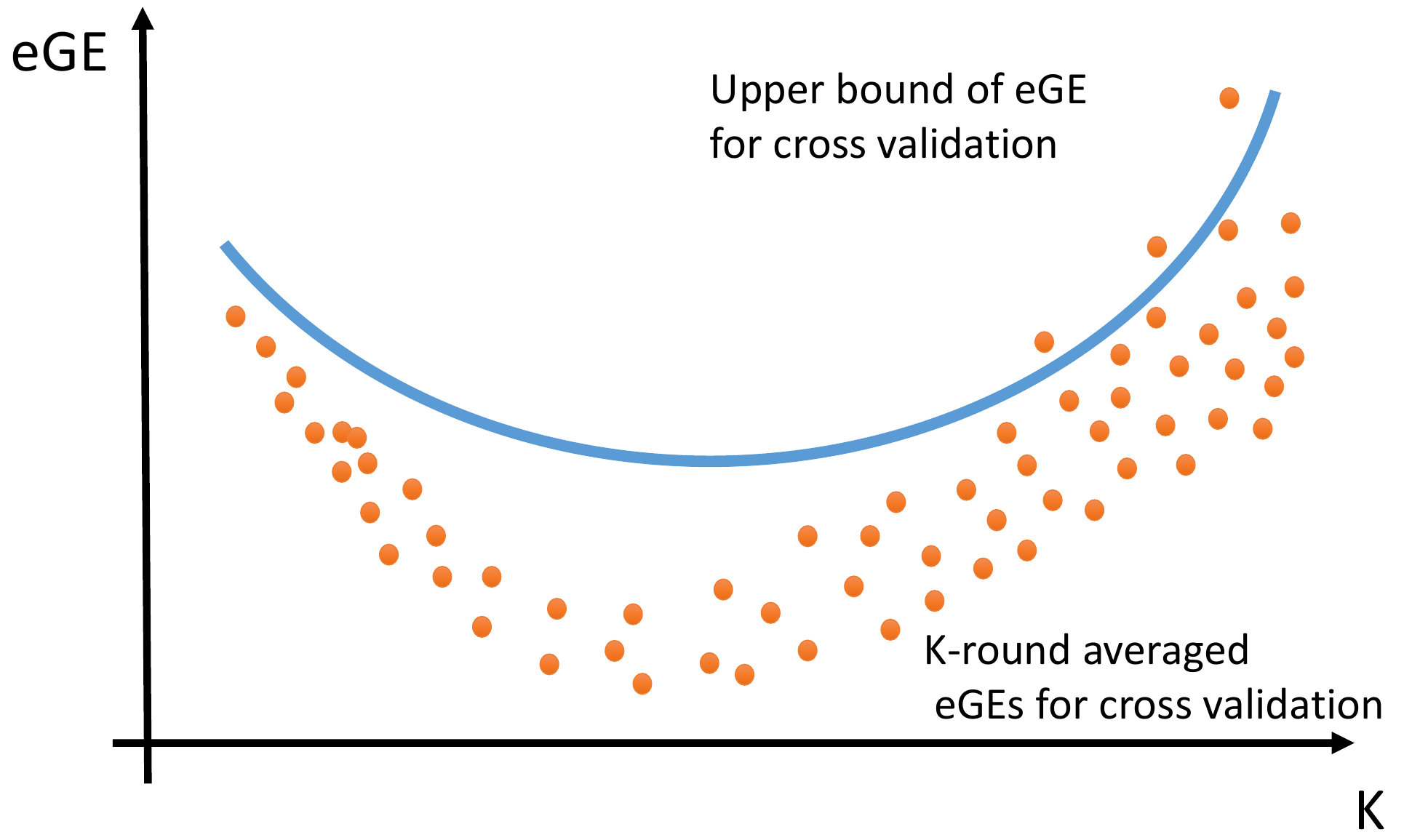}}
	
    \caption{\label{fig:Ktradeoff}Representation of the bias-variance trade-off for cross-validation eGE}

\end{figure}

\begin{enumerate}

    \item   (\textit{The upper bound of the eGE}) Similar to eq.~(\ref{eq:thm2.1}), eq.~(\ref{eq:thm2.2}) serves as the upper bound of the averaged eGE generated by cross-validation. Both equations show the eTE-eGE trade-off and reveal the effect of a heavy tail on GA.

    \item  (\textit{Tuning the cross-validation hyperparameter $K$}) Eq.~(\ref{eq:thm2.2}) characterizes how the hyperparameter $K$ affects the averaged eGE from cross-validation (also called the cross-validation error in the literature). As explained above, the random partitioning in cross-validation introduces sub-sampling randomness. With a given sample and fixed $K$, sub-sampling randomness will produce a different averaged eGE each time cross-validation is performed. When $K$ changes, the size of each fold changes, implying the training and test sets also change. When $K$ is large, the test sets become small, increasing sub-sampling randomness. When $K$ is small, the training sets become small, increasing sub-sampling randomness. For extremum estimators like OLS, the bias-variance trade-off is straightforward to analyze for different $p$ because the sample is fixed. In contrast, the sub-sampling randomness introduced by cross-validation, the bias-variance trade-off for averaged eGE of cross-validation cannot be studied with the given training and test set when $K$ changes. As a result, in order to characterize and control for the influence of sub-sampling randomness, we establish the bias-variance trade-off for cross-validation by its upper bound, after running cross validation multiple times, as is illustrated in Figure~\ref{fig:Ktradeoff}.

	\begin{enumerate}

        \item \textit{(Large bias, small variance)} When $K$ is small, $n_t$ is smaller in each round of in-sample estimation. Hence, as shown in Figure~\ref{fig:Ktradeoff1}, the eTE in each round, $\mathcal{R}_{n_t}(b_{train}|Y_{t}^q,X_{t}^q)/(1-\sqrt{\epsilon})$, is more biased from the population error. As shown in Figure~\ref{fig:Ktradeoff2}, the $K$-round averaged eTE,    		$\frac{1}{K}\sum_{q=1}^{K}\mathcal{R}_{n_t}(b_{train}|Y_{t}^q,X_{t}^q)/(1-\sqrt{\epsilon})$, is more biased away from the true population error as $K$ gets smaller. As a result, the RHS of eq.~(\ref{eq:thm2.2}) suffers more from finite-sample bias. However, since small $K$ implies that more data is used for eGE calculation in each round ($n_s$ is not very small), in each round the eGE on the test set should not be very volatile. Thus, the $K$-round averaged eGE for cross-validation is not very volatile, which is shown by the fact that $\varsigma_{cv}$ is not very large in eq.~(\ref{eq:thm2.2}).

        \item \textit{(Small bias, large variance)} When $K$ is large, $n_s$ is small and the test set in each round is small. Hence, with large $K$, the eGE in each round may be hard to bound from above, which implies that the averaged eGE from $K$ rounds is more volatile. As a result, $\varsigma_{cv}$ tends to be large. However, with large $K$, the RHS term $\frac{1}{K}\sum_{q=1}^{K}\mathcal{R}_{n_t} (b_{train}|Y_{t}^q,X_{t}^q) / (1-\sqrt{\epsilon})$ tends to be closer to the true population error, so the averaged eGE suffers less from bias.
	\end{enumerate}	
	
As shown in Figure~\ref{fig:Ktradeoff2}, the averaged eGE from cross-validation follows a typical bias-variance trade-off by value of $K$. If $K$ is small, the averaged eGE is computationally cheap and less volatile but more biased away from the population error. As $K$ gets larger, the averaged eGE becomes computationally expensive and more volatile but less biased away from the population error. This result exactly matches the \citet{kohavi1995study} simulation study. More specifically, by tuning $K$ to the lowest upper bound, we can find the $K$ that maximizes the GA from cross-validation.

\end{enumerate}

Theorems~\ref{thm2.1} and~\ref{thm2.2} establish the finite-sample and asymptotic properties of GA analysis for any extremum estimator. In finite-sample analysis, the results capture the trade-off between eGE and eTE, which can be used to measure the GA of an econometric model. In asymptotic analysis, eGE minimization is consistent. As a result, GA can be implemented as a criterion for model selection, and directly connects to the theoretical properties for model selection methods such as penalized extremum estimation, the various information criteria and maximum a posteriori (MAP) estimation. Minimizing eGE works especially well for penalized regression. As shown in Algorithm~1, penalized regression estimation returns a $b_{\lambda}$ for each $\lambda$. Each value of $\lambda$ generates a different model and a different eGE. Intuitively, Theorems~\ref{thm2.1} and~\ref{thm2.2} guarantee that the model with the minimum eGE from $\left\{ b_{\lambda}\right\}$ has the best empirical generalization ability. In the next section, we study the finite-sample and asymptotic properties of eGE for all penalized regressions.


\section{Finite-sample and asymptotic properties of eGE for penalized regression}

Using the classical concentration inequalities in Section~2, we established the upper bound for the finite-sample eGE of the extremum estimator given any random sample of any size. We also revealed the trade-off between eTE and eGE for model selection and derived the properties of eGE under validation and cross-validation. In this section, we apply the framework and results from Section~2 to penalized regression.

\subsection{Definition of penalized regression}

Firstly, we formally define penalized regression and its two most popular variants: ridge regression ($\mathcal{L}_2$-penalized regression) and the lasso ($\mathcal{L}_1$-penalized regression).

\begin{defn}[Penalized regression, $\mathcal{L}_2$-eGE and $\mathcal{L}_2$-eTE]\end{defn}

		\begin{enumerate}
            \item   (\textit{General form}) The general form of the objective function for penalized regression is as follows
                \begin{equation}
			        \min_{b_\lambda}
                    \frac{1}{n}\left(\left\Vert Y-Xb_{\lambda}\right\Vert_{2}\right)^{2}
			      	+\lambda\;\mathrm{Penalty}(\Vert b_{\lambda}\Vert_{\gamma}).
			    \label{lasso-type}
			    \end{equation}
                where $\mathrm{Penalty}(\Vert\cdot\Vert_{\gamma})$ stands for the penalty term, which is a function of the $\mathcal{L}_{\gamma}$ norm of the $b_{\lambda}$.

            \item   (\textit{$b_\lambda$ and $b_{pen}$}) We denote $b_{\lambda}$ to be the solution of eq.~(\ref{lasso-type}) given the value of the penalty parameter $\lambda$ while $b_{pen}$ is defined to be the model with the minimum eGE among all alternative $\{b_{\lambda}\}$, as in Algorithm~1 in Section~1.

            \item   (\textit{Lasso and ridge}) The objective functions for lasso ($\mathcal{L}_1$ penalty) and ridge regression ($\mathcal{L}_2$ penalty), respectively, are
			    \begin{equation}
			      	\min_{b_\lambda}
			      	\frac{1}{n}
			      	\left(\left\Vert Y-Xb_{\lambda}\right\Vert _{2}\right)^{2}
			      	+\lambda\Vert b_{\lambda}\Vert_{1},
			    \label{lasso-reg}
			    \end{equation}
                and
			    \begin{equation}
			      	\min_{b_\lambda}
			      	\frac{1}{n}
			      	\left(\left\Vert Y-Xb_{\lambda}\right\Vert _{2}\right)^{2}
			      	+\lambda\Vert b_{\lambda}\Vert_{2}.
			      	\label{ridge-reg}
			    \end{equation}

                \item   \textit{($\mathcal{L}_2$ error for regression)} the eTE and eGE for regression are defined in $\mathcal{L}_2$ form respectively as follows:
			        \begin{align*}
			      	    \mathcal{R}_{n_t}(b_{train}|Y_{t},X_{t}) & =\frac{1}{n_t}\,
			      	    \Vert Y_{t}-X_{t}b_{train}\Vert_{2}^{2} \\
			      	    \mathcal{R}_{n_s}(b_{train}|Y_{s},X_{s}) & =\frac{1}{n_s}\,
			      	    \Vert Y_{s}-X_{s}b_{train}\Vert_{2}^{2}
			        \end{align*}

		\end{enumerate}

\begin{figure}[ht]
	\centering
	\subfloat[\label{figpenreg:fig3}boundaries for $\mathcal{L}_{0}$, $\mathcal{L}_{0.5}$,
	$\mathcal{L}_1$, $\mathcal{L}_2$ and $\mathcal{L}_\infty$ penalties]
	{\includegraphics[width=0.25\paperwidth]{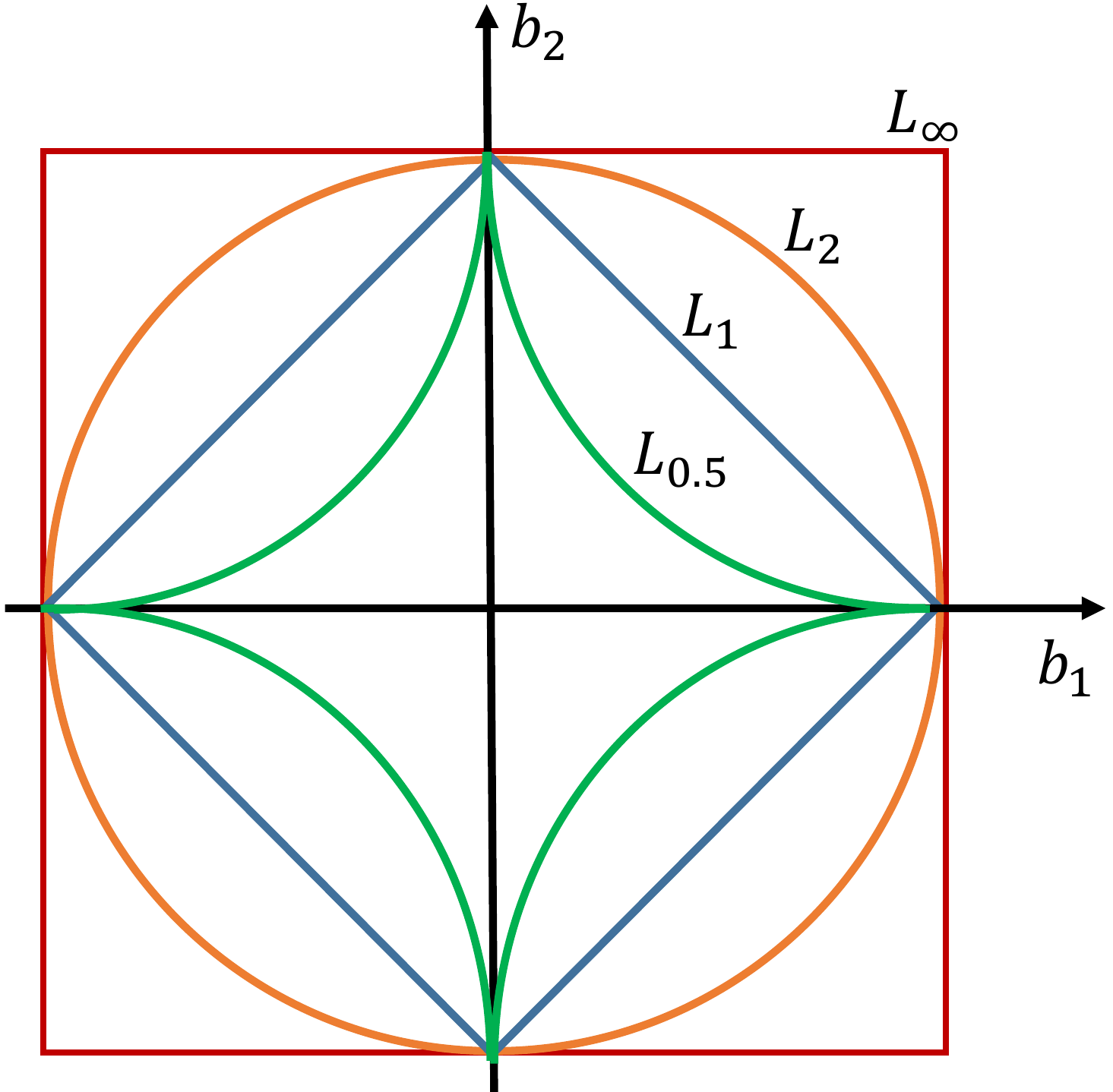}}
	\subfloat[\label{figpenreg:fig1}$\mathcal{L}_1$ penality (lasso)]
	{\includegraphics[width=0.25\paperwidth]{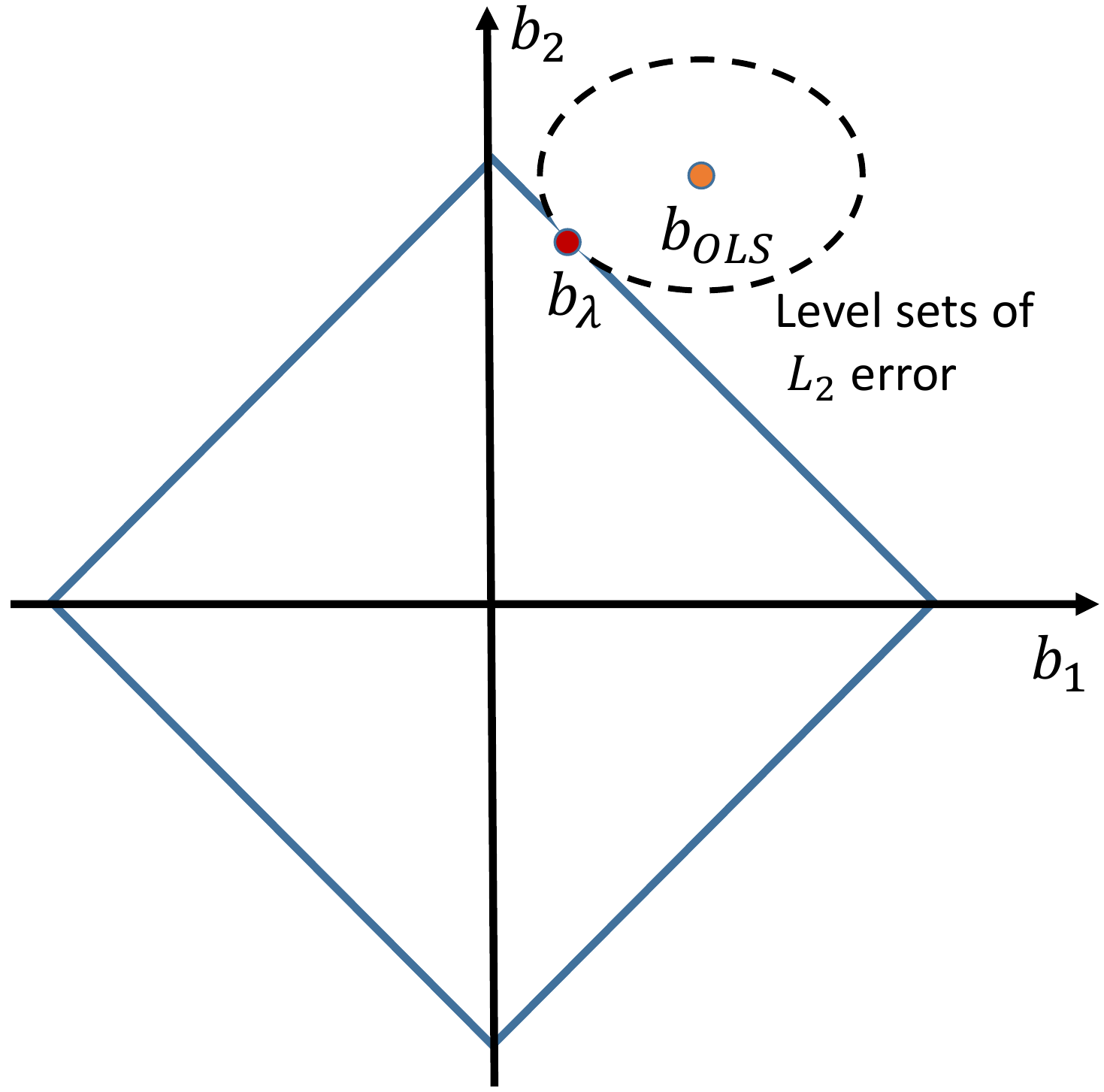}}
	\subfloat[\label{figpenreg:fig2}$\mathcal{L}_{0.5}$ penality]
	{\includegraphics[width=0.25\paperwidth]{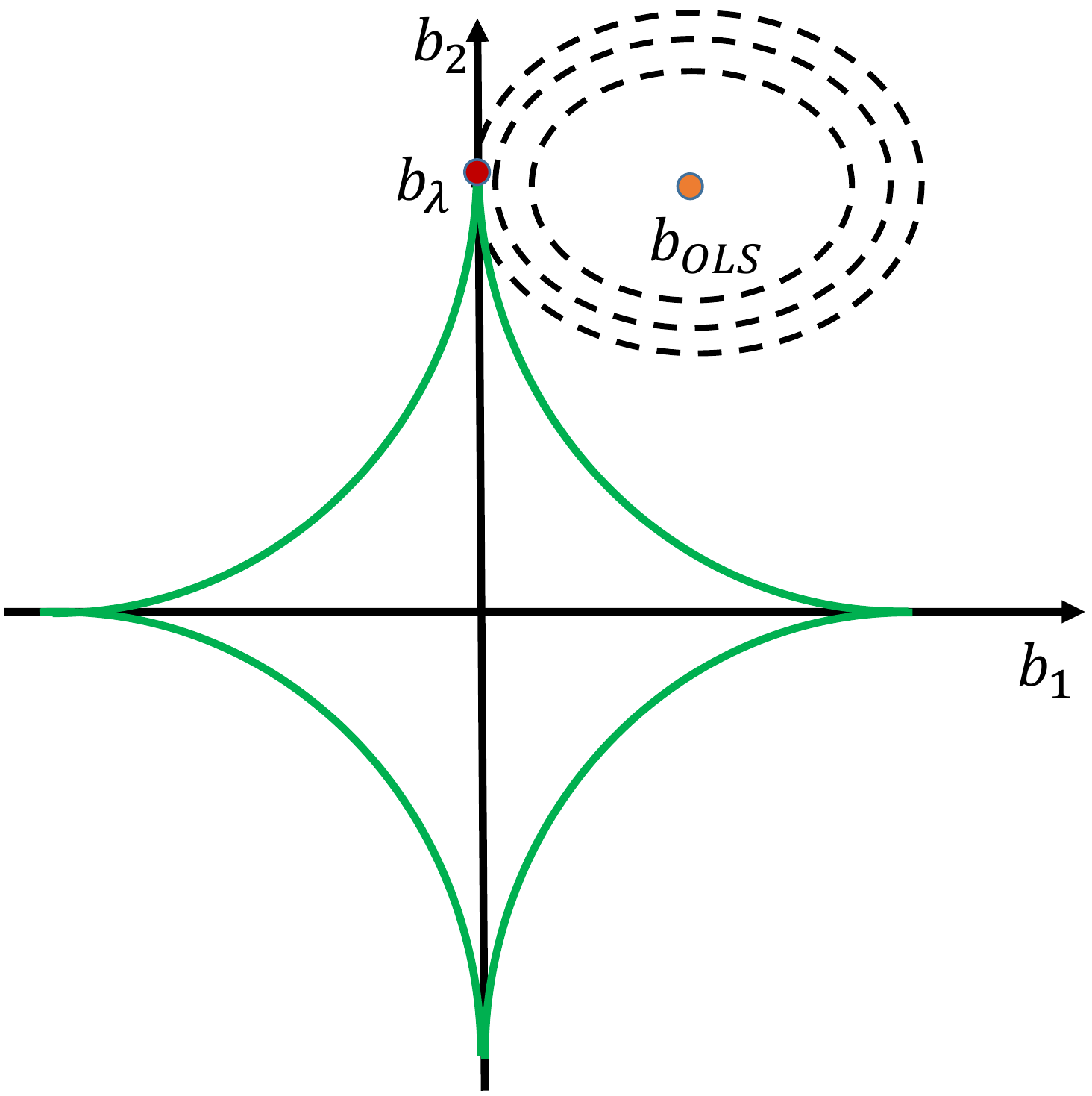}}

	\caption{\label{figpenreg}Illustration of various penalized regressions}
\end{figure}

The idea behind penalized regression is illustrated in Figure~\ref{figpenreg} where $b_{\lambda}$ refers to the penalized regression estimates for some $\lambda$ and $b_{OLS}$ refers to the OLS estimates. As shown in Figure~\ref{figpenreg:fig3}, different $\mathcal{L}_{\gamma}$ norms correspond to different boundaries for the estimation feasible set. For the $\mathcal{L}_1$ penalized regression (lasso), the feasible set is a diamond since each coefficient is equally penalized by the $\mathcal{L}_1$ norm. The feasible area shrinks under a $\mathcal{L}_{0.5}$ penalty. Hence, as shown in Figure~\ref{figpenreg:fig3}, given the same $\lambda$, the smaller is $\gamma$, the more likely $b_{\lambda}$ is to be a corner solution. Hence, given the same $\lambda$, under the $\mathcal{L}_{0.5}$ penalty variables are more likely to be dropped than with the $\mathcal{L}_{1}$ or $\mathcal{L}_{2}$ penalty.\footnote{For $0<\gamma<1$, the penalized regression may be a non-convex programming problem. While general algorithms have not been found for non-convex optimization, \citet{strongin2013global}, \citet{yan2001global} and \citet{noor2008differentiable} have developed functioning algorithms. For $\gamma=0$, the penalized regression becomes a discrete programming problem, which can be solved by Dantzig-type methods (see \citet{candestao07}).} In special cases when $\gamma=0$ and $\lambda$ is fixed at $2$ ($\ln n_t$), the $\mathcal{L}_0$ penalized regression is identical to the Akaike (Bayesian) information criterion.

The last important comment is that penalized regression primarily focuses on overfitting. By contrast, OLS minimizes the eTE without any penalty, typically causing a large eGE (as shown in Figure~\ref{fig:VCineqov}). There is also the possibility that OLS predicts the data poorly, causing both the eTE and eGE to be large. The latter refers to underfitting and is shown in Figure~\ref{fig:VCinequn}. We are more capable of dealing with overfitting than underfitting despite the fact that it possible to quantify GA or the eGE.\footnote{See eq.~(\ref{eq1:thm3.1}) and (\ref{eq2:thm3.1}).} Typically overfitting in OLS is caused by including too many variables, which we can resolve by reducing $p$. However, underfitting in OLS is typically due to a lack of data (variables) and the only remedy is to collect additional relevant variables.

\subsection{Schematics and assumptions for eGE minimization with penalized regression}

As shown in Section~2, eGE minimization improves finite-sample GA, implying the estimator has a lower eGE on out-of-sample data. In this section, we implement the schematics of eGE minimization on penalized regression. We demonstrate: (1) specific error bounds for any penalized regression, (2) a general $\mathcal{L}_2$ consistency property for penalized regression estimates, (3) that the upper bound for the $\mathcal{L}_2$ difference between $b_{pen}$ and $b_{OLS}$ is a function of the eGE, the tail property of the loss distribution and sample exogeneity.

\begin{figure}
	\centering
	\includegraphics[width=0.5\paperwidth]{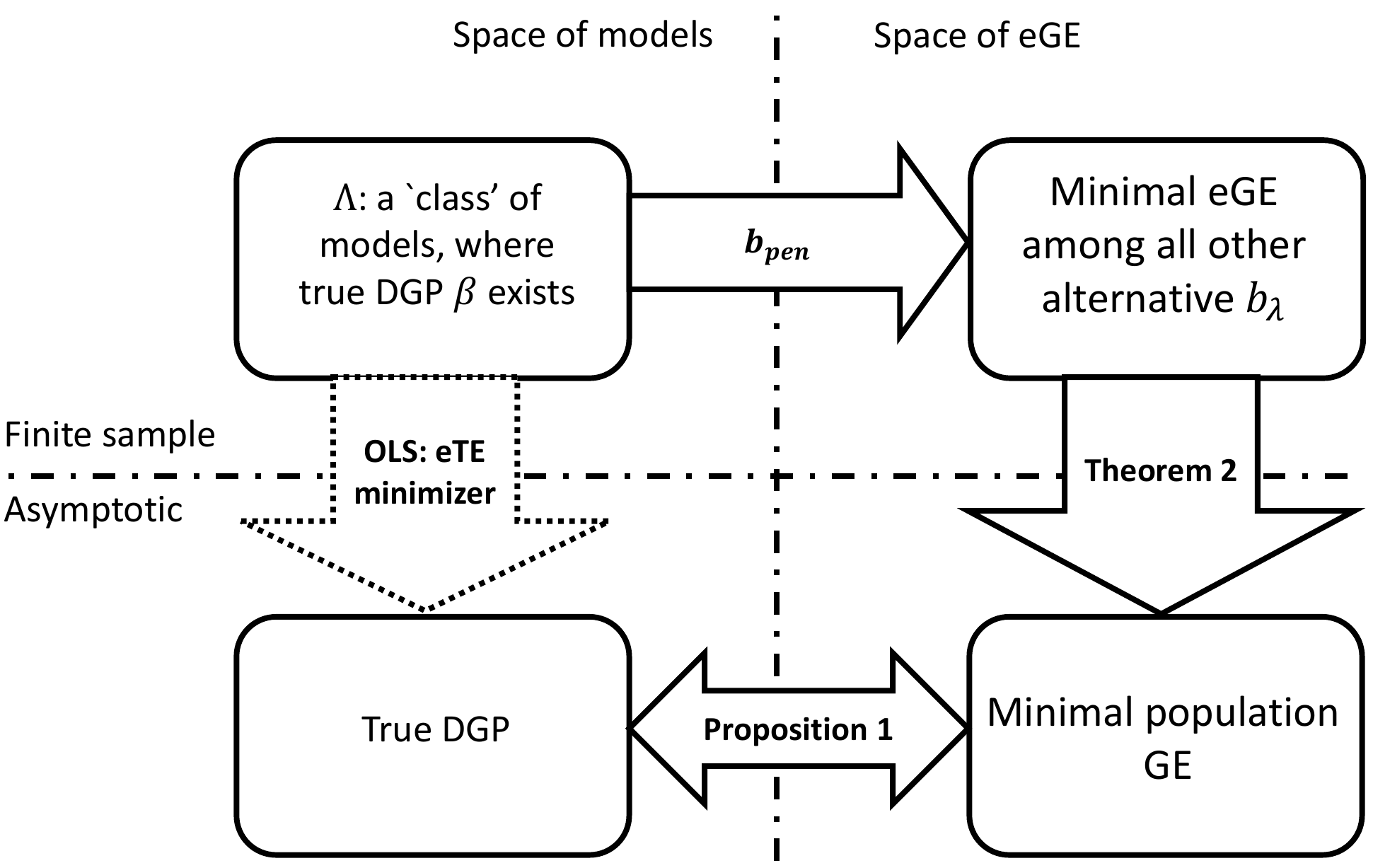}

	\caption{\label{fig:outline}Outline of proof strategy}
\end{figure}

The classic route to derive asymptotic or finite-sample properties for regression is through analyzing the properties of the estimate in the space of the eTE. In contrast, to study how penalized regression improves GA or eGE and balances the in-sample and out-of-sample fit, we reformulate the asymptotic and finite-sample problems in the space of the eGE. We show that, under the framework of eGE minimization, a number of finite-sample properties of penalized regression can be explained by eGE or the finite-sample GA.

In asymptotic analysis, consistency is typically considered to be one of the most fundamental properties. To ensure that eGE minimization is a reliable estimation approach, we prove that the penalized regression, which is a specific form of the eGE minimizer, converges to the true DGP as $n\rightarrow\infty$. Essentially, we show that penalized regression bijectively maps $b_{pen}$ to the minimal eGE among $\{b_{\lambda}\}$ on the test set. To bridge between the finite sample and asymptotic results we need to show that if

\begin{itemize}
    \item   the true DGP $\beta$ is bijectively assigned to the minimal eGE in population, and

	\item  $\min_{b\in b_{\lambda}}
	       \frac{1}{n_s}\sum_{i=1}^{n_s}\Vert Y_{s}-X_{s}b\Vert_{2}^{2}
	       \rightarrow
	       \min_b \int\Vert y-\mathbf{x}^{T}b\Vert_{2}^{2}~\mathrm{d}F(y, \mathbf{x})$,
\end{itemize}
then $b_{pen}$ is consistent in probability or $\mathcal{L}_2$, or
        \[
	       b_{pen} =
	       \argmin\{\mbox{eGEs of } \{b_{\lambda}\}\}
	       \overset{\mathbf{P}~\mathrm{or}~\mathcal{L}_2}{\rightarrow}
	       \argmin_b \int\Vert y_{s}-\mathbf{x}_{s}^{T}b\Vert_{2}^{2}~\mathrm{d}F(y, \mathbf{x})
	       = \beta.
        \]

At the outset, we stress that each variable in $(Y,X)$ must be standardized before implementing penalized regression. Without standardization, as shown by \citep{tibshirani96}, the penalized regression may be influenced by the magnitude (units) of the variables. After standardization, of course, $X$ and $Y$ are unit- and scale-free.

To ensure the consistency of penalized regression, we require the following three additional assumptions.

\bigskip
\noindent
\textbf{Further assumptions}

\begin{enumerate}

    \item[\textbf{A4.}]    The true DGP is $Y = X\beta + u$.

	\item[\textbf{A5.}]    $\mathbb{E} \left( u^T X \right) = \mathbf{0}$.

	\item[\textbf{A6.}]    No perfect collinearity in $X$.

\end{enumerate}
The assumptions \textbf{A4} to \textbf{A6} restrict the true DGP $\beta$ to be identifiable. Otherwise, there might exist another model that is not statistically different from the true DGP. The assumptions are quite standard for linear regression.

\subsection{Necessary propositions for final results}

Under assumptions \textbf{A1} to \textbf{A6}, we show that the true DGP is the most generalizable model, yielding Proposition~3.1.
%
%
\begin{prop}[Identification of $\beta$ in the space of eGE]
Under assumptions \textbf{A1} to \textbf{A6}, the true DGP, $Y = X\beta + u$, is the one and only one offering the minimal eGE as $\widetilde{n}\rightarrow\infty$.

\label{prop3.1}
\end{prop}
Proposition~\ref{prop3.1} states that there is a bijective mapping between $\beta$ and the global minimum eGE in the population. If \textbf{A5} or \textbf{A6} are violated, there may exist variables in the sample that render the true DGP not to be the model with minimum eGE in population. As shown in Algorithm~1,  penalized regression picks the model with the minimum eGE in $\{b_\lambda\}$ to be $b_{pen}$. As a result, we also need to prove that, when the sample size is large enough, the true DGP is included in $\{b_{\lambda}\}$, the list of models from which validation or cross-validation selects. This is shown in Proposition~3.2.
%
%
\begin{prop}[Existence of $\mathcal{L}_2$ consistency]
Under assumptions \textbf{A1} to \textbf{A6} and Proposition~\ref{prop3.1}, there exists at least one $\widetilde{\lambda}$ such that
$\lim_{\widetilde{n} \rightarrow \infty} \Vert  b_{\widetilde{\lambda}} - \beta \Vert_2 = 0$.

\label{prop3.2}
\end{prop}

\begin{figure}[ht]

	\centering

	\subfloat[\label{figlassoshrink:fig1}under-shrinkage]
	{\includegraphics[width=0.25\paperwidth]{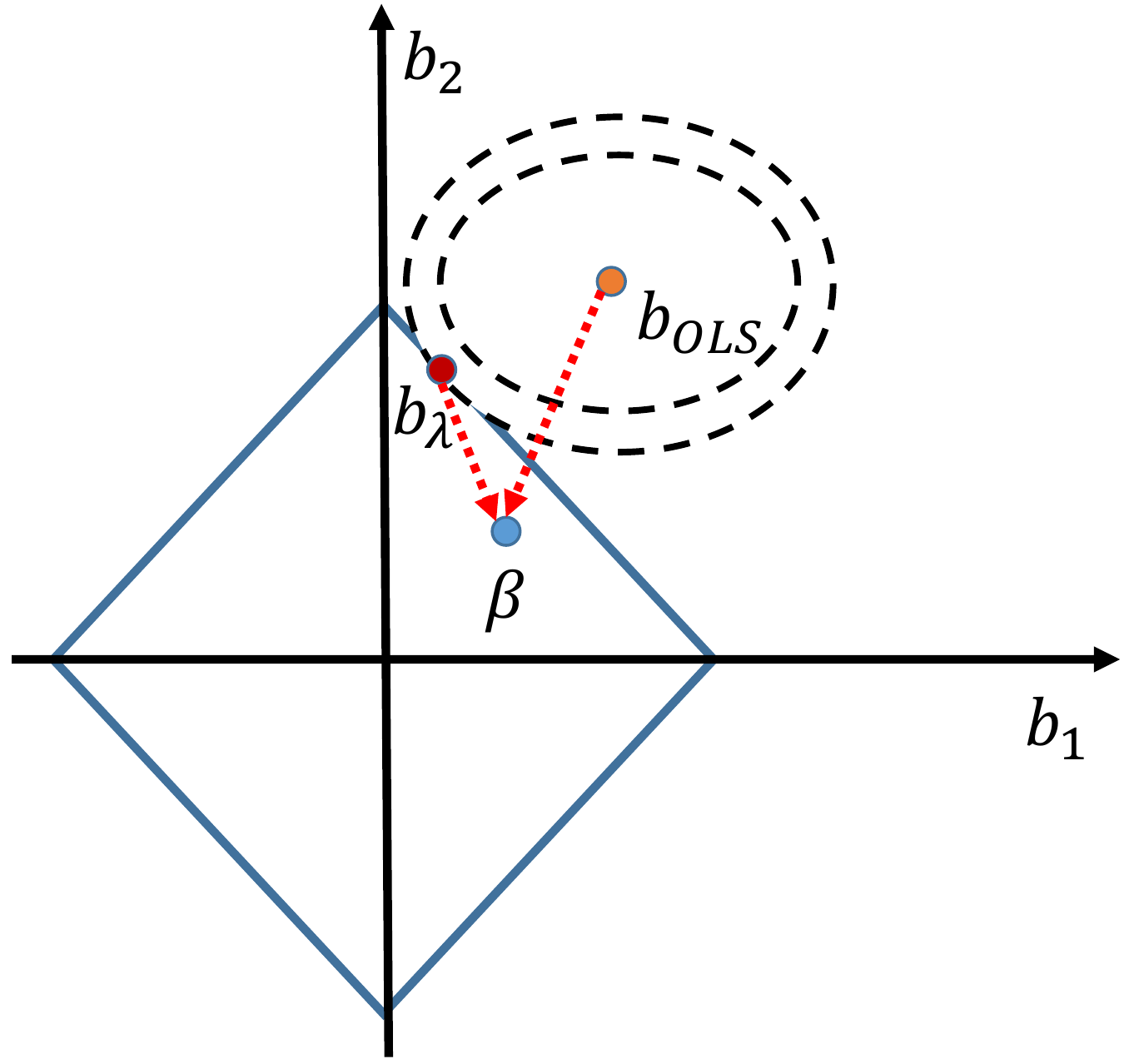}}
	\subfloat[\label{figlassoshrink:fig2}perfect shrinkage]
	{\includegraphics[width=0.25\paperwidth]{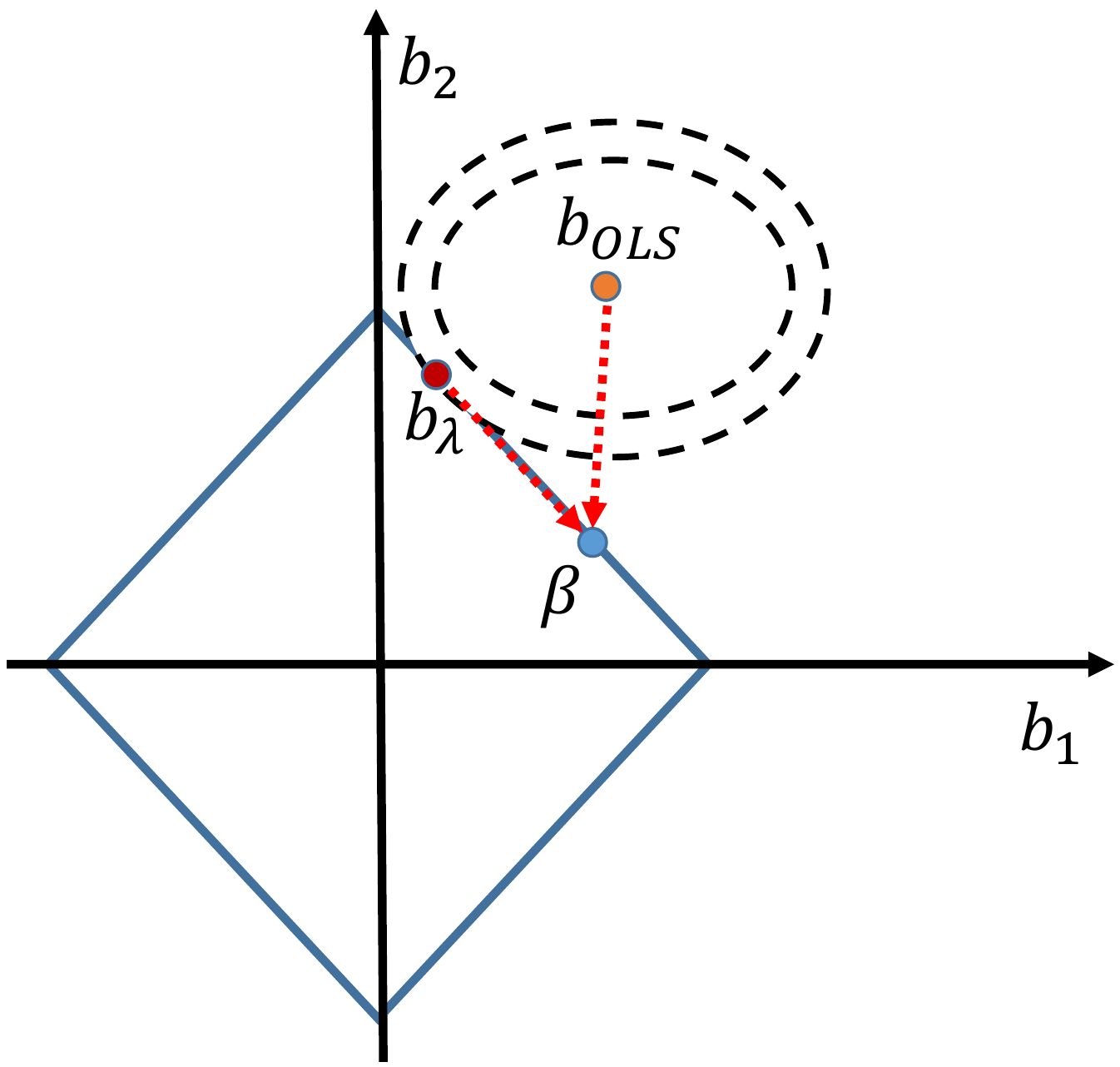}}
	\subfloat[\label{figlassoshrink:fig3}over-shrinkage]
	{\includegraphics[width=0.25\paperwidth]{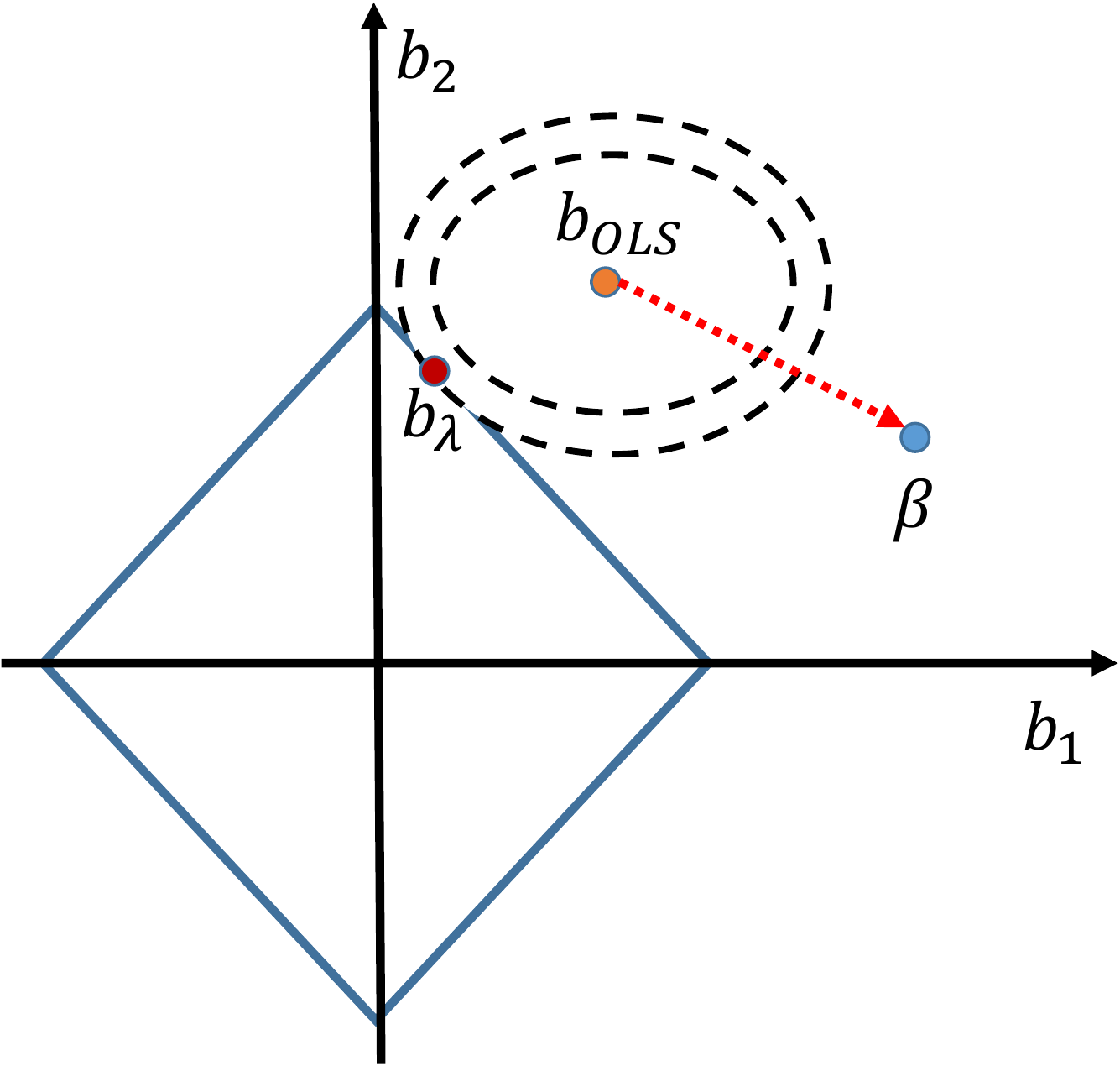}}

	\caption{\label{figlassotypeshrink:fig}Various types of shrinkage under an $\mathcal{L}_1$ penalty}

\end{figure}

Since the penalized regression can be sketched as a constrained minimization of the empirical error, we can illustrate Proposition~\ref{prop3.1} and~\ref{prop3.2} in Figure~\ref{figlassotypeshrink:fig} using lasso as the example of a penalized regression. In Figure~\ref{figlassotypeshrink:fig}, the parallelogram-shaped feasible sets are determined by the $\mathcal{L}_\gamma$ penalty, $b_{\lambda}$ refers to the solution of eq.~(\ref{lasso-type}), $\beta$ refers to the true DGP, and $b_{OLS}$ refers to the OLS estimates. Different values for $\lambda$ imply different areas for the feasible set of the constrained minimization; the area of the feasible set gets smaller as the value of $\lambda$ gets larger. Hence, one of three cases may occur: (i) as shown in Figure~\ref{figlassoshrink:fig1}, for a small value of $\lambda$, $\beta$ lies in the feasible set (under-shrinkage) and offers the minimum eTE in the population; (ii) as shown in Figure~\ref{figlassoshrink:fig2}, for the oracle $\lambda$, $\beta$ is located precisely on the boundary of the feasible set (perfect-shrinkage) and still offers the minimum eTE in the population; (iii) as shown in Figure~\ref{figlassoshrink:fig3}, for a large value of $\lambda$, $\beta$ lies outside the feasible set (over-shrinkage). In cases (i) and (ii), the constraints become inactive as $\widetilde{n}\rightarrow\infty$, so $\lim_{\widetilde{n}\rightarrow\infty}b_{\lambda}=\lim_{\widetilde{n}\rightarrow\infty}b_{OLS}=\beta$. However, in case (iii), $\lim_{\widetilde{n}\rightarrow\infty}b_{\lambda}\neq\beta$. Therefore, tuning the penalty parameter $\lambda$ is critical for the theoretical properties of penalized regression estimates.

\subsection{Main results for penalized regression estimates}

As shown above, intuitively the penalized regression estimate is expected be consistent in some norm or measure as long as we can be sure that for a specific $\lambda$, $\beta$ lies in the feasible set and offers the minimum eTE in population. However, in practice we may not know a priori which $\lambda$ causes over-shrinkage and which does not, especially when the number of variables, $p$, is not fixed. As a result, we need to a method such as cross-validation or validation to tune the value of $\lambda$. Thus, as a direct application of eGE minimization in Section~2, we use GA/eGE analysis to show that eGE minimization guarantees the model selected by penalized regression, $b_{pen}$, asymptotically converges in $\mathcal{L}_{2}$ to the true DGP.

In the following section we analyze the finite-sample and asymptotic properties of the penalized regression estimate in two scenarios: $n\geqslant p$ and $n < p$. In the case where $n\geqslant p$, OLS is feasible, so we take the OLS estimate for the unpenalized regression estimate. However, when $n<p$, OLS is not feasible, and we use forward stagewise regression (FSR) for the unpenalized regression estimate. Hereafter, $b_{OLS}$ is the OLS estimate from the training set.

\subsubsection{Case: $n\geqslant p$}

Firstly, by specifying eq.~(\ref{eq:thm2.1}) and (\ref{eq:thm2.2}) in the context of regression, we can establish the upper bound of the eGE, as shown in Lemma~\ref{lem3.1}.
%
%
\begin{lem}[Upper bound of the eGE for the OLS estimate]
Under \textbf{A1} to \textbf{A6}, if we assume $u \sim Gaussian(0,var(u))$,

    \begin{enumerate}

    \item (\textbf{Validation}) The following bound for the eGE holds with probability at
	    least $\varpi(1-1/n_t)$ for $b_{OLS}$, $\forall\varpi\in\left(0,1\right)$.
		\begin{equation}
			\frac{1}{n_s}(\Vert e_s \Vert_2)^2
			\leqslant \frac{(\Vert e_t \Vert_2)^2}{n_t(1-\sqrt{\epsilon})}
			+ \frac{2 (\mathrm{var}(u))^2}{n_s\sqrt{1-\varpi}},
			\label{eq1:lem3.1}
		\end{equation}
        where $e_s$ is the OLS eGE, $e_t$ is the OLS eTE, and $\epsilon$ is defined in Lemma~\ref{lem2.1}.

    \item (\textbf{$K$-fold cross-validation}) The following bound for the eGE holds with probability at least $\varpi(1-1/K)$ for $b_{OLS}$, $\forall\varpi\in\left(0,1\right)$.
		\begin{equation}
			\frac{1}{K}\sum^{K}_{j=1}\frac{(\Vert e_s^j \Vert_2)^2}{n/K}
			\leqslant
			\frac{n(K-1)\sum_{q=1}^K (\Vert e_t^q \Vert_2)^2}
            {K^2(1-\sqrt{\epsilon})} + \frac{2 (\mathrm{var}(u))^2 }{ \sqrt{1-\varpi}\cdot (n/K)^2},
			\label{eq2:lem3.1}
		\end{equation}
        where $e_s^j$ is the eGE of OLS estimate in the $j$th round, $ e_t^q$ is the eTE of OLS in the $q$th round, while $\epsilon$ and $\varsigma$ are defined in Lemma~\ref{lem2.1}.

   \end{enumerate}    	

\label{lem3.1}	
\end{lem}

Eq.~(\ref{eq1:lem3.1}) and (\ref{eq2:lem3.1}) show that the higher the variance of $u$ in true DGP, the higher the upper bound of the eGE in validation and cross-validation. Based on eq.~(\ref{eq2:lem3.1}), the lowest upper bound of the cross-validation eGE is determined by minimizing the expectation of the RHS of eq.~(\ref{eq2:lem3.1}), yielding the way to find the optimal $K$ as follows.
%
%
\begin{cor}[The optimal $K$ for penalized regression]
Under \textbf{A1} to \textbf{A6} and eq.~(\ref{eq2:lem3.1}) from Lemma~\ref{lem3.1}, if we assume the OLS error term $u \sim Gaussian(0, \mathrm{var}(u))$, the optimal $K$ for cross-validation in regression (the minimum expected upper bound of eGE) is defined:
	\[
        K^* = \argmin_{K} \frac{\mathrm{var}(u)}{1-\sqrt{\epsilon}} +
                          \frac{4(\mathrm{var}(u))^2}{\sqrt{1-\varpi}(n/K)^2}
	\]

\label{cor3.1}
\end{cor}

Similar to eq.~(\ref{eq:thm2.1}), eq.~(\ref{eq1:lem3.1}) and (\ref{eq2:lem3.1}) respectively measure the upper bound of the eGE for the OLS estimate using validation and cross-validation. In standard practice, neither validation nor cross-validation are implemented as part of OLS estimation and hence the eGE of the OLS estimate is rarely computed. Eq.~(\ref{eq1:lem3.1}) and (\ref{eq2:lem3.1}) show that eGE can be computed without having to carry out validation or cross-validation.

In penalized regression, the penalty parameter $\lambda$ can be tuned by validation or $K$-fold cross-validation. For $K\geqslant 2$, we have $K$ different test sets for tuning $\lambda$ and $K$ different training sets for estimation. Based on eq.~(\ref{eq1:lem3.1}) and (\ref{eq2:lem3.1}), an upper bound for the $\mathcal{L}_2$ predicted difference between $b_{OLS}$ and $b_{pen}$ can be established under validation and cross-validation, as shown below.
%
%
\begin{prop}[$\mathcal{L}_2$ distance between the penalized and unpenalized predicted values]
Under \textbf{A1} to \textbf{A6} and based on Lemma~\ref{lem3.1}, Proposition~\ref{prop3.1}, and Proposition~\ref{prop3.2},

	\begin{enumerate}

        \item (\textbf{Validation}) The following bound holds for the validated $b_{pen}$ with probability $\varpi(1-1/n_{t})$
			\begin{equation}
				\frac{1}{n_{s}}(\Vert X_{s}b_{OLS}-Xb_{pen}\Vert _{2})^{2}
				\leqslant
				\left(\frac{1}{n_{t}}\frac{\Vert e_{t}\Vert_{2}^{2}}{1-\sqrt{\epsilon}}
				-\frac{1}{n_{s}}\Vert e_{s}\Vert _{2}^{2}\right)
				+\frac{4}{n_{s}}\Vert e_{s}^{T}X_{s}\Vert_{\infty}\Vert b_{OLS}\Vert _{1}+\varsigma
				\label{eq1:prop3.3}
			\end{equation}
            where $\varsigma$ is defined in Lemma~\ref{lem3.1}.

        \item (\textbf{$K$-fold cross validation}) The following bound holds for the $K$-fold cross-validated $b_{pen}$ with probability $\thickbar{\varpi}(1-1/n_{t})$
		      \begin{eqnarray}
                \frac{1}{K}\sum_{q=1}^K
                \frac{1}{n_s}(\Vert X_s^q b_{OLS}^q -X_s^q b_{pen}^q \Vert_{2})^{2}
		      	& \leqslant &
                \left\vert\frac{1}{n_t}\frac{\frac{1}{K}\sum_{q=1}^K
                \left\Vert e_t^q \right\Vert_{2}^{2}}{1-\sqrt{ \epsilon}} -
		      	\frac{1}{K}\sum_{q=1}^K\frac{1}{n_s}
                \left\Vert e_s^q \right\Vert_{2}^{2}\right\vert \\ \notag
		      	& + &
                \frac{1}{K}\sum_{q=1}^K
                \frac{4}{n_s}\left\Vert\left( e_s^q \right)^TX_s^q\right\Vert_{\infty}		      	\left\Vert b_{OLS}^q\right\Vert_{1} +
                \varsigma_{cv}.
		      	\label{eq2:prop3.3}
		      \end{eqnarray}
            where $e_t^q$ is the eTE of the OLS estimate in the $q$th round of cross-validation, $e_s^q$ is the eGE of the OLS estimate in the $q$ round of cross-validation and $b_{OLS}^q$ is the OLS estimate in the $q$th round of cross-validation.

	\end{enumerate}

\label{prop3.3}
\end{prop}

We can now derive the upper bound of $\Vert b_{OLS} - b_{pen}\Vert_2$, as shown in Theorem~\ref{thm3.1}.
%
%
\begin{thm}[$\mathcal{L}_2$ distance between the penalized and unpenalized estimates]
Under \textbf{A1} to \textbf{A6} and based on Propositions~\ref{prop3.1}, \ref{prop3.2}, and \ref{prop3.3},

	\begin{enumerate}

		\item (\textbf{Validation}) The following bound holds with probability $\varpi(1-1/n_{t})$:
		      \begin{equation}
		      	\Vert b_{OLS}-b_{pen}\Vert _{2} \leqslant
		      	\sqrt{\left\vert\frac{1}{\rho n_{t}}\frac{\Vert e_{t}\Vert_{2}^{2}}{(1-\sqrt{\epsilon})}
		      		-\frac{1}{\rho n_{s}}\Vert e_{s}\Vert_{2}^{2}\right\vert}
		      	+\sqrt{\frac{4}{\rho n_{s}}\Vert e_{s}^{T}X_{s}\Vert_{\infty}\Vert b_{OLS}\Vert_{1}}
		      	+ \left(\frac{\varsigma}{\rho}\right)^{\frac{1}{2}}
		      	\label{eq1:thm3.1}
		      \end{equation}
              where $\rho$ is the minimal eigenvalue of $X^{T}X$ and $\varsigma$ is defined in Lemma~\ref{lem3.1}.

        \item (\textbf{$K$-fold cross validation}) The following bound holds with probability $\varpi(1-1/n_{t})$:
		      \begin{eqnarray}
		      	\frac{1}{K}\sum_{q=1}^K (\Vert b_{OLS}^q - b_{pen}^q \Vert _{2})^{2} &
		      	\leqslant	      	
		      	\left\vert \frac{1}{K}\sum_{q=1}^K \frac{1}{n_t\cdot\overline{\rho}}
		      	\frac{ \left\Vert e_t^q \right\Vert_{2}^{2}}{1-\sqrt{\epsilon}}
		      	-\frac{1}{K}\sum_{q=1}^K \frac{1}{n_s \cdot \bar{\rho} }
		      	\left\Vert e_s^q \right\Vert_{2}^{2} \right\vert
		      	\notag \\
		      	& +\frac{1}{K}\sum_{q=1}^K \frac{4}{ n_s \cdot \bar{\rho} }
		      	\left\Vert (e_s^q)^TX_s^q \right\Vert_{\infty}
		      	\left\Vert b_{OLS}^q \right\Vert_{1}
		      	+\frac{\varsigma}{\bar{\rho}}
		      	\label{eq2:thm3.1}
		      \end{eqnarray}
              where $\bar{\rho}$ is defined as $\min\left\{\rho_q\vert\rho_q \mbox{ is the minimal eigenvalue of} \left(X^q_s\right)^{T}X^q_s, \forall q\right\}$.

    \end{enumerate}

\label{thm3.1}
\end{thm}

Some important remarks apply to Theorem~\ref{thm3.1}. The LHS of eq.~(\ref{eq1:thm3.1}) essentially measures how much the penalized regression estimate differs from the OLS estimate under validation. The RHS of eq.~(\ref{eq1:thm3.1}) essentially captures the maximum $\mathcal{L}_2$ difference between $b_{OLS}$ and $b_{pen}$. As shown in eq.~(\ref{eq1:thm3.1}), the maximum difference depends on the GA of the true DGP and the GA of the OLS model in several different forms.

\begin{itemize}

    \item   The first term on the RHS of eq.~(\ref{eq1:thm3.1}) (ignoring $1/\rho$) is the difference between the eGE from OLS and the upper bound of the population error for the OLS estimate, or, equivalently, \emph{how far the GA of the OLS estimate is from its maximum}. The larger the GA of $b_{OLS}$, the less overfitting OLS generates, the closer the eGE of $b_{OLS}$ is to the upper bound of the population error, and the smaller the first term on the RHS of eq.~(\ref{eq1:thm3.1}).

    \item   The second term on the RHS of eq.~(\ref{eq1:thm3.1}) (ignoring $4/\rho$) measures the empirical endogeneity of the error term of the OLS estimate on the test set. On the training set $e_t^T X_s=\mathbf{0}$, but in general $e_s^T X_s\neq 0$ on the test set. Hence, $\frac{1}{n_{s}}\Vert e_{s}^{T}X_{s}\Vert_{\infty}\Vert b_{OLS}\Vert_{1}$ \emph{measures the GA for the empirical moment condition of the OLS estimate on out-of-sample data}.\footnote{Because we standardize the test and training data, the moment condition $\mathbb{E}(e_s)=0$ holds directly.} The more generalizable the OLS estimate, the closer $e_s^T X_s$ is to zero on out-of-sample data, and the smaller the second term on the RHS of eq.~(\ref{eq1:thm3.1}).

    \item   The third term on the RHS of eq.~(\ref{eq1:thm3.1}) is affected by $\varsigma$, which \emph{measures the heaviness of the tail in the loss distribution for the OLS estimate}. Similar to the comments of Theorem~\ref{thm2.1}, the OLS loss distribution affects the GA of the OLS estimate. The heavier the loss tail, the more volatile the eGE on out-of-sample data, and the more difficult it is to bound the eGE for OLS.

    \item   All three RHS terms in eq.~(\ref{eq1:thm3.1}) are affected by $\rho$, the minimum eigenvalue of $X_x^TX_s$, which can be thought of as a \emph{measure of the curvature of the objective function for penalized regression}. The larger the minimum eigenvalue, the more convex the objective function. Put another way, it is easier to identify the true DGP $\beta$ from the alternatives as $n$ get larger.

\end{itemize}

The interpretation of eq.~(\ref{eq2:thm3.1}) is similar to eq.~(\ref{eq1:thm3.1}) adjusting for cross-validation. Hence, the first term on the RHS of eq.~(\ref{eq2:thm3.1}) (ignoring $1/\overline{\rho}$) stands for how far away the \textit{average} GA of OLS estimate is from its maximum in $K$ rounds of validation. The second term on the RHS of eq.~(\ref{eq2:thm3.1}) (ignoring $4/\overline{\rho}$) indicates \textit{on average} how generalizable the empirical moment condition of the OLS estimate is with out-of-sample data in $K$ rounds of validation. Similarly, $\overline{\varsigma}$ indicates \textit{on average} the heaviness in the tail of the loss distribution in $K$ rounds of validation. As a direct result of Theorem~\ref{thm3.1}, the $\mathcal{L}_2$ consistency for the penalized regression estimate is established as follows.
%
%
\begin{cor}[$\mathcal{L}_2$ consistency for the penalized regression estimate when $n\geqslant p$]
Under \textbf{A1} to \textbf{A6} and Propositions~\ref{prop3.1}, \ref{prop3.2}, and \ref{prop3.3}, if we assume the error term $u \sim Gaussian(0, \mathrm{var}(u))$, then $b_{pen}$ converges in the $\mathcal{L}_{2}$ norm to the true DGP if \textup{$\lim_{n\rightarrow\infty}p/\widetilde{n}=0$}

\label{cor3.2}
\end{cor}

\begin{figure}
	\centering
	\subfloat[\label{figillustrate:fig1}$\mathcal{L}_{1}$ penalty]
	{\includegraphics[width=0.35\paperwidth]{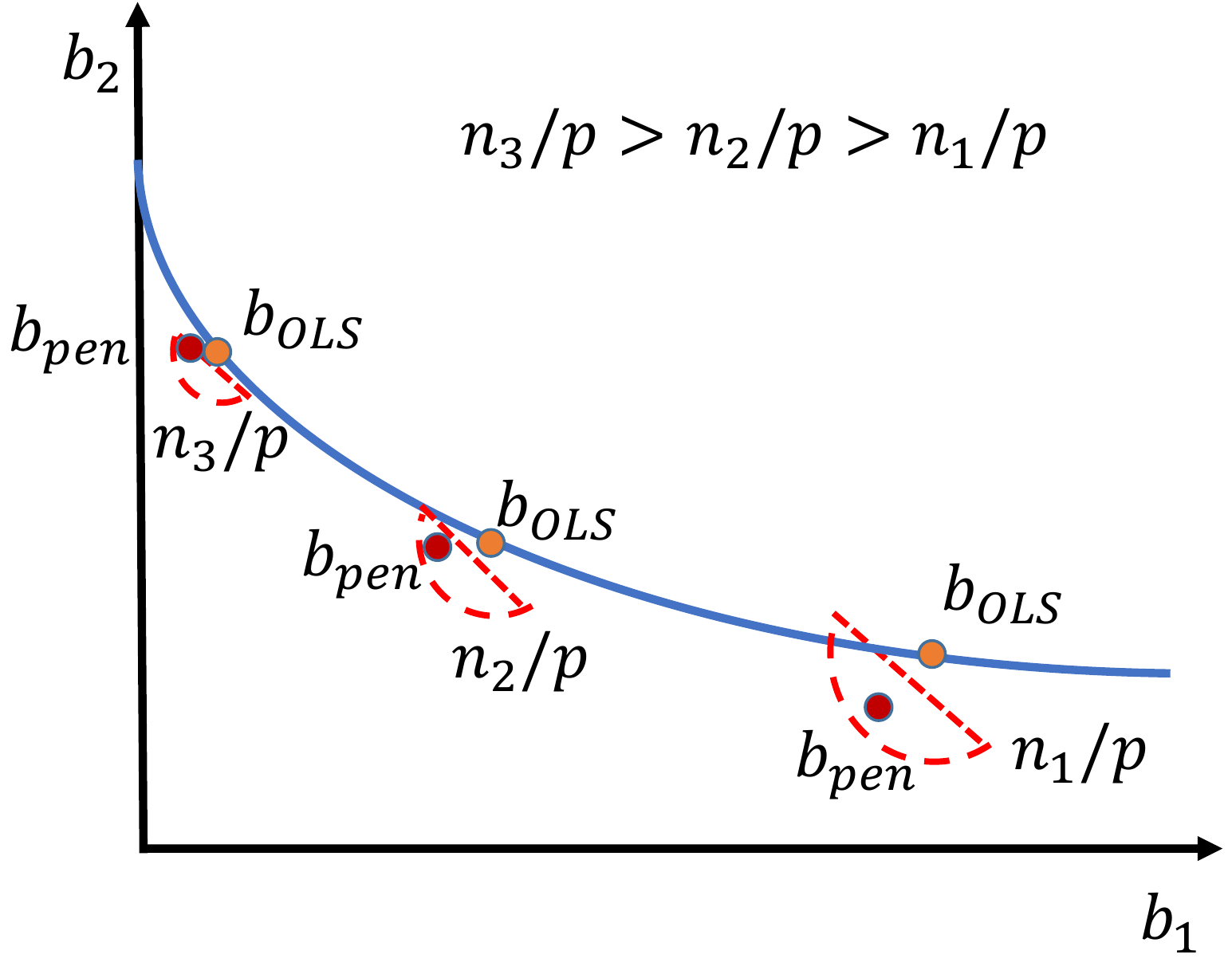}}
	\subfloat[\label{figillustrate:fig2}$\mathcal{L}_{0.5}$ penalty]
	{\includegraphics[width=0.35\paperwidth]{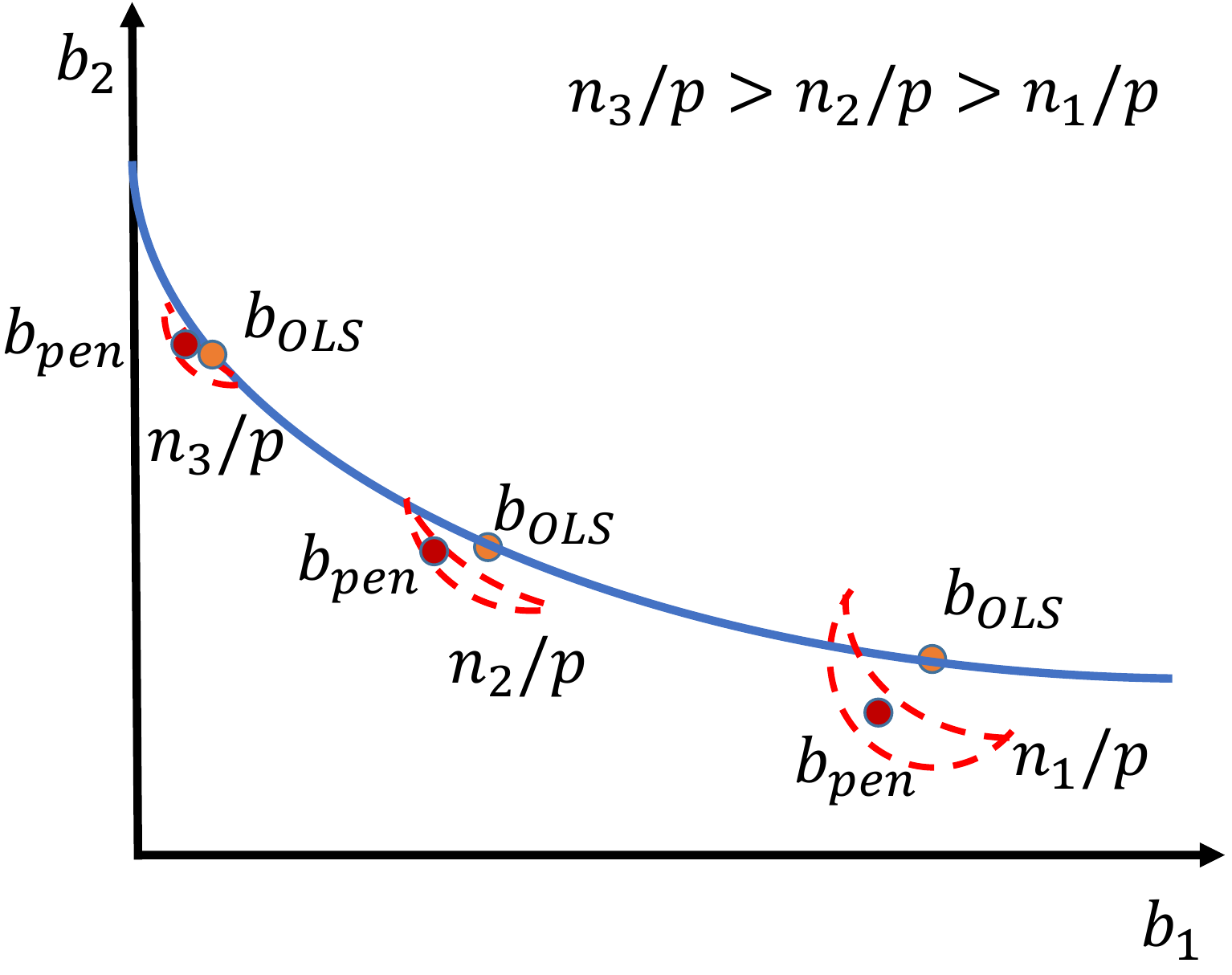}}

	\caption{\label{figillustrate}Illustration of the relation between $b_{OLS}$ and $b_{pen}$}
\end{figure}

Theorem~\ref{thm3.1} and Corollary~\ref{cor3.2} are illustrated in Figure~\ref{figillustrate}. Typically, due to the poor GA of the OLS estimate, the penalized regression estimate $b_{pen}$ will not lie on the same convergence path as the OLS estimate. However, Theorem~\ref{thm3.1} shows that the deviation of $b_{pen}$ from the convergence path is bounded. Furthermore, $b_{pen}$ typically lies within an $\epsilon$-ball centered on $b_{OLS}$ whose radius is a function of the eGEs of the OLS estimate and the true DGP. Also, as shown in Figures~\ref{figlassoshrink:fig1} and \ref{figlassoshrink:fig2}, $b_{pen}$ always lies within the feasible set parameterized by $\lambda\Vert b\Vert_{\gamma}$. Hence, as shown in Figures~\ref{figillustrate:fig1} and \ref{figillustrate:fig2}, $b_{pen}$ typically is located in the small area at the intersection of the $\mathcal{L}_{\gamma}$ feasible area and $\epsilon$-ball. Unless the optimal $\lambda$ from validation or cross-validation is $0$, the OLS estimate will never be in the feasible area of the penalized regression estimate, which is why the intersection region is always below the $b_{OLS}$. As $n/p$ increases, the $\epsilon$-ball becomes smaller, the penalized regression estimate gets closer to the OLS estimate, and both converge to $\beta$.

\subsubsection{Case: $n < p$}

Typically, to ensure $b_{OLS}$ can identify the true DGP $\beta$, we require that $(\Vert e\Vert_2)^2$ is strongly convex, or that the minimal eigenvalue of $X^T X$, $\rho$, is strictly larger than 0. However, if $p>n$, $\rho=0$ and the space of $(\Vert e\Vert_2)^2$ is flat in some direction. As a result, the $\Vert b_{OLS}\Vert_1$ is not closed-form, the true DGP $\beta$ cannot be identified and Eqs.~(\ref{eq1:thm3.1}) and (\ref{eq2:thm3.1}) are trivial.

To make the results above non-trivial, we need to ensure $\beta$ is still identifiable even when $p>n$. Put another way, we need to ensure that the strong convexity of the space $(\Vert e\Vert_2)^2$ is still reserved for the $p>n$ case and $\beta$ is still identifiable. This is guaranteed by the sparse eigenvalue condition \citep{bickeletal09, meinshausenyu09, zhang10}---see the proof of Proposition~\ref{prop3.4} (below) in Appendix~1 for the details.

Regression can at most estimate $\max(p,n)$ coefficients. When $p>n$, penalized regression has to drop some variables to make it estimable, implying that $\gamma>1$ does not apply to the $p>n$ case. Hence, for the $p>n$ case, we focus only on the $\mathcal{L}_{1}$ penalized regression, i.e., the lasso. As shown by \citet{efronall04, zhang10}, lasso may be thought of as a forward stagewise regression (FSR) with an $\mathcal{L}_1$ norm constraint.\footnote{The method of solving lasso by forward selection is the least angle regression (LARS). For details of LARS and its consistency, see \citet{efronall04} and \citet{zhang10}.} Hence, lasso regression can be treated as a form of GA/eGE control for FSR when $p>n$. As shown by \citet{zhang10}, even though FSR is a greedy algorithm that may result in overfitting in finite samples, it is still $\mathcal{L}_2$ consistent under the sparse eigenvalue condition

Thus, for $p>n$, we set the FSR estimate, $b_{FSR}$, to be the unpenalized regression estimate, and the $\mathcal{L}_1$ penalized regression estimate, $b_{pen}$, to be the penalized regression estimate. In Proposition~\ref{prop3.4} and Corollary~\ref{cor3.3}, we show that the lasso preserves the properties and interpretations of the $n\geqslant p$ case by reducing the overfitting inherent in FSR.
%
%
\begin{prop}[$\mathcal{L}_2$ distance between the $\mathcal{L}_1$ penalized and unpenalized FSR estimates]
Under \textbf{A1} to \textbf{A6} and based on Lemma~\ref{lem3.1}, Propositions~\ref{prop3.1}, and~\ref{prop3.2} and the sparse eigenvalue condition,

	\begin{enumerate}

        \item (\textbf{Validation}) The following bound holds with probability $\varpi(1-1/n_{t})$
		    \begin{eqnarray}
		      	\Vert b_{FSR}-b_{pen}\Vert_{2}
                & \leqslant
		      	\sqrt{\left|\frac{1}{\rho_{re}n_{t}}
                \frac{\Vert e_{t}\Vert_{2}^{2}}{(1-\sqrt{\epsilon})} -
                \frac{1}{\rho_{re}n_{s}}\Vert e_{s}\Vert_{2}^{2}\right|} \nonumber \\
		      	& + \sqrt{\frac{4}{\rho_{re}n_{s}}
                \Vert e_{s}^{T}X_{s}\Vert_{\infty}\Vert b_{FSR}\Vert_{1}} + \left(\frac{\varsigma}{\rho_{re}}\right)^{\frac{1}{2}}
		      	\label{eq1:prop3.4}
		    \end{eqnarray}
            where $\rho_{re}$ is the minimum of the sparse eigenvalues of $X^{T}X$ and $b_{FSR}$ is the FSR estimate.

        \item (\textbf{$K$-fold cross-validation}) The following bound holds with probability $\varpi(1-1/n_{t})$
		    \begin{eqnarray}
		    	\frac{1}{K}\sum_{q=1}^K \left\Vert b_{FSR}^q - b_{pen}^q\right\Vert _{2}^{2}
                & \leqslant
		      	\left\vert \frac{1}{K}\sum_{q=1}^K \frac{1}{n_t\cdot \bar{\rho}_{re}}
		      	\frac{\left\Vert e_t^q \right\Vert_{2}^{2}}{1-\sqrt{\epsilon}}
		      	- \frac{1}{K}\sum_{q=1}^K \frac{1}{n_s\cdot \bar{\rho}_{re}}
		      	\left\Vert e_s^q \right\Vert_{2}^{2}\right\vert \notag \\
		      	& +\frac{1}{K}\sum_{q=1}^K\frac{4}{n_s\cdot \bar{\rho}_{re}}
		      	\left\Vert \left( e_s^q\right)^TX_s^q\right\Vert_{\infty}
                \left\Vert b_{FSR}\right\Vert_{1}
		      	+ \frac{\varsigma}{\bar{\rho}_{re}}
		      	\label{eq2:prop3.4}
		    \end{eqnarray}
            where $\bar{\rho}_{re}$ is defined as $\min\left[\rho_{re}^q~\vert~\rho_{re}^q \mbox{ is the minimal restricted eigenvalue of }\left(X^q_s\right)^{T}X^q_s, \forall q\right]$ and  $b_{FSR}^q$ is the FSR estimator in the $q$th round of validation.

	\end{enumerate}

\label{prop3.4}
\end{prop}
%
%
\begin{cor}[$\mathcal{L}_2$ consistency for the $\mathcal{L}_1$-penalized regression estimate when $n<p$]
Under \textbf{A1} to \textbf{A6} and based on Propositions~\ref{prop3.1}, \ref{prop3.2}, \ref{prop3.2} and the sparse eigenvalue condition, $b_{pen}$ converges in the $\mathcal{L}_{2}$ norm asymptotically to the true DGP if \textup{$\lim_{n\rightarrow\infty}\log(p)/\widetilde{n}=0$}.

\label{cor3.3}
\end{cor}

\begin{figure}
	\centering
	\includegraphics[width=0.45\paperwidth]{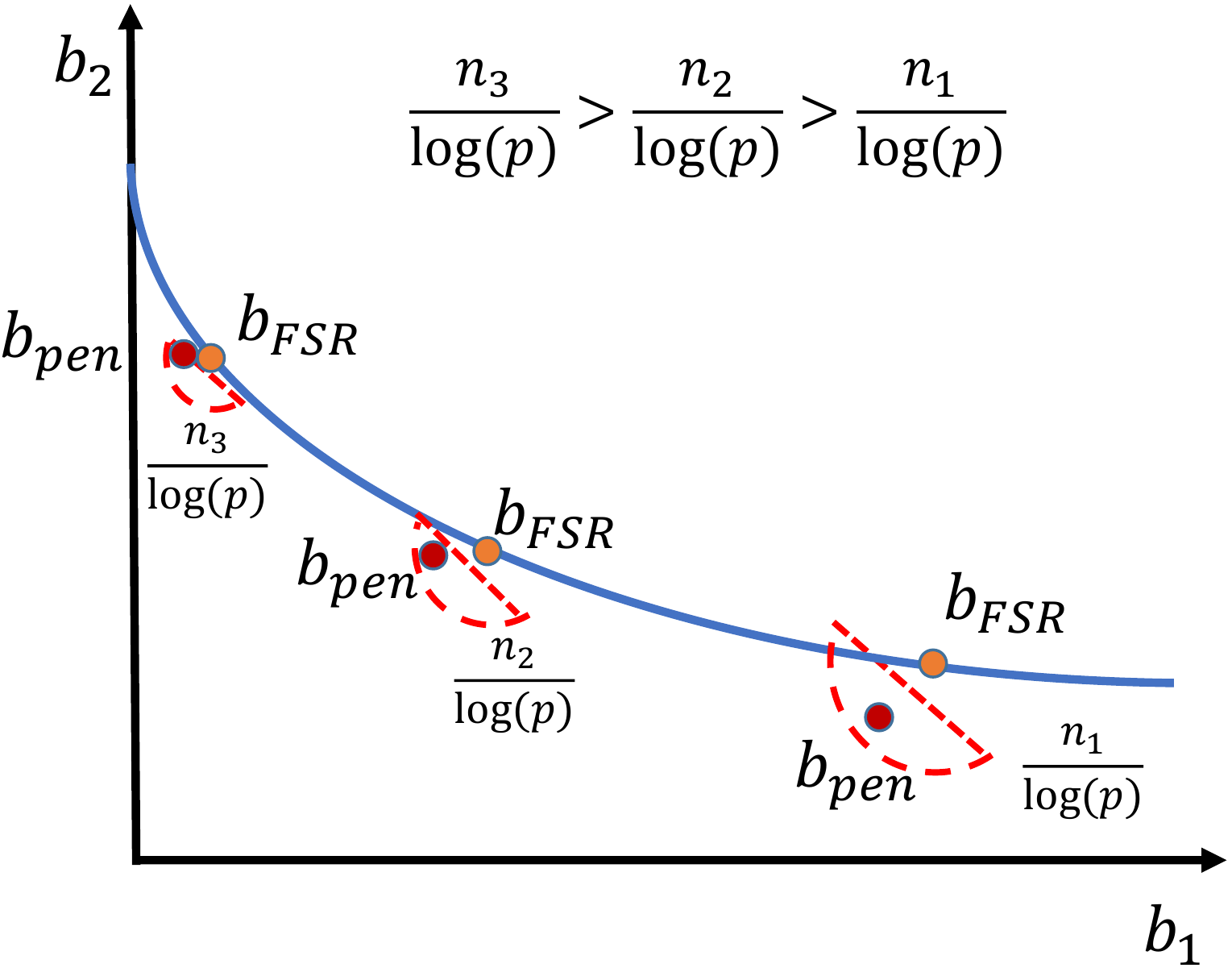}

	\caption{\label{figillustrateFSR}Illustration of the relation between the FSR estimate and the $\mathcal{L}_1$ penalized FSR estimate}
\end{figure}

The interpretation of Proposition~\ref{prop3.4} and Corollary~\ref{cor3.3} is very similar to the interpretation of Eqs.~(\ref{eq1:thm3.1}) and (\ref{eq2:thm3.1}), which specifies the upper bound of $\mathcal{L}_2$ difference between $b_{FSR}$ and $b_{pen}$ as a function of GA of the FSR estimate and the population error. Hence, the interpretation for Proposition~\ref{prop3.4} and Corollary~\ref{cor3.3} can also be illustrated in Figure~\ref{figillustrateFSR}.


\section{Simulation Study}

In sections~2 and~3, we use eTE to measure in-sample fit and eGE to measure out-of-sample fit. However, to measure the GA and degree of overfitting, we need to compare the eTE and eGE to the total sum of square for the training set and test set, respectively. To summarize the in-sample and out-of-sample goodness of fit, we propose the generalized $R^2$:
\begin{equation}
	GR^{2} =
	\left(1 - \frac{\mathcal{R}_{n_s}(b_{train}|Y_{s},X_{s})}{\mathrm{TSS}\left(Y_{s}\right)}\right)
	\times
	\left(1 - \frac{\mathcal{R}_{n_t}(b_{train}|Y_{t},X_{t})}{\mathrm{TSS}\left(Y_{t}\right)}\right) \\
	= R^2_s \times R^2_t
	\label{grsq}
\end{equation}
where $R^2_s$ is the $R^2$ for the test set and $R^2_t$ for the training set. If $b_{train}$ is consistent, both $\mathcal{R}_{n_s}(b_{train}|Y_{s},X_{s})$ and $\mathcal{R}_{n_t}(b_{train}|Y_{t},X_{t})$ converge to the same limit in probability as $\widetilde{n}\rightarrow\infty$.

\begin{table}[h!]
	\centering
	\caption{Schematic table  and $GR^2$}
	\begin{tabular}{|l|c|c|}\hline
		\diagbox[width=4em]{$R^2_t$}{$R^2_s$}
                  & high                         & low
        \\ \hline
		high      & high $GR^2$, the ideal model & relatively low $GR^2$, overfitting
        \\ \hline
		low       & relatively low $GR^2$        & low $GR^2$, underfitting
        \\ \hline
	\end{tabular}
	\label{table1}
\end{table}

Clearly $GR^2$ considers both the in-sample fit and the out-of-sample fit. Intuitively, there are four scenarios for $GR^2$. As summarized in Table~\ref{table1}, a model that fits both the training set and the test set well will have high $R^2_t$ and $R^2_s$ values and hence a high $GR^2$. When overfitting occurs, the $R^2_t$ will be relatively high and the $R^2_s$ will be low, reducing the $GR^2$. In contrast, the selected model may fit the training set and test set poorly, which is called underfitting. When underfitting occurs, the $R^2_t$ and $R^2_s$ will be low, reducing the $GR^2$ further. It is unlikely but possible that the model estimated on the training set fits the test set better (the $R^2_s$ is high while the $R^2_t$ low).

In the simulations, we illustrate the fact that penalized regression, by constraining the $\mathcal{L}_\gamma$ norm of the regression estimate, obtains a superior GA compared with OLS or FSR. Penalized regression is less efficient at model selection when the norm of penalty term $\gamma > 1$, as illustrated in Figures~\ref{figpenreg} and~\ref{figillustrate}. Thus, we focus on the $\mathcal{L}_1$ penalized or lasso-type regression.

For the simulations, we assume the outcome $y$ is generated by the following DGP:
\[
	y = X'\beta+u = X_{1}^T\beta_{1} + X_{2}^T\beta_{2} + u
\]
where $X=\left(x_{1},\cdots,x_{p}\right)\in\mathbb{R}^{p}$ is generated by a multivariate Gaussian distribution with zero mean, $\mbox{var}(x_{i}) = 1$, $\mbox{corr}(x_{i},x_{j}) = 0.9, \forall i,\thinspace j$, $\beta_{1}=\left(2,4,6,8,10,12\right)^T$ and $\beta_{2}$ is a $(p - 6)$-dimensional zero vector. $u$ is generated by a Gaussian distribution with zero mean. Here $x_{i}$ does not cause $x_{j}$ and there is no causal relationship between $u$ and $x_{i}$. We set the sample size at 250, $p$ is set at 200 or 500 and $\mathrm{var}(u)$ at 1 or 5. Hence, we have four different cases. In each case, we repeat the simulation 50 times. In each simulation, we apply the Algorithm~1 to find the estimate of $\beta$ and calculate its distance to the true value, the eGE, as well as our goodness-of-fit measure $GR^{2}$. As a comparison, we also apply OLS in the $n\geqslant p$ cases and the FSR algorithm for the $n<p$ cases.

Boxplots (see Appendix~2) summarize the estimates of all the coefficients in $\beta_{1}$ (labeled $b_1$ to $b_6$) along with the four worst estimates among the coefficients in $\beta_{2}$ (labeled $b_7$ to $b_{10}$), where `worst' refers to estimates with the largest bias. The lasso and OLS/FSR estimates and histograms of the $GR^{2}$ are reported for each case in Figures~9--12 (Appendix~2). Lastly, the distance between the estimates and the true values, the eGE, and the $GR^{2}$ (all averages across the 50 simulations) are reported in Table~1 for all four cases.

\begin{figure}[t]
\section*{Appendix 2}
	\centering
	\subfloat[Lasso estimates]
	{\includegraphics[width=0.33\paperwidth]{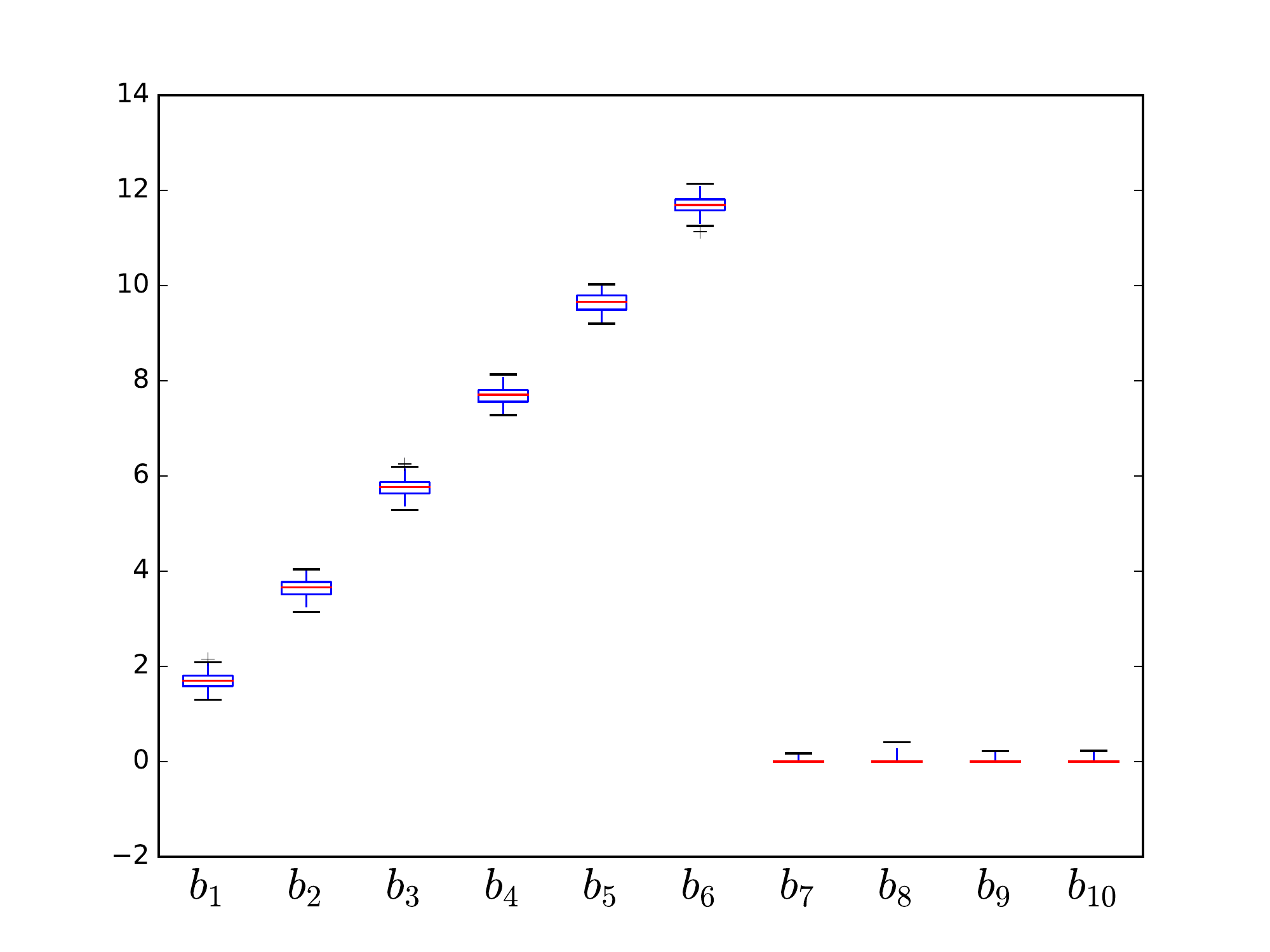}}
	\subfloat[OLS estimates]
	{\includegraphics[width=0.33\paperwidth]{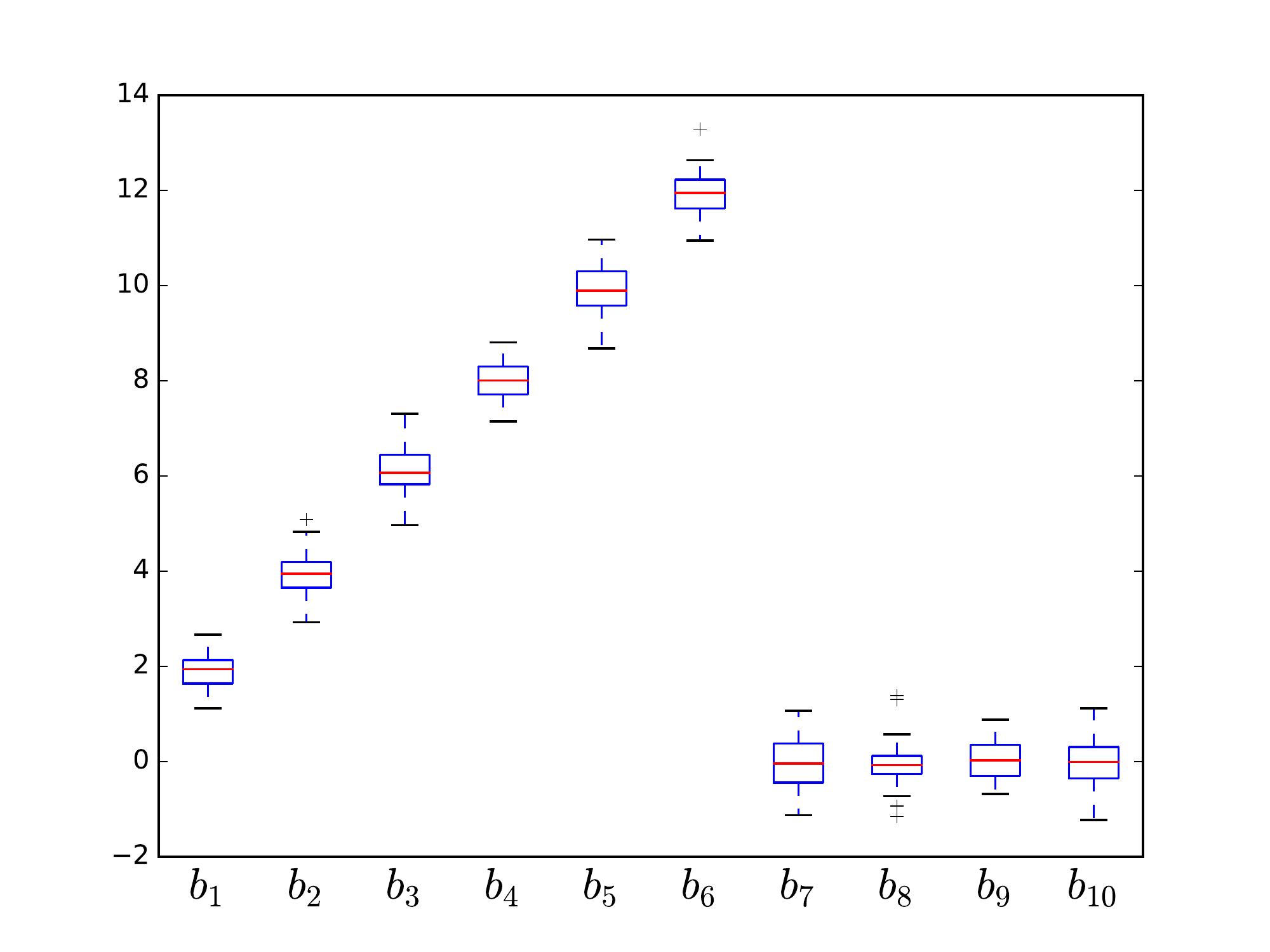}}

	\subfloat[Histogram of $GR^{2}$]
	{\includegraphics[width=0.6\paperwidth]{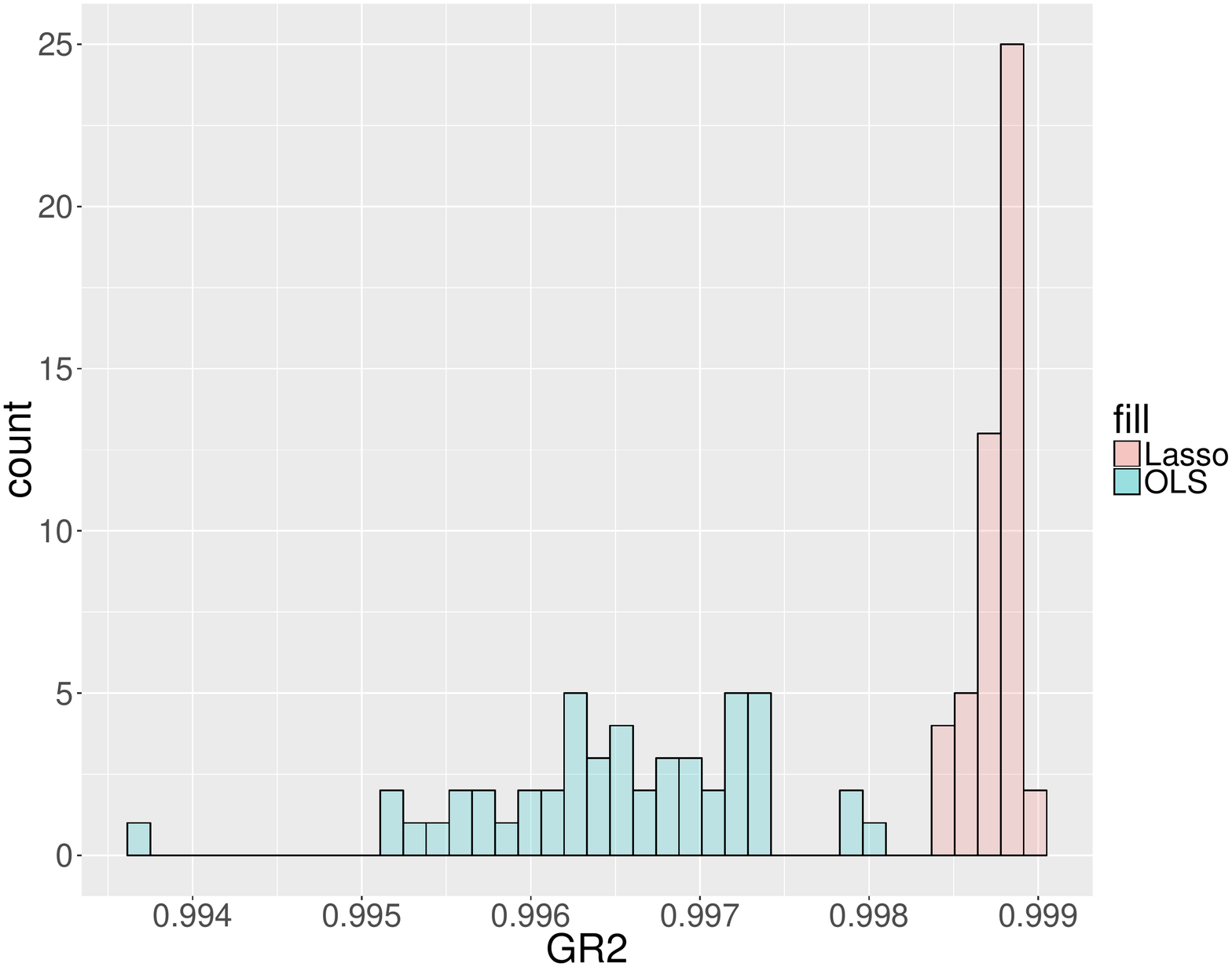}}

	\caption{\label{p=200v=1}Boxplots of estimates and $GR^2$ for
            DGP $n=250,\thinspace p=200, \thinspace \mathrm{var}(u)=1$ }
\end{figure}

When $n>p$, as we can see from the boxplot in Figure~(\ref{p=200v=1}) and (\ref{p=200v=5}), both lasso and OLS perform well; in the case $var(u)=1$ both lasso and OLS perform better than when $var(u)=5$. All the coefficient estimates are centered around the true values, and the deviations are relatively small. However, lasso outperforms OLS for the estimates of $\beta_{2}$ by having much smaller deviations. Indeed, a joint significance test ($F$ test) fails to reject the null hypothesis that all coefficients in $\beta_{2}$ are zero for the OLS estimates. As shown in Figure~(\ref{p=200v=1}) and (\ref{p=200v=5}), the $GR^{2}$ for the lasso is marginally larger than for OLS, but the differences are inconsequential.

\begin{figure}[ht]
	\centering
	\subfloat[Lasso estimates]
	{\includegraphics[width=0.33\paperwidth]{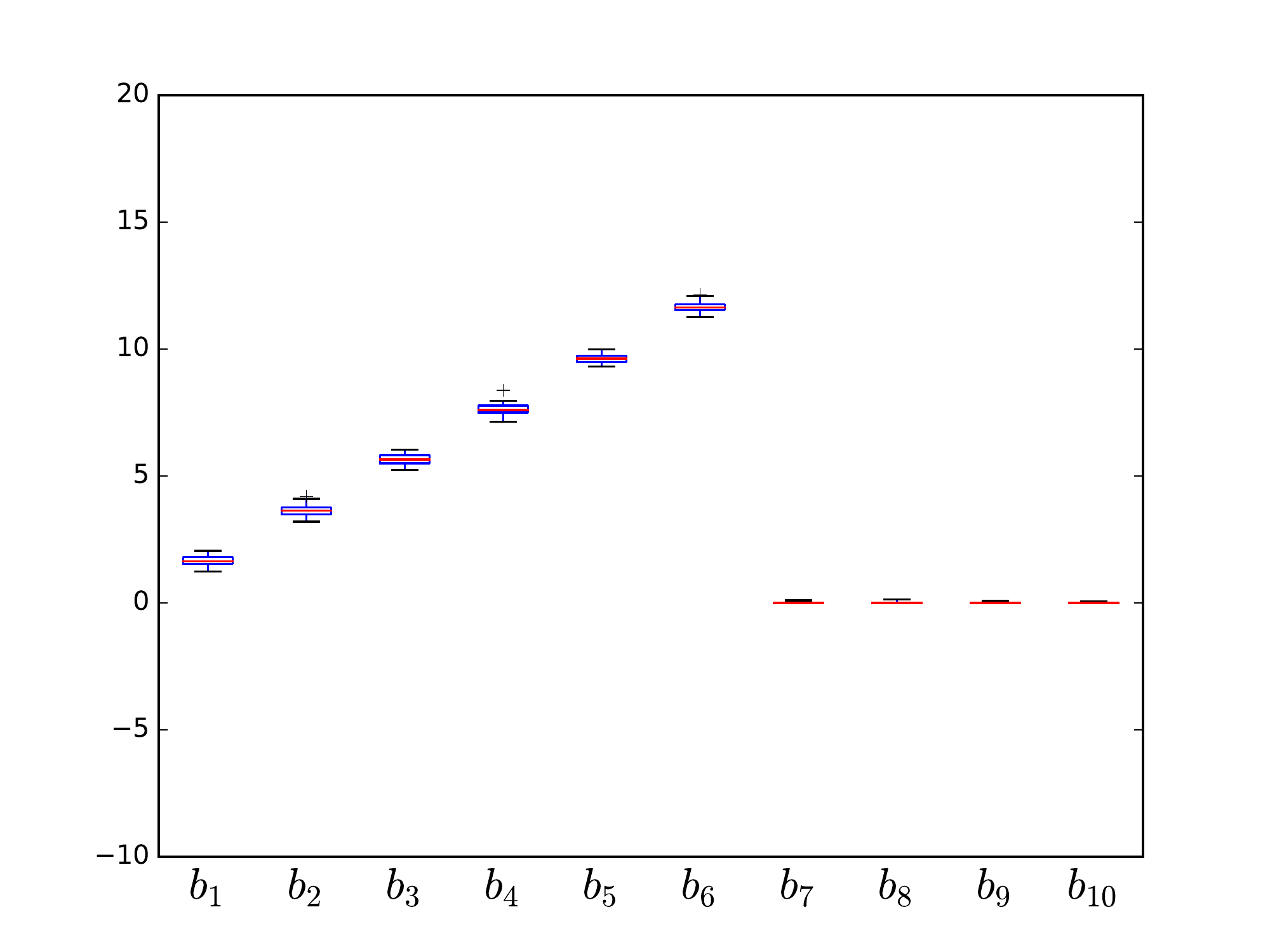}}
	\subfloat[FSR estimates]
	{\includegraphics[width=0.33\paperwidth]{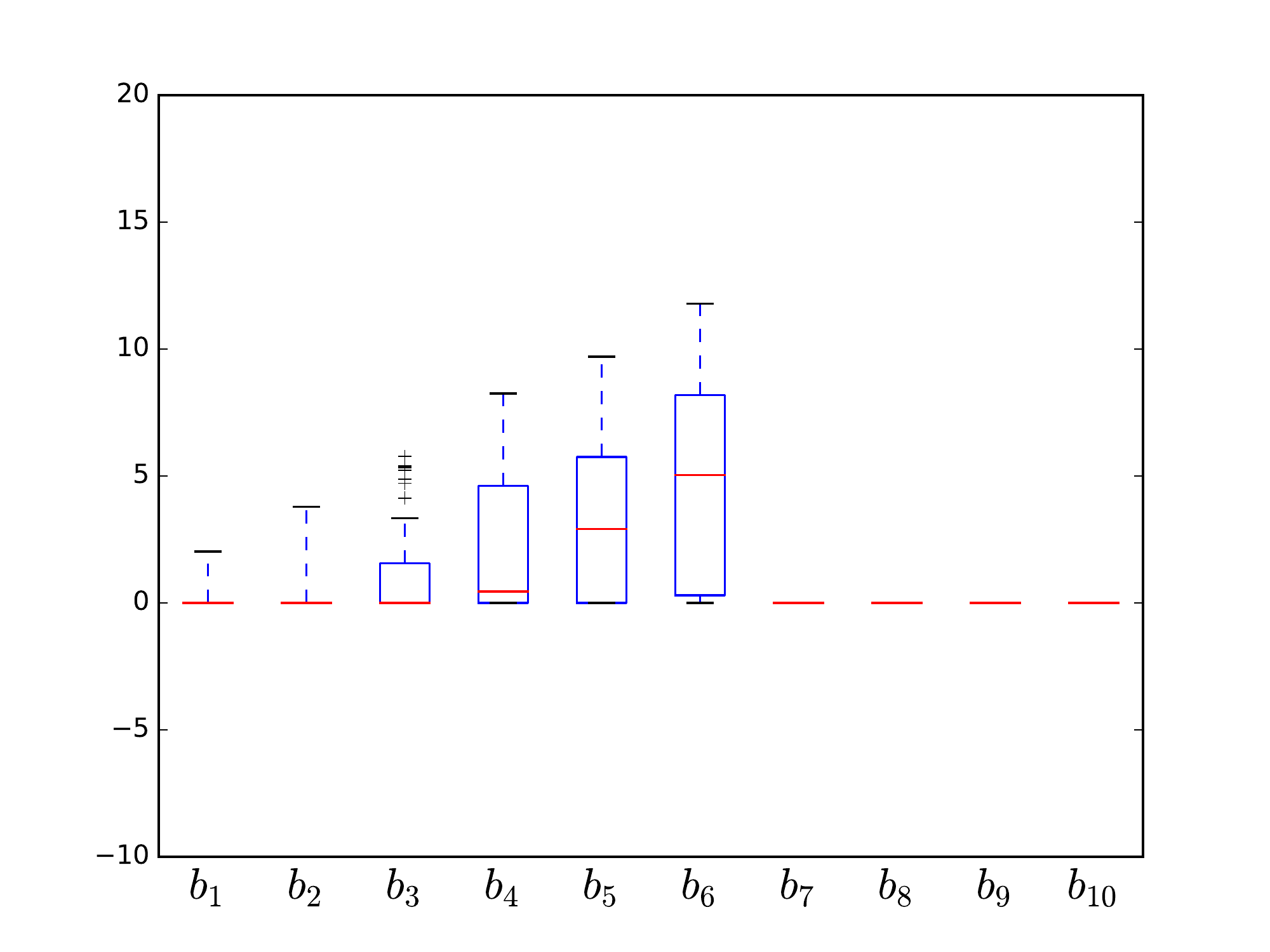}}

	\subfloat[Histogram of $GR^{2}$]
	{\includegraphics[width=0.6\paperwidth]{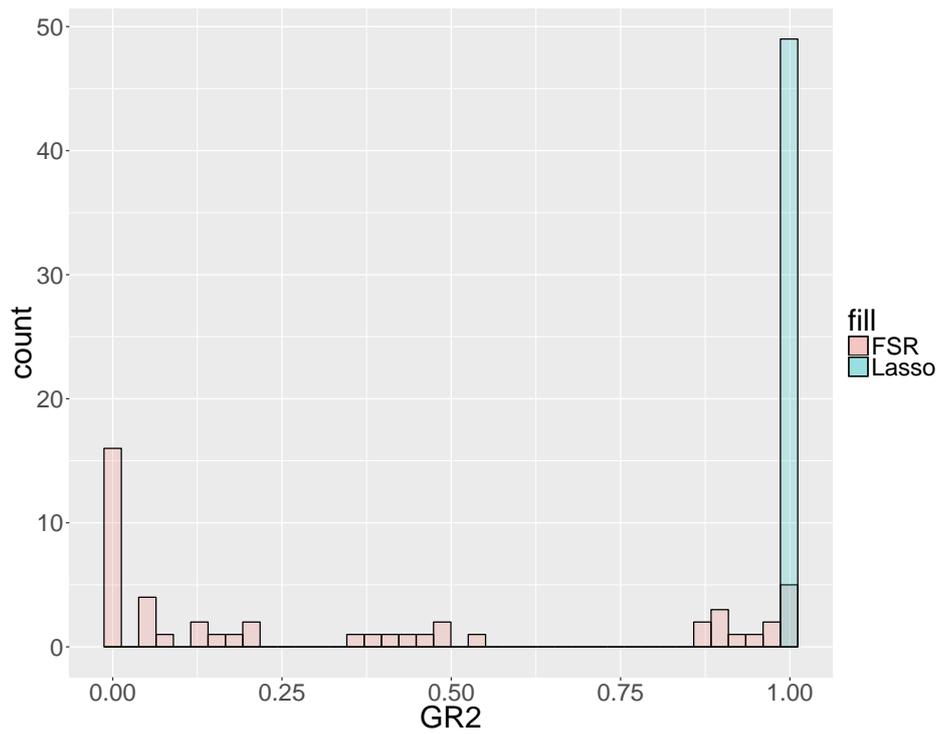}}

	\caption{\label{p=500v=1}Boxplots of estimates and $GR^2$ for
            DGP $n=250,\thinspace p=500, \thinspace \mathrm{var}(u)=1$ }
\end{figure}

\begin{figure}[ht]
	\centering
	\subfloat[Lasso estimates]
	{\includegraphics[width=0.33\paperwidth]{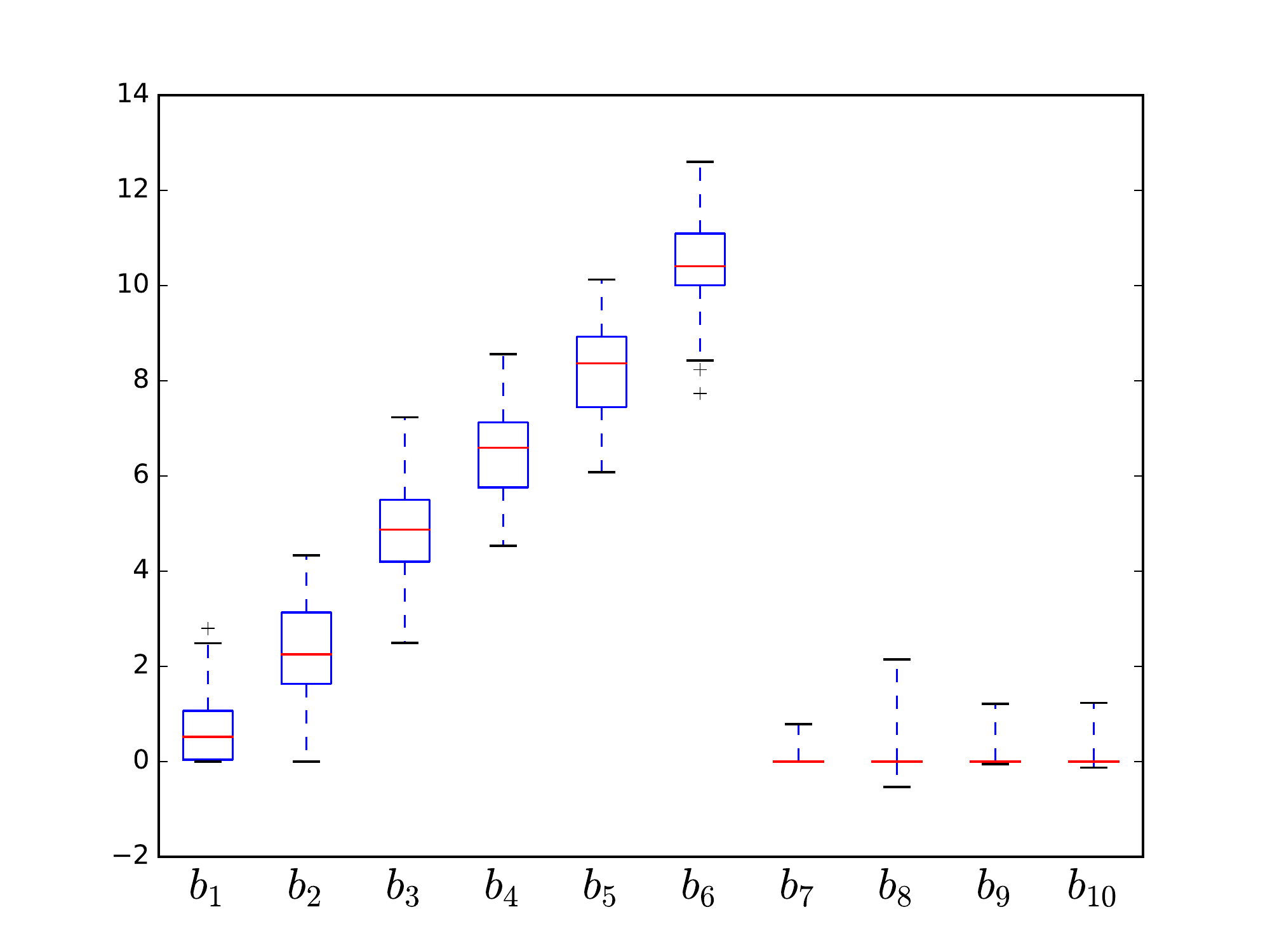}}
	\subfloat[OLS estimates]
	{\includegraphics[width=0.33\paperwidth]{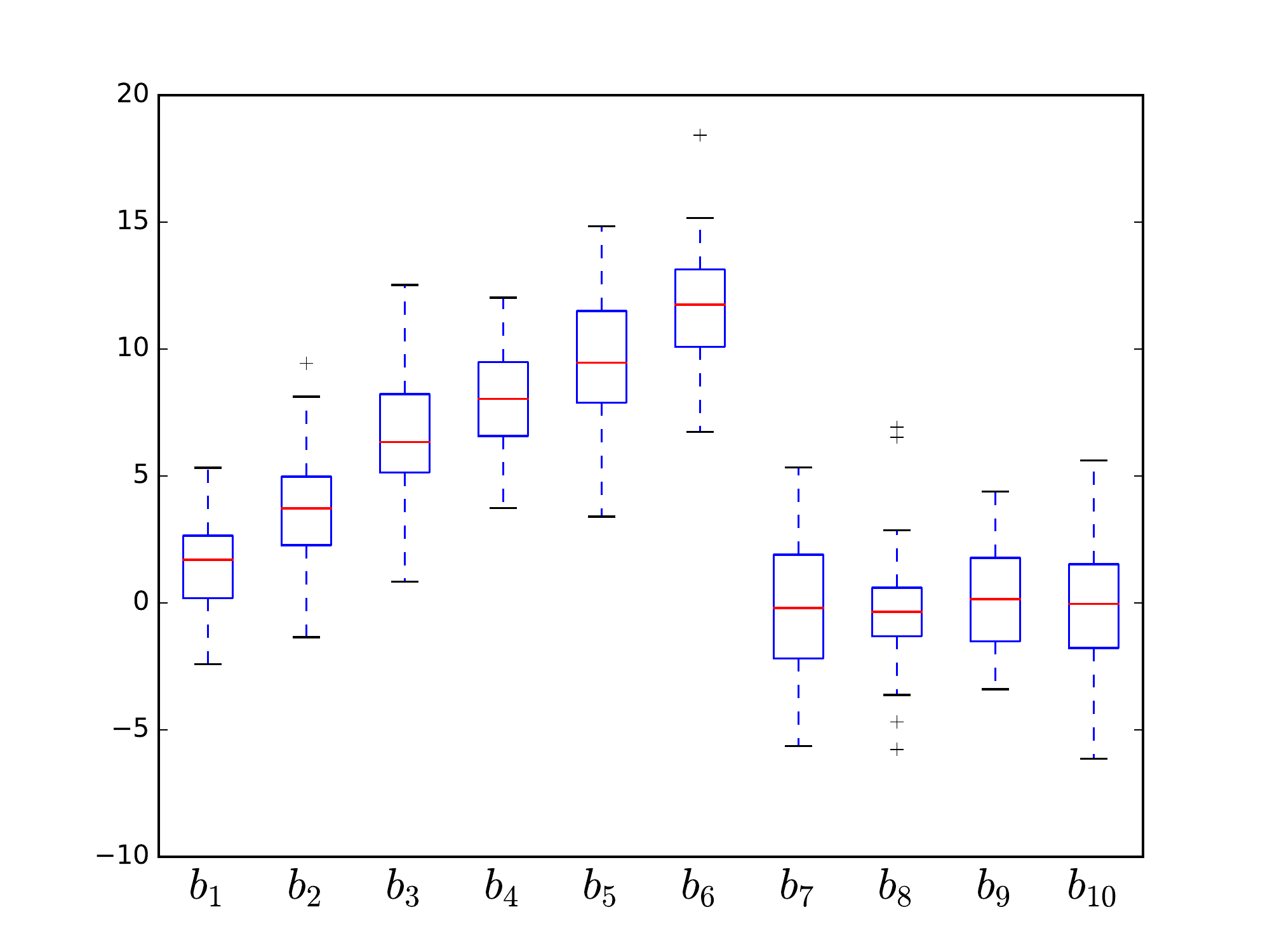}}

	\subfloat[Histogram of $GR^{2}$]
	{\includegraphics[width=0.6\paperwidth]{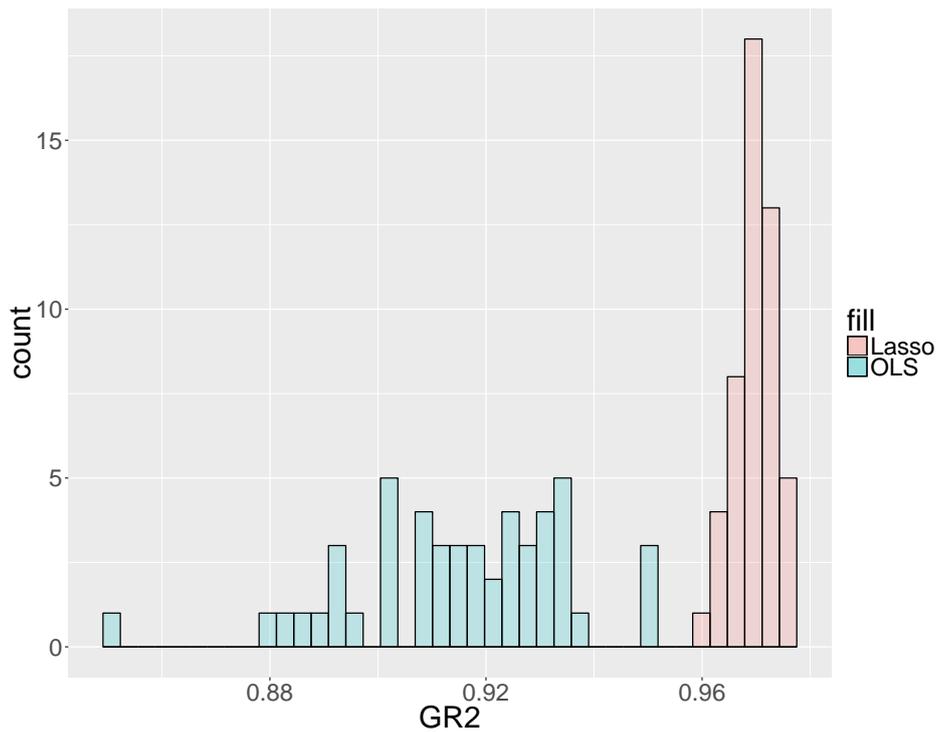}}

	\caption{\label{p=200v=5}Boxplots of estimates and $GR^2$ for
            DGP $n=250,\thinspace p=200, \thinspace \mathrm{var}(u)=5$ }
\end{figure}

\begin{figure}[ht]
	\centering
	\subfloat[Lasso estimates]
	{\includegraphics[width=0.33\paperwidth]{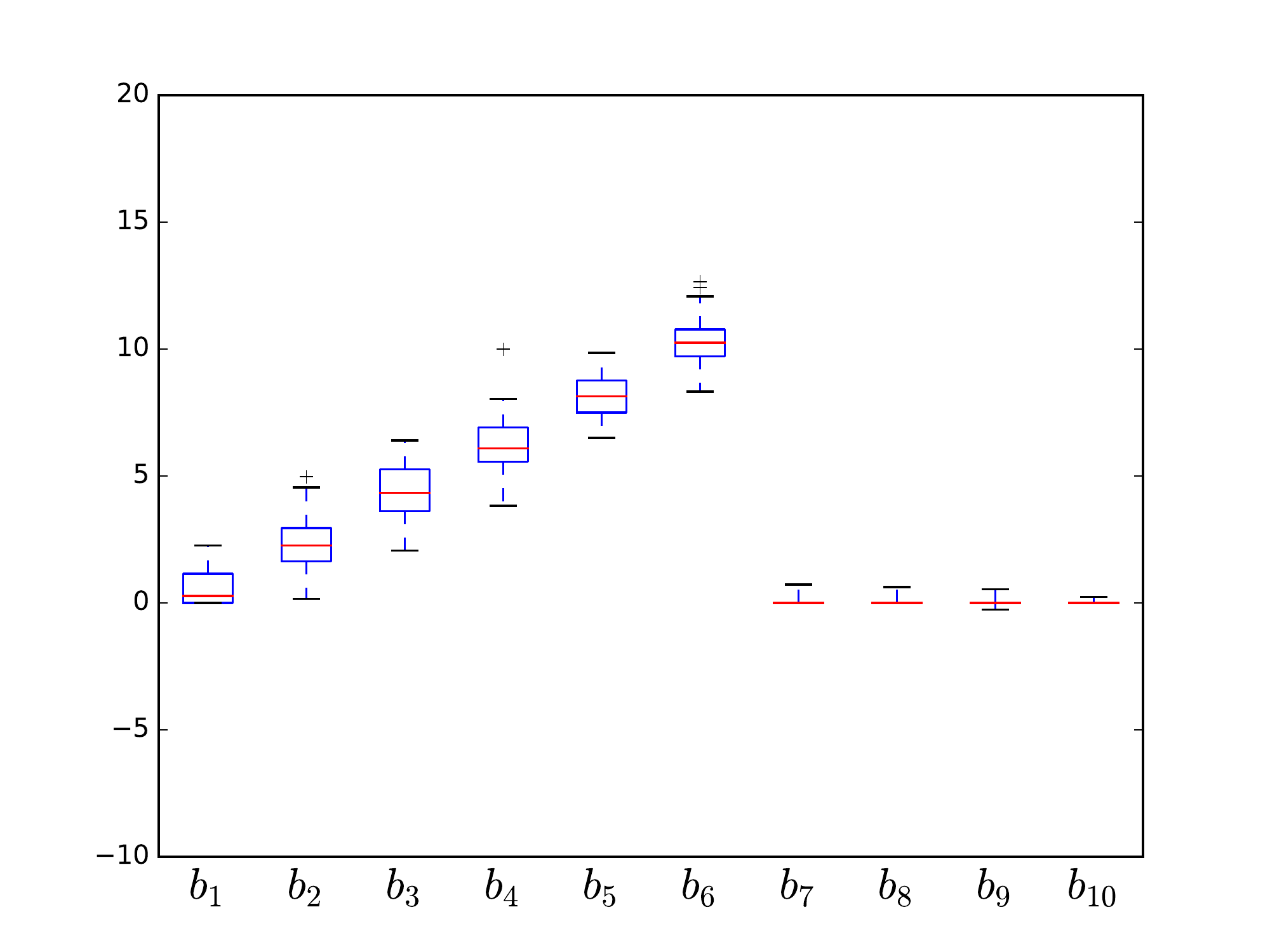}}
	\subfloat[FSR estimates]
	{\includegraphics[width=0.33\paperwidth]{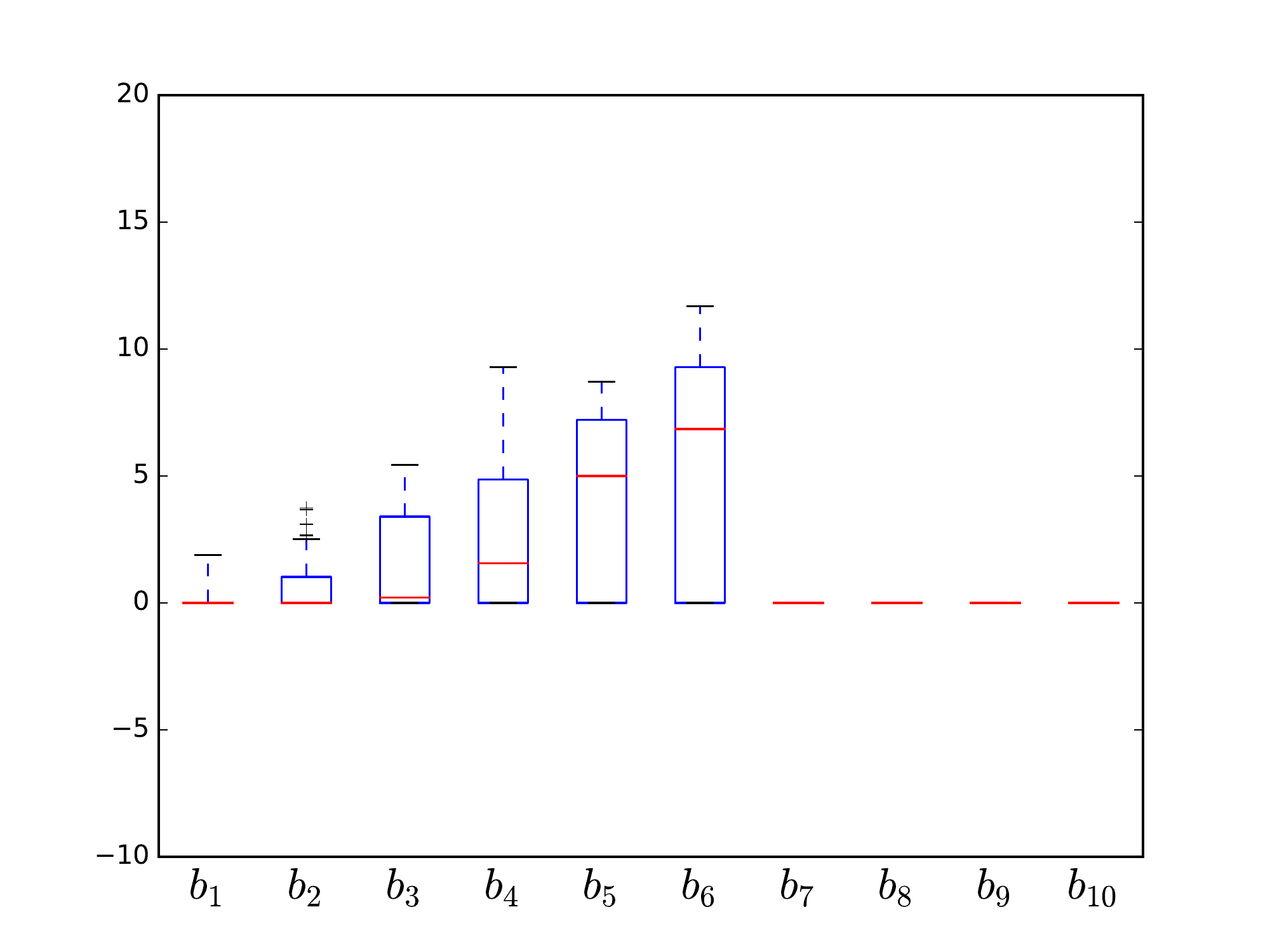}}

	\subfloat[Histogram of $GR^{2}$]
	{\includegraphics[width=0.6\paperwidth]{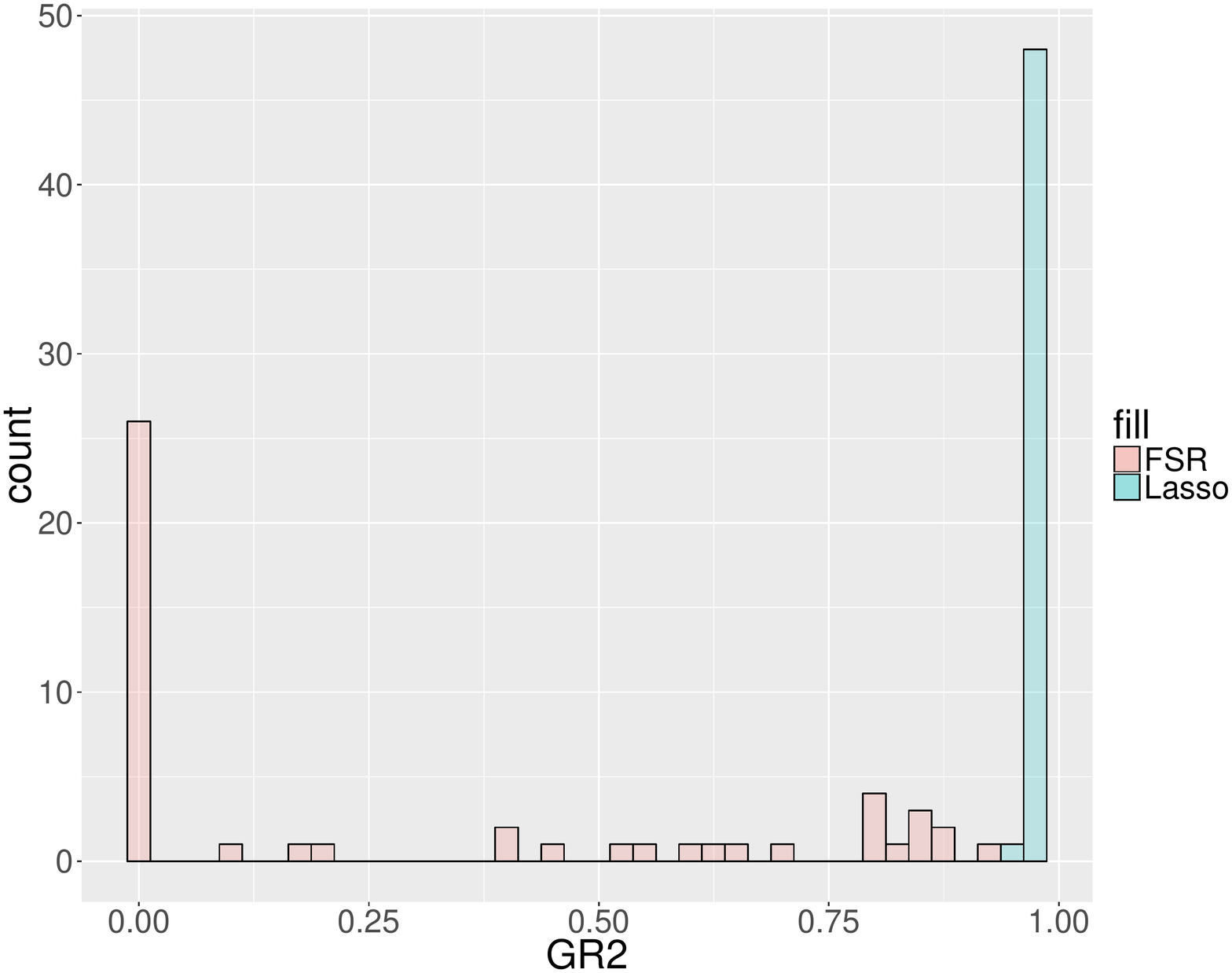}}

	\caption{\label{p=500v=5}Boxplots of estimates and $GR^2$ for
            DGP $n=250,\thinspace p=500, \thinspace \mathrm{var}(u)=5$ }
\end{figure}

When $n<p$, the regression model is not identified, OLS is infeasible, and we apply FSR. As shown in Figures~(\ref{p=500v=1}) and (\ref{p=500v=5}), lasso still performs well and correctly selects the variables with non-zero coefficients. In contrast, although FSR also correctly identifies the non-zero coefficients, its biases and deviations are much larger than for the lasso. For the $p=500$ case shown in Figures~(\ref{p=500v=1}) and (\ref{p=500v=5}), the $GR^2$ for the FSR indicate that the FSR estimates are unreliable. Generally speaking, overfitting is controlled well by lasso (all the $GR^2$ are close to 1) whereas the performance of FSR is mixed, as reflected by the deteriorating $GR^2$ as $p$ increases. This suggests that, by imposing an $\mathcal{L}_1$ penalty on estimates, lasso mitigates the overfitting problem and that the advantage of lasso is likely to be more pronounced as $p$ increases.

\begin{table}[ht]
	\caption{Average bias, training error, generalization error, in-sample $R^2$, out-of-sample $R^2$, and $GR^{2}$ for lasso and OLS/FSR when $n=250$}
	\centering

	\begin{tabular}{l.....}
		\toprule
		Measure & \multicolumn{2}{c}{$var(u) = 1$} & \multicolumn{2}{c}{$var(u) = 5$} \\
        \cmidrule(r{0.5em}){2-3} \cmidrule(r{0.5em}){4-5}
        & \multicolumn{1}{c}{$p = 200$}
		& \multicolumn{1}{c}{$p = 500$}
		& \multicolumn{1}{c}{$p = 200$}
		& \multicolumn{1}{c}{$p = 500$} \\
		\midrule
		Bias\\
		\hspace*{4mm}$b_{Lasso}$   & 0.7923 & 0.8810   & 3.8048   & 4.1373     \\
		\hspace*{4mm}$b_{OLS/FSR}$ & 0.9559 & 11.7530  & 4.7797   & 13.7622    \\
		Training error\\
		\hspace*{4mm}Lasso         & 0.9167 & 0.8625   & 22.2476  & 21.1334    \\
		\hspace*{4mm}OLS/FSR       & 0.2164 & 832.9988 & 5.4097   & 1034.2636  \\
		Generalization error\\
		\hspace*{4mm}Lasso         & 1.1132 & 1.1478   & 27.8672  & 28.5125    \\
		\hspace*{4mm}OLS/FSR       & 5.2109 & 852.5822 & 134.8725 & 1070.6329  \\
		$R^2$, in-sample\\
		\hspace*{4mm}Lasso         & 0.9994 & 0.9994   & 0.9866   & 0.9867   \\
		\hspace*{4mm}OLS/FSR       & 0.9999 & 0.4678   & 0.9967   & 0.3619   \\
		$R^2$, out-of-sample\\
		\hspace*{4mm}Lasso         & 0.9993 & 0.9993   & 0.9830   & 0.9826   \\
		\hspace*{4mm}OLS/FSR       & 0.9967 & 0.4681   & 0.9181   & 0.3627   \\
		$GR^{2}$\\
		\hspace*{4mm}Lasso         & 0.9988 & 0.9987   & 0.9698   & 0.9695   \\
		\hspace*{4mm}OLS/FSR       & 0.9965 & 0.3659   & 0.9151   & 0.2935   \\
		\bottomrule
	\end{tabular}
\end{table}

Table~1 reinforces the impressions from the boxplots and histograms. When $p=200$ OLS of course performs extremely well in terms of training error and more poorly in terms of generalization error while its $GR^2$ is very close to the lasso value. For $n<p$ what is noteworthy is the stable performance of the lasso relative to that of FSR. The training errors, generalization errors, and $GR^2$ are particulary poor for FSR, again illustrating the advantage of the lasso in avoiding overfitting.

\section{Conclusion}

In this paper, we study the performance of penalized and unpenalized extremum estimators from the perspective of generalization ability (GA), the ability of a model to predict outcomes in new samples from the same population. We analyze the GA of penalized regression estimates for the $n\geq p$ and the $n<p$ cases. We propose inequalities for the extremum estimators, which bound empirical out-of-sample prediction errors as a function of in-sample errors, sample sizes, model complexity and the heaviness in the tail of the error distribution. The inequalities serve not only to quantify GA, but also to illustrate the trade-off between in-sample and out-of-sample fit, which in turn may be used for tuning estimation hyperparameters, such as the number of folds $K$ in cross-validation or $n_t/n_s$ in validation. We show that some finite-sample and asymptotic properties of the penalized estimators are explained directly by their GA. Furthermore, we use the bounds to quantify the $\mathcal{L}_2$-norm difference between the penalized and corresponding unpenalized regression estimates.

Our work sheds new light on penalized regression and on the applicability of GA for model selection, as well as further insight into the bias-variance trade-off. In this paper, we focus mainly on implementing penalized regression. However, other penalty methods, such as penalized MLE, functional regression, principle component analysis and decision trees, potentially fit the GA framework. Furthermore, the results we establish for penalized regression and GA may be implemented with other empirical methods, like the EM algorithm, clustering, mixture and factor modeling, Bayes networks, and so on.

In providing a general property for all penalized regressions, the generalization error bounds are necessarily conservative. Finer error bounds may well be derived by focusing on specific penalized regression methods. Lastly, as an early attempt to incorporate GA analysis into econometrics, we focus on the i.i.d.\ case. However, it is clear that the framework has the potential to be generalized to non-i.i.d.\ data in settings like $\alpha$- and $\beta$-mixing stationery time series data as well as dependent and non-identical panel data.

\newpage{}
\section*{References}

\bibliographystyle{plainnat}
\bibliography{LASSOrefs}

\begin{thebibliography}{62}
\providecommand{\natexlab}[1]{#1}
\providecommand{\url}[1]{\texttt{#1}}
\expandafter\ifx\csname urlstyle\endcsname\relax
  \providecommand{\doi}[1]{doi: #1}\else
  \providecommand{\doi}{doi: \begingroup \urlstyle{rm}\Url}\fi

\bibitem[Akaike(1973)]{akaike73}
Hirotugu Akaike.
\newblock Information theory and an extension of the maximum likelihood
  principle.
\newblock In B.~N. Petrov and F.~Csaki, editors, \emph{2\textsuperscript{nd}
  International Symposium on Information Theory, Tsahkadsor, Armenia, USSR},
  pages 267--281. Budapest: Akademiai Kaido, 1973.

\bibitem[Amemiya(1985)]{amemiya1985advanced}
Takeshi Amemiya.
\newblock \emph{Advanced econometrics}.
\newblock Harvard university press, 1985.

\bibitem[Belloni et~al.(2013)Belloni, Chernozhukov, Fern{\'a}ndez-Val, and
  Hansen]{belloni2013program}
Alexandre Belloni, Victor Chernozhukov, Ivan Fern{\'a}ndez-Val, and Chris
  Hansen.
\newblock Program evaluation with high-dimensional data.
\newblock \emph{arXiv preprint arXiv:1311.2645}, 2013.

\bibitem[Bickel et~al.(2009)Bickel, Ritov, and Tsybakov]{bickeletal09}
Peter~J Bickel, Ya'acov Ritov, and Alexandre~B Tsybakov.
\newblock Simultaneous analysis of lasso and dantzig selector.
\newblock \emph{The Annals of Statistics}, 37:\penalty0 1705--1732, 2009.

\bibitem[Blundell et~al.(2004)Blundell, Dias, Meghir, and
  Reenen]{blundell2004evaluating}
Richard Blundell, Monica~Costa Dias, Costas Meghir, and John Reenen.
\newblock Evaluating the employment impact of a mandatory job search program.
\newblock \emph{Journal of the European economic association}, 2\penalty0
  (4):\penalty0 569--606, 2004.

\bibitem[Breiman(1995)]{breiman95}
Leo Breiman.
\newblock Better subset regression using the nonnegative garrote.
\newblock \emph{Technometrics}, 37\penalty0 (4):\penalty0 373--384, 1995.

\bibitem[Candes and Tao(2007)]{candestao07}
Emmanuel~J. Candes and Terence Tao.
\newblock The dantzig selector: statistical estimation when $p$ is much larger
  than $n$.
\newblock \emph{The Annals of Statistics}, pages 2313--2351, 2007.

\bibitem[Caner(2009)]{caner09}
Mehmet Caner.
\newblock {Lasso}-type gmm estimator.
\newblock \emph{Econometric Theory}, 25\penalty0 (1):\penalty0 270--290, 2009.

\bibitem[Cesa-Bianchi et~al.(2004)Cesa-Bianchi, Conconi, and
  Gentile]{cesa2004generalization}
Nicolo Cesa-Bianchi, Alex Conconi, and Claudio Gentile.
\newblock On the generalization ability of on-line learning algorithms.
\newblock \emph{IEEE Transactions on Information Theory}, 50\penalty0
  (9):\penalty0 2050--2057, 2004.

\bibitem[Chickering et~al.(2004)Chickering, Heckerman, and Meek]{chickering04}
David~M. Chickering, David Heckerman, and Christopher Meek.
\newblock Large-sample learning of {B}ayesian networks is {NP}-hard.
\newblock \emph{Journal of Machine Learning Research}, 5:\penalty0 1287--1330,
  2004.

\bibitem[Dolton(2006)]{dolton2006econometric}
Peter Dolton.
\newblock \emph{The econometric evaluation of the New Deal for Lone Parents}.
\newblock PhD thesis, Department of Economics, University of Michigan, 2006.

\bibitem[Efron et~al.(2004)Efron, Hastie, Johnstone, and
  Tibshirani]{efronall04}
Bradley Efron, Trevor Hastie, Iain Johnstone, and Robert Tibshirani.
\newblock Least angle regression.
\newblock \emph{The Annals of statistics}, 32\penalty0 (2):\penalty0 407--499,
  2004.

\bibitem[Frank and Friedman(1993)]{frankfriedman93}
Ildiko~E. Frank and Jerome~H. Friedman.
\newblock A statistical view of some chemometrics regression tools.
\newblock \emph{Technometrics}, 35\penalty0 (2):\penalty0 109--135, 1993.

\bibitem[Friedman et~al.(2010)Friedman, Hastie, and Tibshirani]{friedman10}
Jerome Friedman, Trevor Hastie, and Robert Tibshirani.
\newblock A note on the group lasso and a sparse group lasso.
\newblock \emph{arXiv preprint arXiv:1001.0736}, 2010.

\bibitem[Friedman et~al.(1997)Friedman, Geiger, and Goldszmidt]{friedman97}
Nir Friedman, Dan Geiger, and Moises Goldszmidt.
\newblock Bayesian network classifiers.
\newblock \emph{Machine Learning}, 29\penalty0 (2-3):\penalty0 131--163, 1997.

\bibitem[Friedman et~al.(2000)Friedman, Linial, Nachman, and Pe'er]{friedman00}
Nir Friedman, Michal Linial, Iftach Nachman, and Dana Pe'er.
\newblock Using {B}ayesian networks to analyze expression data.
\newblock In \emph{Proceedings of the Fourth Annual International Conference on
  Computational Molecular Biology}, RECOMB '00, pages 127--135, New York, NY,
  USA, 2000. ACM.

\bibitem[Fu(1998)]{fu98}
Wenjiang~J. Fu.
\newblock Penalized regressions: the bridge versus the {Lasso}.
\newblock \emph{Journal of computational and graphical statistics}, 7\penalty0
  (3):\penalty0 397--416, 1998.

\bibitem[Gechter(2015)]{gechter2015generalizing}
Michael Gechter.
\newblock Generalizing the results from social experiments: Theory and evidence
  from mexico and india.
\newblock \emph{manuscript, Pennsylvania State University}, 2015.

\bibitem[Hall and Marron(1991)]{hall1991}
Peter Hall and James~S Marron.
\newblock Local minima in cross-validation functions.
\newblock \emph{Journal of the Royal Statistical Society. Series B
  (Methodological)}, pages 245--252, 1991.

\bibitem[Hall et~al.(2011)Hall, Racine, and Li]{hall2011}
Peter Hall, Jeff Racine, and Qi~Li.
\newblock Cross-validation and the estimation of conditional probability
  densities.
\newblock \emph{Journal of the American Statistical Association}, 2011.

\bibitem[Heckerman et~al.(1995)Heckerman, Geiger, and Chickering]{heckerman95}
David Heckerman, Dan Geiger, and David~M. Chickering.
\newblock Learning {B}ayesian networks: The combination of knowledge and
  statistical data.
\newblock \emph{Machine learning}, 20\penalty0 (3):\penalty0 197--243, 1995.

\bibitem[Hoeffding(1963)]{hoeffding63}
Wassily Hoeffding.
\newblock Probability inequalities for sums of bounded random variables.
\newblock \emph{Journal of the American Statistical Association}, 58\penalty0
  (301):\penalty0 13--30, 1963.

\bibitem[Hoerl and Kennard(1970)]{hoerlkennard70}
Arthur~E. Hoerl and Robert~W. Kennard.
\newblock Ridge regression: Applications to nonorthogonal problems.
\newblock \emph{Technometrics}, 12\penalty0 (1):\penalty0 69--82, 1970.

\bibitem[Hu and Zhou(2009)]{hu2009}
Ting Hu and Ding-Xuan Zhou.
\newblock Online learning with samples drawn from non-identical distributions.
\newblock \emph{Journal of Machine Learning Research}, 10\penalty0
  (Dec):\penalty0 2873--2898, 2009.

\bibitem[Huang et~al.(2008)Huang, Horowitz, and Ma]{huangall08}
Jian Huang, Joel~L. Horowitz, and Shuangge Ma.
\newblock Asymptotic properties of bridge estimators in sparse high-dimensional
  regression models.
\newblock \emph{The Annals of Statistics}, pages 587--613, 2008.

\bibitem[Kakade and Tewari(2009)]{kakade2009generalization}
Sham~M Kakade and Ambuj Tewari.
\newblock On the generalization ability of online strongly convex programming
  algorithms.
\newblock pages 801--808, 2009.

\bibitem[Knight and Fu(2000)]{knightfu00}
Keith Knight and Wenjiang Fu.
\newblock Asymptotics for {Lasso}-type estimators.
\newblock \emph{Annals of statistics}, pages 1356--1378, 2000.

\bibitem[Kohavi et~al.(1995)]{kohavi1995study}
Ron Kohavi et~al.
\newblock A study of cross-validation and bootstrap for accuracy estimation and
  model selection.
\newblock In \emph{Ijcai}, volume~14, pages 1137--1145, 1995.

\bibitem[McDonald et~al.(2011)McDonald, Shalizi, and Schervish]{shalizi2011}
Daniel~J McDonald, Cosma~Rohilla Shalizi, and Mark Schervish.
\newblock Generalization error bounds for stationary autoregressive models.
\newblock \emph{arXiv preprint arXiv:1103.0942}, 2011.

\bibitem[Meinshausen and B{\"u}hlmann(2006)]{meinshausenbuhlmann06}
Nicolai Meinshausen and Peter B{\"u}hlmann.
\newblock High-dimensional graphs and variable selection with the {Lasso}.
\newblock \emph{The Annals of Statistics}, pages 1436--1462, 2006.

\bibitem[Meinshausen and Yu(2009)]{meinshausenyu09}
Nicolai Meinshausen and Bin Yu.
\newblock {Lasso}-type recovery of sparse representations for high-dimensional
  data.
\newblock \emph{The Annals of Statistics}, pages 246--270, 2009.

\bibitem[Michalski and Yashin(1986)]{yashin1986}
A.~Michalski and A.I. Yashin.
\newblock Structural minimization of risk on estimation of heterogeneity
  distributions.
\newblock 1986.
\newblock URL \url{http://pure.iiasa.ac.at/2785/1/WP-86-076.pdf}.

\bibitem[Mohri and Rostamizadeh(2009)]{mohri2009rademacher}
Mehryar Mohri and Afshin Rostamizadeh.
\newblock Rademacher complexity bounds for non-iid processes.
\newblock pages 1097--1104, 2009.

\bibitem[Newey and McFadden(1994)]{neweymcfadden94}
Whitney~K. Newey and Daniel McFadden.
\newblock Large sample estimation and hypothesis testing.
\newblock \emph{Handbook of econometrics}, 4:\penalty0 2111--2245, 1994.

\bibitem[Noor(2008)]{noor2008differentiable}
Muhammad~Aslam Noor.
\newblock Differentiable non-convex functions and general variational
  inequalities.
\newblock \emph{Applied Mathematics and Computation}, 199\penalty0
  (2):\penalty0 623--630, 2008.

\bibitem[Pearl(2015)]{pearl2015detecting}
Judea Pearl.
\newblock Detecting latent heterogeneity.
\newblock \emph{Sociological Methods \& Research}, page 0049124115600597, 2015.

\bibitem[Schwarz(1978)]{schwarz78}
Gideon~E. Schwarz.
\newblock Estimating the dimension of a model.
\newblock \emph{Annals of Statistics}, 6\penalty0 (2):\penalty0 461--464, 1978.

\bibitem[Shao(1997)]{shao97}
Jun Shao.
\newblock Asymptotic theory for model selection.
\newblock \emph{Statistica Sinica}, 7:\penalty0 221--242, 1997.

\bibitem[Skrondal and Rabe-Hesketh(2004)]{skrondal2004generalized}
Anders Skrondal and Sophia Rabe-Hesketh.
\newblock \emph{Generalized latent variable modeling: Multilevel, longitudinal,
  and structural equation models}.
\newblock Crc Press, 2004.

\bibitem[Smale and Zhou(2009)]{smale2009}
Steve Smale and Ding-Xuan Zhou.
\newblock Online learning with markov sampling.
\newblock \emph{Analysis and Applications}, 7\penalty0 (01):\penalty0 87--113,
  2009.

\bibitem[Stock and Watson(2012)]{stock2012generalized}
James~H Stock and Mark~W Watson.
\newblock Generalized shrinkage methods for forecasting using many predictors.
\newblock \emph{Journal of Business \& Economic Statistics}, 30\penalty0
  (4):\penalty0 481--493, 2012.

\bibitem[Stone(1974)]{stone74}
M.~Stone.
\newblock Cross-validatory choice and assessment of statistical predictions.
\newblock \emph{Journal of the Royal Statistical Society, Series B
  (Methodological)}, 36\penalty0 (2):\penalty0 111--147, 1974.

\bibitem[Stone(1977)]{stone77}
M.~Stone.
\newblock An asymptotic equivalence of choice of model by cross-validation and
  akaike's criterion.
\newblock \emph{Journal of the Royal Statistical Society, Series B
  (Methodological)}, 39\penalty0 (1):\penalty0 44--47, 1977.

\bibitem[Strongin and Sergeyev(2013)]{strongin2013global}
Roman~G Strongin and Yaroslav~D Sergeyev.
\newblock \emph{Global optimization with non-convex constraints: Sequential and
  parallel algorithms}, volume~45.
\newblock Springer Science \& Business Media, 2013.

\bibitem[Tang(2007)]{tang2007hoeffding}
Yongqiang Tang.
\newblock A hoeffding-type inequality for ergodic time series.
\newblock \emph{Journal of Theoretical Probability}, 20\penalty0 (2):\penalty0
  167--176, 2007.

\bibitem[Tibshirani(1996)]{tibshirani96}
Robert Tibshirani.
\newblock Regression shrinkage and selection via the {Lasso}.
\newblock \emph{Journal of the Royal Statistical Society, Series B
  (Methodological)}, 58:\penalty0 267--288, 1996.

\bibitem[Vapnik and Chervonenkis(1971{\natexlab{a}})]{vc71}
Vladimir~N. Vapnik and Alexey~Y. Chervonenkis.
\newblock On the uniform convergence of relative frequencies of events to their
  probabilities.
\newblock \emph{Theoretical Probability and its Applications}, 16\penalty0
  (2):\penalty0 264--280, 1971{\natexlab{a}}.

\bibitem[Vapnik and Chervonenkis(1971{\natexlab{b}})]{vc71b}
Vladimir~N. Vapnik and Alexey~Y. Chervonenkis.
\newblock Theory of uniform convergence of frequencie of appearance of
  attributes to their probabilities and problems of defining optimal solution
  by empiric data.
\newblock \emph{Avtomatika i Telemekhanika}, \penalty0 (2):\penalty0 42--53,
  1971{\natexlab{b}}.

\bibitem[Vapnik and Chervonenkis(1974{\natexlab{a}})]{vc74}
Vladimir~N. Vapnik and Alexey~Y. Chervonenkis.
\newblock The method of ordered risk minimization, {I}.
\newblock \emph{Avtomatika i Telemekhanika}, \penalty0 (8):\penalty0 21--30,
  1974{\natexlab{a}}.

\bibitem[Vapnik and Chervonenkis(1974{\natexlab{b}})]{vc74b}
Vladimir~N. Vapnik and Alexey~Y. Chervonenkis.
\newblock On the method of ordered risk minimization, {II}.
\newblock \emph{Avtomatika i Telemekhanika}, \penalty0 (9):\penalty0 29--39,
  1974{\natexlab{b}}.

\bibitem[Wang and Feng(2005)]{wang2005learning}
Li-Wei Wang and Ju-Fu Feng.
\newblock Learning gaussian mixture models by structural risk minimization.
\newblock In \emph{2005 International Conference on Machine Learning and
  Cybernetics}, volume~8, pages 4858--4863. IEEE, 2005.

\bibitem[Wellner(1981)]{wellner1981glivenko}
Jon~A Wellner.
\newblock A glivenko-cantelli theorem for empirical measures of independent but
  non-identically distributed random variables.
\newblock \emph{Stochastic Processes and their Applications}, 11\penalty0
  (3):\penalty0 309--312, 1981.

\bibitem[Xu et~al.(2016)Xu, Hong, and Fisher]{xuall16b}
Ning Xu, Jian Hong, and Timothy~C.G. Fisher.
\newblock Clustered model selection and clustered model averaging: Simultaneous
  heterogeneity control and modeling.
\newblock {S}chool of Economics, The University of Sydney, 2016.

\bibitem[Yan and Ma(2001)]{yan2001global}
Liexiang Yan and Dexian Ma.
\newblock Global optimization of non-convex nonlinear programs using line-up
  competition algorithm.
\newblock \emph{Computers \& Chemical Engineering}, 25\penalty0 (11):\penalty0
  1601--1610, 2001.

\bibitem[Yu(1994)]{yu1994rates}
Bin Yu.
\newblock Rates of convergence for empirical processes of stationary mixing
  sequences.
\newblock \emph{The Annals of Probability}, pages 94--116, 1994.

\bibitem[Yu and Joachims(2009)]{yu2009learning}
Chun-Nam~John Yu and Thorsten Joachims.
\newblock Learning structural svms with latent variables.
\newblock In \emph{Proceedings of the 26th annual international conference on
  machine learning}, pages 1169--1176. ACM, 2009.

\bibitem[Yu(1993)]{yu1993glivenko}
Hao Yu.
\newblock A glivenko-cantelli lemma and weak convergence for empirical
  processes of associated sequences.
\newblock \emph{Probability theory and related fields}, 95\penalty0
  (3):\penalty0 357--370, 1993.

\bibitem[Zhang(2010)]{zhang10}
Cun-Hui Zhang.
\newblock Nearly unbiased variable selection under minimax concave penalty.
\newblock \emph{The Annals of Statistics}, 38:\penalty0 894--942, 2010.

\bibitem[Zhang and Huang(2008)]{zhanghuang08}
Cun-Hui Zhang and Jian Huang.
\newblock The sparsity and bias of the {Lasso} selection in high-dimensional
  linear regression.
\newblock \emph{The Annals of Statistics}, 36:\penalty0 1567--1594, 2008.

\bibitem[Zhang(2009)]{zhang09}
Tong Zhang.
\newblock On the consistency of feature selection using greedy least squares
  regression.
\newblock \emph{Journal of Machine Learning Research}, 10:\penalty0 555--568,
  2009.

\bibitem[Zhao and Yu(2006)]{zhaoyu06}
Peng Zhao and Bin Yu.
\newblock On model selection consistency of {Lasso}.
\newblock \emph{The Journal of Machine Learning Research}, 7:\penalty0
  2541--2563, 2006.

\bibitem[Zou(2006)]{zou06}
Hui Zou.
\newblock The adaptive {Lasso} and its oracle properties.
\newblock \emph{Journal of the American statistical association}, 101\penalty0
  (476):\penalty0 1418--1429, 2006.

\end{thebibliography}

\newpage{}

\section*{Appendix 1}

\begin{proof}
\textbf{Theorem~\ref{thm2.1}} 
Since $b_{train}=\mathrm{argmin}_{b}\;\mathcal{R}_{n_t}\left(b\vert Y_t,X_t\right)$, eq.~{(\ref{eq:lem2.1})} forms an upper bound for the generalization error with probability $1-1/n_t$, $\forall b$,
\[
    \mathcal{R}(b_{train}|Y,X) \leqslant \mathcal{R}_{n_t}(b_{train}|Y_{t},X_{t})\left(1-\sqrt{\epsilon}\right)^{-1}
\]
where $\mathcal{R}_{n_t}(b_{train}|Y_{t},X_{t})$ stands for the training error on $(Y_{t},X_{t})$, $\mathcal{R}(b_{train}|Y,X)$ stands for the true population error of $b_{train}$ and $\epsilon=(1/n_t)\left\{h\ln\left[\left(n_t/h\right)\right]+h-\ln\left(1/n_t\right)\right\} $.

To use eq.~(\ref{eq:lem2.1}) to quantify the relation between eGE and eTE, we need to consider whether the loss function $Q(b_{train}\vert y,\mathbf{x})$ has no tail, a light tail, or a heavy tail.

\begin{description}

\item[\textbf{No tail.}] If the loss function $Q(\cdot)$ is bounded between $[0,B]$, where $B\in(0,\infty)$, then from Hoeffding's inequality \citep{hoeffding63} for the extremum estimator $b_{train}$, the empirical process satisfies, $\forall\varsigma\geqslant0$,
    \begin{equation}
     	\mathrm{P}\left\{\vert\mathcal{R}_{n_s}(b_{train}|X_{s},Y_{s})-\mathcal{R}(b_{train}|X,Y)\vert
	  	\leqslant\varsigma\right\}
		\geqslant
        1 - 2\exp\left(-\frac{2n\varsigma}{B}\right)
    \end{equation}	
	If we define $\varpi=1-2\exp(-2n^2\varsigma/\sum_{i=1}^{n}B_i)$, then	
	\begin{equation}
		\varsigma=\frac{B}{n}\ln\sqrt{\frac{2}{1-\varpi}}
	\end{equation}
	This implies, for any extremum estimator $b_{train}$	
	\begin{equation}
        \mathrm{P}\{\mathcal{R}_{n_s}(b_{train}|X_{s},Y_{s})\leqslant\mathcal{R}(b_{train}|X,Y)+\varsigma\} \geqslant
        \varpi.
	\end{equation}	
    Since both the training and test set are randomly sampled from the population, eq.~(\ref{eq:lem2.1}) can be modified as follows: $\forall\varsigma\geqslant0,\,\forall\tau_{1}\geqslant0,\,\exists\, N_{1}\in\mathbb{R}^{+}$ subject to	
	\begin{equation}
		\mathrm{P}\left\{\mathcal{R}_{n_s}\left(b_{train}|X_{s},Y_{s}\right) \leqslant
        \frac{\mathcal{R}_{n_t}(b_{train}|X_{t},Y_{t})}{1-\sqrt{\epsilon}}+
		\varsigma\right\}
		\geqslant
        \varpi\left(1-\frac{1}{n_t}\right)
	\end{equation}

\item[\textbf{Light tail.}] Suppose the loss function $Q(\cdot)$ is unbounded, but still $\mathcal{F}$-measurable, and possesses a finite $\nu$th moment when $\nu>2$. Based on Chebyshev's inequality for the extremum estimator $b_{train}$, the empirical process satisfies, $\forall\varsigma\geqslant0$,
    \begin{eqnarray}
     	\mathrm{P}\left\{\vert\mathcal{R}_{n_s}(b_{train}|X_{s},Y_{s})-\mathcal{R}(b_{train}|X,Y)\vert
	  	\leqslant\varsigma\right\}
		& \geqslant &
        1 - \frac{\mathrm{var}(Q(b_{train}\vert y,\mathbf{x}))}{n\varsigma}
    \end{eqnarray}	
	If we define $\varpi = 1 - \mathrm{var}(Q(b_{train}\vert y,\mathbf{x}))/n\varsigma$, then	
	\begin{equation}
		\varsigma=\frac{\mathrm{var}(Q(b_{train}\vert y,\mathbf{x}))}
		{n(1-\varpi)}
	\end{equation}
	This implies, for any extremum estimator $b_{train}$	
	\begin{equation}
        \mathrm{P}\{\mathcal{R}_{n_s}(b_{train}|X_{s},Y_{s})\leqslant\mathcal{R}(b_{train}|X,Y)+\varsigma\} 		\geqslant
        \varpi.
	\end{equation}	
    Since both the training and test set are randomly sampled from the population, eq.~(\ref{eq:lem2.1}) can be modified as follows: $\forall\varsigma\geqslant0,\,\forall\tau_{1}\geqslant0,\,\exists\, N_{1}\in\mathbb{R}^{+}$ subject to	
	\begin{equation}
		\mathrm{P}\left\{\mathcal{R}_{n_s}\left(b_{train}|X_{s},Y_{s}\right) \leqslant\frac{\mathcal{R}
		_{n_t}(b_{train}|X_{t},Y_{t})}{1-\sqrt{\epsilon}}+
		\varsigma\right\}
		\geqslant \varpi\left(1-\frac{1}{n_t}\right)
	\end{equation}

\item[\textbf{Heavy tail.}] Suppose the loss function $Q(\cdot)$ is unbounded, but still $\mathcal{F}$-measurable, and has heavy tails with the property that, for $1<\nu\leqslant 2$, $\exists\tau$, such that
	\begin{equation}
		\sup
		\frac{\sqrt[\nu]{\int[Q(b_{train}\vert y,\mathbf{x})]^\nu dF\left(y,\mathbf{x}\right)}}
		{\int Q(b_{train}\vert y,\mathbf{x})dF\left(y,\mathbf{x}\right)}
		\leqslant
        \tau.
	\end{equation}	
    Based on the Bahr-Esseen inequality for the extremum estimator $b_{train}$, the empirical process satisfies, $\forall\varsigma\geqslant0$,	
	\begin{eqnarray}
        \mathrm{P}\{\vert\mathcal{R}(b_{train}|Y,X)-\mathcal{R}_{n_s}(b_{train}|Y_{s},X_{s})\vert
        \leqslant\varsigma\}
		& \geqslant & 1-2\cdot \frac{\mathbb{E}\left[Q(b_{train}\vert y,\mathbf{x})^\nu\right]}
		{\varsigma^p\cdot n_s^{\nu-1}} \notag \\
		& \geqslant & 1-2\tau^\nu\cdot \frac{\left(\mathbb{E}\left[Q(b_{train}\vert y,\mathbf{x})\right]
		\right)^\nu}{\varsigma^\nu\cdot n_s^{\nu-1}}
	\end{eqnarray}	
	If we define
	\begin{eqnarray}
        \varpi=1-2\tau^\nu\cdot\frac{\left(\mathbb{E}\left[Q(b_{train}\vert y,\mathbf{x})\right]\right)^\nu}
        {\varsigma^\nu\cdot n_s^{\nu-1}}
	\end{eqnarray}	
  	then  	
  	\begin{equation}
        \varsigma=\frac{\sqrt[\nu]{2}\cdot\tau\left(\mathbb{E}
        \left[Q(b_{train}\vert y,\mathbf{x})\right]\right)}
        {\sqrt[\nu]{1-\varpi}\cdot n_s^{1-1/{\nu}}}
	\end{equation}
	This implies, for any extremum estimator $b_{train}$	
	\begin{equation}
        \mathrm{P}\{\mathcal{R}_{n_s}(b_{train}|X_{s},Y_{s})
        \leqslant\mathcal{R}(b_{train}|X,Y)+\varsigma\}
        \geqslant
        \varpi.
	\end{equation}	
    Since both the training and test set are randomly sampled from the population, eq.~(\ref{eq:lem2.1}) could be modifed and relaxed as follows: $\forall\varsigma\geqslant0,\,\forall\tau_{1}\geqslant0,
    \,\exists\, N_{1}\in\mathbb{R}^{+}$ subject to	
	\begin{equation}
		\mathrm{P}\left\{\mathcal{R}_{n_s}\left(b_{train}|X_{s},Y_{s}\right) \leqslant\frac{\mathcal{R}
		_{n_t}(b_{train}|X_{t},Y_{t})}{1-\sqrt{\epsilon}}+
		\varsigma\right\}
		\geqslant \varpi\left(1-\frac{1}{n_t}\right)
	\end{equation}	
\end{description}
\end{proof}

\begin{proof}
\textbf{Theorem~\ref{thm2.2}}
The upper bound of eGE for cross-validation can be established by adapting eq.~(\ref{eq:lem2.1}) and (\ref{eq:thm2.1}). The proof is quite similar to Theorem~\ref{thm2.1}, except for the fact eq.~(\ref{eq:lem2.1}) and (\ref{eq:thm2.1}) measure the upper bound by one-round eTE while eq.~(\ref{eq:thm2.2}) uses averaged multiple-round eTE,
\[
	\frac{1}{K}\sum_{q=1}^{K}\mathcal{R}_{n_t}(b_{train}|Y_{t}^q,X_{t}^q).
\]
Hence, eq.~(\ref{eq:lem2.1}) can be generalized as follows
\begin{eqnarray}
	\mathcal{R}(b_{train}|Y,X)
	& = & \mathbb{E}[Q(b_{train}|y,\mathbf{x})] \notag \\
	& = & \mathbb{E}[\frac{1}{n_t}\sum_i Q(b_{train}|y_{t}^i,\mathbf{x}_{t}^i)]\notag \\
	& = & \mathbb{E}[\mathcal{R}_{n_t}(b_{train}|Y_{t},X_{t})]\notag \\
	& \leqslant &
	\frac{\frac{1}{K}\sum_{q=1}^{K}\mathcal{R}_{n_t}(b_{train}|Y_{t}^q,X_{t}^q)}{1-\sqrt{\epsilon}}.
\end{eqnarray}	
Also, since	
\begin{eqnarray}
	\mathcal{R}(b_{train}|Y,X)
	& = & \mathbb{E}[Q(b_{train}|y,\mathbf{x})] \notag \\
	& = & \mathbb{E}[\frac{1}{n_s}\sum_{i=1}^{n_s}Q(b_{train}|y_{s}^i,\mathbf{x}_{s}^i)] \notag	\\
	& = & \mathbb{E}[\mathcal{R}_{n_s}(b_{train}|Y_{s},X_{s})] \notag \\
	& = & \mathbb{E}[\frac{1}{K}\sum_{q=1}^{K} \mathcal{R}_{n_s}(b_{train}|	Y_{s}^q,X_{s}^q)],
\end{eqnarray}	
we can derive the following inequality
\begin{eqnarray}
	&   & \mathrm{P}\{\vert \mathbb{E}[\frac{1}{K}\sum_{q=1}^{K}
    \mathcal{R}_{n_s}(b_{train}|Y_{s},X_{s})] -
	\frac{1}{K}\sum_{q=1}^{K}
    \mathcal{R}_{n_s}(b_{train}|Y_{s},X_{s}) \vert \leqslant \varsigma_{cv}\} \notag \\
	& = & \mathrm{P}\{\vert \mathcal{R}(b_{train}|Y,X) -
	\frac{1}{K}\sum_{q=1}^{K}
    \mathcal{R}_{n_s}(b_{train}|Y_{s},X_{s}) \vert	\leqslant \varsigma_{cv}\} \notag \\
	& \geqslant & \varpi
\end{eqnarray}	
where, similar to the proof of Theorem~\ref{thm2.1},
\begin{equation}
	\varsigma_{cv} =
	\left\{
		\begin{array}{ll}
			(\sqrt[\nu]{2} \cdot \tau \cdot \mathcal{R}(b_{train}|Y,X) /
                   (\sqrt[\nu]{1-\varpi}\cdot n_s^{1-1/{\nu}})
			& \mbox{if } \nu \in(1,2] \\
			(B\ln\sqrt{2/(1-\varpi)})/n_s
            & \mbox{if } Q(\cdot) \in(0,B] \mbox{ and B is bounded} \\
			\mathrm{var}( Q(b_{train}\vert y, \mathbf{x})) / ((n_s)^2(1-\varpi))
			& \mbox{if } \nu\in(2,\infty)
        \end{array}
    \right.	
\end{equation}	
As a result, we have
\begin{equation}
	\frac{1}{K}\sum_{j=1}^{K}\mathcal{R}_{n_s}(b_{train}|Y_{s}^j,X_{s}^j)
	\leqslant
	\frac{\frac{1}{K}\sum_{q=1}^{K}\mathcal{R}_{n_t}(b_{train}|Y_{t}^q,X_{t}^q)}
	{(1-\sqrt{\epsilon})} + \varsigma_{cv}
\end{equation}
\end{proof}

\begin{proof}
\textbf{Proposition~\ref{prop3.1}}
Given A1--A6, the true DGP is
\[
	y_{i} = x_{i}^{T}\beta + u_{i}, \quad i=1,\ldots,n.
\]
Proving that the true DGP has the lowest eGE is equivalent to proving, in a test set, that
\begin{equation}
    \frac{\sum_{i=1}^n\left(y_{i}-x_{i}^{T}\beta\right)^{2}}{n}
    \leqslant
    \frac{\sum_{i=1}^n\left(y_{i}-x_{i}^{T}b\right)^{2}}{n},
    \label{a1p1}
\end{equation}
which is equivalent to proving that
\begin{align*}
	0 & \leqslant \frac{1}{n} \sum_{i=1}^n\left[\left(y_{i}-x_{i}^{T}b\right)^{2}-
           \left(y_{i}-x_{i}^{T}\beta\right)^{2}\right] \\
	\iff
	0 & \leqslant \frac{1}{n}
	  \sum_{i=1}^n\left(y_{i}-x_{i}^{T}b+y_{i}-x_{i}^{T}\beta\right)
	\left(y_{i}-x_{i}^{T}b-y_{i}+x_{i}^{T}\beta\right) \\
	\iff
	0 & \leqslant \frac{1}{n}
	  \sum_{i=1}^n\left(y_{i}-x_{i}^{T}b+y_{i}-x_{i}^{T}\beta\right)
	  \left(x_{i}^{T}\beta-x_{i}^{T}b\right).
\end{align*}
Defining $\delta=\beta-b$, it follows,
\begin{align*}
	0      & \leqslant \frac{1}{n}
	\sum_{i=1}^n\left(2y_{i}-x_{i}^{T}b-x_{i}^{T}\beta\right)
	\left(x_{i}^{T}\delta\right) \\
	\iff 0 & \leqslant \frac{1}{n}
    \sum_{i=1}^n\left(2y_{i}-x_{i}^{T}\beta+x_{i}^{T}\beta-x_{i}^{T}b-x_{i}^{T}\beta\right)
	\left(x_{i}^{T}\delta\right) \\
	\iff 0 & \leqslant \frac{1}{n}
	\sum_{i=1}^n\left(2y_{i}-2x_{i}^{T}\beta+x_{i}^{T}\delta\right)
	\left(x_{i}^{T}\delta\right) \\
	\iff 0 & \leqslant \frac{1}{n}
	\sum_{i=1}^n\left(2u_{i}+x_{i}^{T}\delta\right)\left(x_{i}^{T}\delta\right)
\end{align*}
Hence, proving Proposition~\ref{prop3.1} is equivalent to proving
\[
	0 \leqslant \frac{1}{n} \sum_{i=1}^n\left(2u_{i}+x_{i}^{T}\delta\right)\left(x_{i}^{T}\delta\right)
\]
Since $\mathbb{E}(X^T u)=\mathbf{0}$ (A2), it follows that
\[
 	\frac{1}{n}\sum_{i=1}^n u_{i}\cdot x_{i}\overset{\mathbf{P}}{\rightarrow}\mathbf{0}
	\iff
	\frac{1}{n}\sum_{i=1}^n\left(u_{i}\cdot x_{i}^{T}\right)\beta\overset{\mathbf{P}}{\rightarrow}0
	\quad\mbox{and}\quad
	\frac{1}{n}\stackrel[i=1]{n}{\sum}\left(u_{i}\cdot x_{i}^{T}\right)b\rightarrow 0
\]
Hence, asymptotically
\[
	\frac{1}{n}\sum_{i=1}^n\left(2u_{i}+x_{i}^{T}\delta\right)\left(x_{i}^{T}\delta\right)=
	\frac{1}{n}\sum_{i=1}^n 2\delta u_{i}x_{i}^{T}+
	\frac{1}{n}\sum_{i=1}^n\left(x_{i}^{T}\delta\right)^{2}
	\overset{\mathbf{P}}{\rightarrow}\mathbb{E}\left(x_{i}^{T}\delta\right)^{2}\geqslant 0
\]
\end{proof}

\bigskip
\begin{proof}
\textbf{Proposition~\ref{prop3.2}}
The proof of Proposition~\ref{prop3.2} is very straightforward. When $\lambda = 0$, $b_{\lambda} = b_{OLS}$. Hence, as $n\rightarrow \infty$, $b_{\lambda=0} = b_{OLS} \overset{\mathcal{L}_2}{\rightarrow} \beta$. Hence, there exists at least one $\lambda$ that guarantees the $\mathcal{L}_2$ consistency. This guarantees that when $n\rightarrow\infty$, $\beta \in \{b_{\lambda}\}$, or the true DGP is in the list of alternative $b_{\lambda}$.
\end{proof}


\bigskip
\begin{proof}
\textbf{Lemma~\ref{lem3.1}}
Eq.~(\ref{eq1:thm3.1}) and (\ref{eq2:thm3.1}) are the direct application of eq.~(\ref{eq:thm2.1}) and (\ref{eq:thm2.2}). Thus, we only need to focus on the last term of the RHS, $\varsigma$ and $\varsigma_{cv}$.

Since the error term $u$ in classical regression analysis is distributed as a zero-mean Gaussian, the loss function of OLS
\[
	Q(b_{OLS}) \sim \mathrm{var(u)} \cdot \chi^2(1).
\]
Hence in eq.~(\ref{eq1:thm3.1}) the last RHS term
\[
	\varsigma = \frac{2(\mathrm{var}(u))^2}{n_s\sqrt{1-\varpi}}
\]
Furthermore, in cross-validation,
\[
	\mathcal{R}_{n_s}(b_{OLS}\vert Y,X) \sim \frac{\mathrm{var(u)}}{n(K-1)/K} \cdot \chi^2(n(K-1)/K).
\]
Hence, in eq.~(\ref{eq2:thm3.1}) the last RHS term
\[
	\varsigma_{cv} = \frac{2(\mathrm{var}(u))^2}{\sqrt{1-\varpi}(n/K)^2}
\]
By substituting the above values for $\varsigma$ and $ \varsigma_{cv}$ into eq.~(\ref{eq:thm2.1}) and (\ref{eq:thm2.2}), we have eq.~(\ref{eq1:thm3.1}) and (\ref{eq2:thm3.1}).
\end{proof}

\bigskip
\begin{proof}
\textbf{Corollary~\ref{cor3.1}}
The optimal $K$ or $n_t/n_s$ can be obtained by finding the smallest expectation of the RHS for eq.~(\ref{eq1:lem3.1}) and (\ref{eq2:lem3.1}).

Since the error term $u$ in classical regression analysis is distributed as a zero-mean Gaussian, the loss function of OLS
\[
	Q(b_{OLS}) \sim \mathrm{var(u)} \cdot \chi^2(1).
\]
As a result,
\[
	\mathcal{R}_{n_s}(b_{OLS}\vert Y,X) \sim \frac{\mathrm{var(u)}}{n(K-1)/K} \cdot \chi^2(n(K-1)/K).
\]
Hence in the  RHS of eq.~(\ref{eq1:thm3.1}),
\[
	\frac{1}{K}\sum_{j=1}^{K}\mathcal{R}_{n_t}(b_{train}|Y_{t}^j,X_{t}^j) 	
	\sim
	 \frac{\mathrm{var}(u)}{n(K-1)/K} \cdot \mathrm{Gamma}(n(K-1)/2,2/K)
\]
Hence, the expectation of RHS for eq.~(\ref{eq2:thm3.1}) is equal to
\[
	\frac{\mathrm{var}(u)}{1-\sqrt{\epsilon}} + \frac{4(\mathrm{var}(u))^2}{\sqrt{1-\varpi}(n/K)^2}
\]
and
\[
	K^{*} = \argmin_{K}\frac{\mathrm{var}(u)}{1-\sqrt{\epsilon}} +
    \frac{4(\mathrm{var}(u))^2}{\sqrt{1-\varpi}(n/K)^2}
\]
\end{proof}

\bigskip
\begin{proof}
\textbf{Proposition~\ref{prop3.3}}
In the proof, $b_{OLS}$ is the OLS estimate learned from the training set $(Y_t,X_t)$ in validation and $b_{OLS}^q$ is the OLS estimate learned from the $q$th training set $(Y_t^q,X_t^q)$ in cross-validation.

\begin{description}
	
    \item[\textbf{Validation.}] As shown in Lemma~\ref{lem3.1}, eq.~(\ref{eq1:lem3.1}) holds with at least probability $\varpi(1-1/n_t)$,
		\begin{equation}
			\mathcal{R}_{n_s}(b_{OLS}|Y_{s},X_{s}) \leqslant
			\frac{\mathcal{R}_{n_t}(b_{OLS}|Y_{t},X_{t})}{1-\sqrt{\epsilon}}
			+ \varsigma	
		\end{equation}
        Also, the validation algorithm guarantees that, among all the $b\in\{ b_{\lambda}\}$, $b_{Lasso}$ has the lowest eGE on the test set,
		\begin{equation}
			\mathcal{R}_{n_s}(b_{pen}|Y_{s},X_{s})
			\leqslant
			\mathcal{R}_{n_s}(b_{OLS}|Y_{s},X_{s})
		\end{equation}
		we have
		\begin{equation}
			\frac{1}{n_s}\left\Vert Y_{s}-X_{s}b_{pen}\right\Vert _{2}^{2} \leqslant
			\frac{\frac{1}{n_t}(\Vert Y_{t}-X_{t}b_{OLS}\Vert_{2})^{2}}
			{1-\sqrt{\epsilon}} + \varsigma
		\end{equation}
		By defining $\Delta=b_{OLS}-b_{pen}$, $Y_{t}-X_{t}b_{OLS}=e_{t}$ and $Y_{s}-X_{s}b_{OLS}=e_{s}$,
		\begin{eqnarray}
		\frac{1}{n_s}(\Vert Y_{s}-X_{s}b_{pen}\Vert_{2})^{2}
		  & = & \frac{1}{n_s}(\Vert Y_{s}-X_{s}b_{OLS}+X_{s}\Delta\Vert_{2})^{2}
		  \notag  \\
		  & = & \frac{1}{n_s}(\Vert e_{s}+X_{s}\Delta\Vert_{2})^{2}
		  \notag     \\
		  & = & \frac{1}{n_s}(e_{s}+X_{s}\Delta)^{T}(e_{s}+X_{s}\Delta)
		  \notag  \\
		  & = & \frac{1}{n_s}(\left\Vert e_{s}\right\Vert_{2}^{2}+2e_{s}^{T}X_{s}\Delta+
			\Delta^{T}X_{s}^{T}X_{s}\Delta)
		\end{eqnarray}
		Hence,
		\begin{equation}
			\frac{1}{n_s} (\Vert Y_{s}-X_{s}b_{pen}\Vert_{2})^{2}
			\leqslant
			\frac{\frac{1}{n_t}(\Vert Y_{t}-X_{t}b_{OLS} \Vert_{2})^{2}}{1-\sqrt{\epsilon}}
			+ \varsigma
		\end{equation}
		implies
		\begin{equation}
			\frac{1}{n_s}(\Vert e_{s} \Vert_{2})^{2}
			+ \frac{2}{n_s}e_{s}^{T}X_{s}\Delta
			+ \frac{1}{n_s}\Delta^{T}X_{s}^{T}X_{s}\Delta
			\leqslant
			\frac{\frac{1}{n_t} (\Vert e_{t} \Vert_{2})^{2}}{1-\sqrt{\epsilon}}
			+ \varsigma.
		\end{equation}
		It follows that
		\begin{equation}
			\frac{1}{n_s} (\Vert X_{s}\Delta \Vert_{2})^{2}
			\leqslant
			\left( \frac{1}{n_t}\frac{ ( \Vert e_{t} \Vert_{2})^{2}}{1-\sqrt{\epsilon}}-
			\frac{1}{n_s} ( \Vert e_{s} \Vert_{2})^{2} \right)-
			\frac{2}{n_s}e_{s}^{T}X_{s}\Delta
			+\varsigma.
		\end{equation}
		By the Holder inequality,
		\begin{equation}
			-e_{s}^{T} X_{s} \Delta
			\leqslant
			\vert e_{s}^{T} X_{s} \Delta \vert
			\leqslant
			\Vert e_{s}^{T} X_{s} \Vert_{\infty}
			\Vert \Delta \Vert_{1}.
		\end{equation}
		It follows that
		\begin{equation}
			\frac{1}{n_s} (\Vert X_{s} \Delta \Vert_{2})^{2}
			\leqslant
			\left( \frac{1}{n_t} \frac{ (\Vert e_{t} \Vert_{2})^{2}}{1-\sqrt{\epsilon}}
			- \frac{1}{n_s} (\Vert e_{s} \Vert_{2})^{2} \right)
			+ \frac{2}{n_s} \Vert e_{s}^{T} X_{s} \Vert_{\infty} \Vert \Delta \Vert _{1}
			+\varsigma.
		\end{equation}
		Also, since $\Vert b_{pen}\Vert_{1} \leqslant \Vert b_{OLS} \Vert_{1}$
		\begin{eqnarray}
			\Vert \Delta \Vert _{1}
			& = & \Vert b_{OLS}-b_{pen} \Vert_{1}
			\notag     \\
			& \leqslant &  \Vert b_{pen} \Vert_{1} + \Vert b_{OLS} \Vert_{1}
			\notag  \\
			& \leqslant & 2 \Vert b_{OLS} \Vert_{1}		
		\end{eqnarray}
		As a result, we have
		\begin{equation}
			\frac{1}{n_s} (\Vert X_{s}\Delta \Vert _{2})^{2}
			\leqslant
			\left( \frac{1}{n_t} \frac{ ( \Vert e_{t} \Vert_{2})^{2}}{1-\sqrt{\epsilon}}
			- \frac{1}{n_s} ( \Vert e_{s} \Vert_{2})^{2} \right)
			+ \frac{4}{n_s} \Vert e_{s}^{T }X_{s} \Vert_{\infty} \Vert b_{OLS} \Vert_{1}
			+\varsigma
		\end{equation}

    \item[\textbf{$K$-fold cross validation.}] If penalized regression is implemented by $K$-fold cross-validation, then based on Lemma~\ref{lem3.1}, the following bound holds with probability at least $(1-1/K)\varpi$
		\begin{equation}
			\frac{1}{K} \sum_{q=1}^{K} \mathcal{R}_{n_s}(b_{OLS}^q|X_s^q,Y_s^q)
			\leqslant
			\frac{1}{K} \sum_{q=1}^{K}
            \frac{\mathcal{R}_{n_t}(b_{OLS}^q|X_t^q,Y_t^q)}{1-\sqrt{\epsilon}}
			+ \varsigma_{cv}.
		\end{equation}
        Since $b_{pen}$ minimizes $(1/K)\sum_{q=1}^K\mathcal{R}_{n_s}(b|X_s^q,Y_s^q)$ among $\{b_{\lambda}\}$,
		\begin{eqnarray}
			\frac{1}{K}\sum_{q=1}^K\mathcal{R}_{n_s}(b_{pen}^q|X_s^q,Y_s^q)
			\leqslant
			\frac{1}{K}\sum_{q=1}^K\mathcal{R}_{n_s}(b_{OLS}^q|X_s^q,Y_s^q),
		\end{eqnarray}
	    it follows that
		\begin{equation}
			\frac{1}{K}\sum_{q=1}^K\mathcal{R}_{n_s}(b_{pen}^q|X_s^q,Y_s^q)
			\leqslant
			\frac{\mathcal{R}_{n_t}(b_{OLS}^q|X_t^{q},Y_t^{q})}{1-\sqrt{\epsilon}}
			+\varsigma_{cv}.
		\end{equation}
        By defining $\Delta^q = b_{OLS}^q - b_{pen}^q$ and $e_s^q = Y_s^q-X_s^q b_{OLS}^q$ we have
		\begin{eqnarray}
			\frac{1}{n_s} (\Vert Y_s^q - X_s^q b_{pen}^q \Vert_{2})^{2}
		  	& = & \frac{1}{n_s}
            ( \Vert Y_s^q - X_s^q b_{OLS}^q + X_s^q \Delta^q \Vert_{2}) ^{2} \notag \\
		  	& = & \frac{1}{n_s} ( \Vert e_s^q + X_s^q \Delta^q \Vert_{2})^{2} \notag \\
		  	& = & \frac{1}{n_s} ( e_s^q + X_s^q \Delta^q )^{T} (e_s^q + X_s^q \Delta^q) \notag \\
		  	& = & \frac{1}{n_s} \left[ (\Vert e_s^q \Vert_{2})^{2} + 2(e_s^q)^{T} X_s^q \Delta^q +
		  	(\Delta^q)^{T} (X_s^q)^{T} X_s^q \Delta^q \right].
		\end{eqnarray}
	    Hence,
		\begin{equation}
			\frac{1}{K}\sum_{q=1}^K \left( \frac{1}{n_s}
            (\Vert Y_s^q - X_s^q b_{pen} \Vert_{2})^{2}\right)
			\leqslant
			\frac{1}{n_t} \frac{\left\Vert Y_t^{q}-X_t^{q}b_{OLS}^q\right\Vert_{2}^{2}}
			{1-\sqrt{\epsilon}}
			+ \varsigma_{cv}
		\end{equation}
	    implies
		\begin{eqnarray}
			 \frac{1}{K}\sum_{q=1}^K \frac{1}{n_s} (\Vert e_s^q \Vert_{2})^{2}
			+\frac{1}{K}\sum_{q=1}^K \frac{2}{n_s} (e_s^q)^{T} X_{s} \Delta
			& + &
			\frac{1}{K}\sum_{q=1}^K \frac{1}{n_s}
            (\Delta^q)^{T} (X_s^q)^{T} (X_s^q) \Delta^q \notag \\
			& \leqslant &
			\frac{1}{K}\sum_{q=1}^K \frac{\frac{1}{n_t} (\Vert e_t^q \Vert_{2})^{2}}
			{1-\sqrt{\epsilon}} + \varsigma_{cv}.
		\end{eqnarray}
	    It follows that
		\begin{eqnarray}
			\frac{1}{K}\sum_{q=1}^K\frac{1}{n_s}\left\Vert X_s^q\Delta\right\Vert_{2}^{2}
			 \leqslant
            &\frac{1}{K}\sum_{q=1}^K \frac{\frac{1}{n_t} (\Vert e_t^q \Vert_{2})^{2}}{1-\sqrt{\epsilon}}
			 -\frac{1}{K}\sum_{q=1}^K\frac{\left\Vert e_s^q \right\Vert_{2}^{2}}{n_s} \notag \\			
			&-\frac{1}{K}\sum_{q=1}^K\frac{2}{n_s}\left(e_s^q\right)^{T}X_{s}^q\Delta^q
			+\varsigma_{cv}.
		\end{eqnarray}
	    By the Holder inequality,
		\begin{equation}
			-1\cdot (e_s^q)^{T} X_s^q \Delta^q
			\leqslant
			\vert ( e_s^q )^{T} X_s^q \Delta^q \vert
			\leqslant \Vert (e_s^q)^{T} X_s^q \Vert_{\infty} \Vert \Delta^q \Vert_{1}.
		\end{equation}
	    It follows that
		\begin{eqnarray}
			\frac{1}{K}\sum_{q=1}^K \frac{1}{n_s} \left(\Vert X_s^q \Delta \Vert_{2}\right)^{2}
			\leqslant
			& \left\vert \frac{1}{K}\sum_{q=1}^K \frac{\frac{1}{n_t}
            (\Vert e_t^q \Vert_{2})^{2}}{1-\sqrt{\epsilon}}
			- \frac{1}{K}\sum_{q=1}^K \frac{ (\Vert e_s^q \Vert_{2})^{2}}{n_s} \right\vert \notag \\
			& +\frac{1}{K}\sum_{q=1}^K \frac{2}{n_s} \Vert (e_s^q)^{T} X_{s}^q \Vert_{\infty}
			\Vert \Delta^q \Vert_{1}
			+ \varsigma_{cv}.
		\end{eqnarray}
	    Also, since $\Vert b_{pen}\Vert_{1} \leqslant \Vert b_{OLS} \Vert_{1}$
		\begin{eqnarray}
			\Vert \Delta^q \Vert _{1}
			& = & \Vert b_{OLS}^q -b_{pen}^q \Vert_{1}   \notag   \\
			& \leqslant & \Vert b_{pen}^q \Vert_{1} + \Vert b_{OLS}^q \Vert_{1} \notag  \\
			& \leqslant & 2\Vert b_{OLS}^q \Vert_{1}
		\end{eqnarray}
	    Therefore, we have
		\begin{eqnarray}
			\frac{1}{K}\sum_{q=1}^K \frac{1}{n_s} \left(\Vert X_s^q \Delta \Vert_{2}\right)^{2}
			\leqslant
			& \left\vert \frac{1}{K}\sum_{q=1}^K \frac{\frac{1}{n_t}
            (\Vert e_t^q \Vert_{2})^{2}}{1-\sqrt{\epsilon}}
			- \frac{1}{K}\sum_{q=1}^K \frac{ (\Vert e_s^q \Vert_{2})^{2}}{n_s} \right\vert \notag \\
			& +\frac{1}{K}\sum_{q=1}^K \frac{4}{n_s} \Vert (e_s^q)^{T} X_{s}^q \Vert_{\infty}
			\Vert b_{OLS}^q \Vert_{1}
			+ \varsigma_{cv}.
		\end{eqnarray}

\end{description}
	
\end{proof}
\bigskip{}

\begin{proof}
\textbf{Theorem~\ref{thm3.1}}
The proof follows from Proposition~\ref{prop3.3}.

\begin{description}
	\item[\textbf{Validation.}] For OLS, $(1/n)\left\Vert X_{s}\Delta\right
       \Vert _{2}^{2}\geqslant\rho\left\Vert \Delta\right\Vert _{2}^{2}$, where $\rho$ is the minimal eigenvalue for $(X_s)^{T}X_s$. Hence,
	   \begin{eqnarray}
		  \rho (\Vert \Delta \Vert_{2})^{2}
		  & \leqslant &
		  \frac{1}{n_s} (\Vert X_{s} \Delta \Vert _{2})^{2} \notag \\
		  & \leqslant &
            \left\vert \frac{1}{n_t} \frac{ (\Vert e_t \Vert_{2})^{2}}{1-\sqrt{\epsilon}}
            - \frac{ \Vert e_s \Vert_{2}^{2}}{n_s} \right\vert \notag \\
		  &           &
            +  \frac{4}{n_s} \Vert (e_s)^{T} X_{s} \Vert_{\infty} \Vert b_{OLS} \Vert_{1} + \varsigma.
	   \end{eqnarray}
	   By the Minkowski inequality, the above can be simplified to
	   \begin{eqnarray}
		  \left\Vert b_{train}-b_{Lasso}\right\Vert _{2}
		  & \leqslant &		
		  \sqrt{\left\vert \frac{1}{n_t \rho} \frac{ (\Vert e_t \Vert_{2})^{2}}{1-\sqrt{\epsilon}}
		  - \frac{ \Vert e_s \Vert_{2}^{2}}{n_s \rho} \right\vert} \notag \\
		  & & +  \sqrt{\frac{4}{n_s \rho} \Vert (e_s)^{T} X_{s} \Vert_{\infty}\Vert b_{OLS} \Vert_{1}}
		  + \sqrt{\frac{\varsigma}{\rho}}.
	   \end{eqnarray}	
	
	\item[\textbf{$K$-fold cross validation.}] For the OLS estimate from the $q$th round,
	   $(1/n)\left\Vert X_{s}^q \Delta\right\Vert _{2}^{2}\geqslant\rho
        \left\Vert \Delta\right\Vert _{2}^{2}$,	where $\rho_q$ is the minimal eigenvalue for $(X_s^q)^{T}X_s^q$ in the $q$th round. Hence, if we define the minimum of all the minimal round-by-round eigenvalues from all $K$ rounds,
	   \[
		  \bar{\rho} = \min\{\rho_q \vert \forall q \in [1,K] \},
	   \]
	   then
	   \begin{eqnarray}
		  \frac{1}{K}\sum_{q=1}^K \bar{\rho} (\Vert \Delta^q \Vert_{2})^{2}
		  & \leqslant &
		  \frac{1}{K}\sum_{q=1}^K \frac{1}{n_s} (\Vert X_{s}^q \Delta^q \Vert _{2})^{2} \notag \\
		  & \leqslant & \left\vert \frac{1}{K}\sum_{q=1}^K \frac{\frac{1}{n_t}
            (\Vert e_t^q \Vert_{2})^{2}}{1-\sqrt{\epsilon}}
		- \frac{1}{K}\sum_{q=1}^K \frac{ (\Vert e_s^q \Vert_{2})^{2}}{n_s} \right\vert \notag \\
		& & +\frac{1}{K}\sum_{q=1}^K \frac{4}{n_s} \Vert (e_s^q)^{T} X_{s}^q \Vert_{\infty}
		\Vert b_{OLS}^q \Vert_{1} + \varsigma_{cv}.
	   \end{eqnarray}
	   Hence,
	   \begin{eqnarray}
		  \frac{1}{K}\sum_{q=1}^K \left(\left\Vert b_{OLS}^q-b_{pen}^q\right\Vert _{2}\right)^2
		  & \leqslant &
		  \frac{1}{K}\sum_{q=1}^K \left|\frac{1}{\bar{\rho} n_t}
          \frac{\left\Vert e_{t}^q \right\Vert_{2}^{2}}{\left(1-\sqrt{\epsilon}\right)}
		-\frac{1}{K}\sum_{q=1}^K \frac{1}{\bar{\rho} n_s}
        \left\Vert e_{s}^q \right\Vert _{2}^{2}\right| \notag \\
		& &
		+\frac{1}{K}\sum_{q=1}^K \frac{4}{\bar{\rho} n_s}
        \left\Vert (e_{s}^q)^{T}X_{s} \right\Vert_{\infty}
		\left\Vert b_{OLS}^q\right\Vert _{1} +\frac{\varsigma}{\bar{\rho}}
	\end{eqnarray}
	
\end{description}
\end{proof}

\bigskip{}
\begin{proof}
\textbf{Corollary~\ref{cor3.2}}
($\mathcal{L}_2$ consistency of  $b_{pen}$)
	
\begin{description}
	
		\item[\textbf{Validation.}] In Theorem~\ref{thm3.1},
			\begin{eqnarray}
				\left\Vert b_{train}-b_{Lasso}\right\Vert _{2}
				& \leqslant &		
				\sqrt{\left\vert \frac{1}{n_t \rho} \frac{ (\Vert e_t \Vert_{2})^{2}}{1-\sqrt{\epsilon}}
				- \frac{ \Vert e_s \Vert_{2}^{2}}{n_s \rho} \right\vert} \notag \\
				& & +  \sqrt{\frac{4}{n_s \rho} \Vert (e_s)^{T} X_{s} \Vert_{\infty}
				\Vert b_{OLS} \Vert_{1}} + \sqrt{\frac{\varsigma}{\rho}}.
			\end{eqnarray}	
		Since
		\[
			\lim_{\tilde{n}/p \rightarrow \infty}
			\frac{1}{n_t} \frac{(\Vert e_t \Vert_{2})^{2}}{1-\sqrt{\epsilon}}
			=
			\lim_{\tilde{n}/p \rightarrow \infty}
            \frac{(\Vert e_s \Vert_{2})^{2}}{n_s}
			=
			\frac{(\Vert u \Vert_{2})^{2}}{n_t},
		\]
		\[
			\lim_{\tilde{n}/\log(p) \rightarrow \infty}\frac{1}{n_s}\Vert (e_s)^{T} X_{s}\Vert_{\infty}
			=	0
			\mbox{ if } u \sim Gaussian(0,var(u)),
		\]
		and
		\[
			\lim_{\tilde{n}/p \rightarrow \infty}\varsigma = 0,
		\]
		as a result, $\Vert b_{pen} - \beta\Vert_2 \rightarrow 0$.
		
		\item[\textbf{$K$-fold cross validation.}] In Theorem~\ref{thm3.1},
			\begin{eqnarray}
				\frac{1}{K}\sum_{q=1}^K (\left\Vert b_{OLS}^q-b_{pen}^q\right\Vert _{2})^2
				& \leqslant &
				\frac{1}{K}\sum_{q=1}^K
                \left|\frac{1}{\bar{\rho} n_t}\frac{\left\Vert e_{t}^q \right\Vert_{2}^{2}}
				{\left(1-\sqrt{\epsilon}\right)}
				-\frac{1}{K}\sum_{q=1}^K \frac{1}{\bar{\rho} n_s}
                \left\Vert e_{s}^q \right\Vert _{2}^{2}\right| \notag \\
				& &
				+ \frac{1}{K}\sum_{q=1}^K \frac{4}{\bar{\rho} n_s}
                \left\Vert (e_{s}^q)^{T}X_{s} \right\Vert_{\infty}
				\left\Vert b_{train}\right\Vert _{1} + \frac{\varsigma}{\bar{\rho}}				
			\end{eqnarray}	
		Since
		\[
			\lim_{\tilde{n}/p \rightarrow \infty}
			\frac{1}{n_t} \frac{(\Vert e_t^q \Vert_{2})^{2}}{1-\sqrt{\epsilon}}
			=
			\lim_{\tilde{n}/p \rightarrow \infty}
			\frac{(\Vert e_s^q \Vert_{2})^{2}}{n_s}
			=
			\frac{(\Vert u \Vert_{2})^{2}}{n_t},
		\]
		\[
			\lim_{\tilde{n}/\log(p) \rightarrow \infty}
			\frac{1}{n_s} \Vert (e_s^q)^{T} X_{s}^q \Vert_{\infty}
			=	0
			~\mbox{if u} \sim Gaussian(0,var(u)),
		\]
		and
		\[
			\lim_{\tilde{n}/p \rightarrow \infty}\varsigma = 0,
		\]
        as a result, $(1/K)\sum_{q=1}^K \left(\left\Vert b_{OLS}^q-b_{pen}^q\right\Vert _{2}\right)^2 \rightarrow 0$.
\end{description}
\end{proof}

\bigskip{}
\begin{proof}
\textbf{Proposition~\ref{prop3.4}}
As shown in the discussion above Proposition~\ref{prop3.4}, while Proposition~\ref{prop3.3} is valid for the $p>n$ case, we cannot derive the $\mathcal{L}_2$ difference between $b_{FSR}$ and $b_{pen}$ because $X\Delta$ is no longer strongly convex. As a result, to derive the upper bound of
$\Vert b_{FSR}-b_{pen}\Vert_2$, we need to use the restricted eigenvalue condition \citep{bickeletal09,meinshausenyu09,zhang09}.

	\begin{description}
    \item[Restricted eigenvalue condition.] For some integer $1\leqslant s \leqslant p$ and a positive number $k_0$, for both FSR and Lasso satisfies the following condition
		\[
			\min_{J_0 \subset \{1,\ldots,p\},\vert J_0 \vert \leqslant s}
			~\min_{\Vert \Delta_{J_0^c} \Vert_1 \leqslant k_0\Vert \Delta_{J_0} \Vert_1}
			\frac{\Vert X\Delta\Vert_2}{\sqrt{n}\Vert \Delta_{J_0}\Vert_2} = \rho_{re} >0
		\]
    where $\Delta_{J_0}$ stands for the difference between two vectors with at most $J_0$ non-zero vectors, and $J_0^c$ is the complement set of $J_0$. Also $J_0$ can be treated as the support of $\Vert \Delta\Vert_0$.
	\end{description}
	
	As a result,
	\begin{description}
	   \item[Validation.] For FSR, $(1/n)\left\Vert X_{s}b_{FSR} - X_{s}b_{pen}\right\Vert_{2}^{2}=
        (1/n)\left\Vert X_{s}\Delta\right\Vert _{2}^{2}\geqslant \rho_{re} 
        \left\Vert \Delta\right\Vert _{2}^{2}$,	where $\rho$ is the minimal eigenvalue for $(X_s)^{T}X_s$. Hence, the restricted eigenvalue condition implies
	   \begin{eqnarray}
		  \rho_{re} (\Vert \Delta \Vert_{2})^{2}
		  & \leqslant &
		  \frac{1}{n_s} (\Vert X_{s} \Delta \Vert _{2})^{2}	\notag \\
		  & \leqslant & \left\vert \frac{1}{n_t} \frac{ (\Vert e_t \Vert_{2})^{2}}{1-\sqrt{\epsilon}}
		  - \frac{ \Vert e_s \Vert_{2}^{2}}{n_s} \right\vert \notag \\
		  & & +  \frac{4}{n_s} \Vert (e_s)^{T} X_{s} \Vert_{\infty}\Vert b_{FSR} \Vert_{1} + \varsigma.
	\end{eqnarray}
	By the Minkowski inequality, the above can be simplified to
	\begin{eqnarray}
		\left\Vert b_{train}-b_{Lasso}\right\Vert _{2}
		& \leqslant &		
		\sqrt{\left\vert \frac{1}{n_t \rho_{re}}
		\frac{ (\Vert e_t \Vert_{2})^{2}}{1-\sqrt{\epsilon}}
		- \frac{ \Vert e_s \Vert_{2}^{2}}{n_s \rho_{re}} \right\vert} \notag \\
		& & + \sqrt{\frac{4}{n_s \rho_{re}} \Vert (e_s)^{T} X_{s} \Vert_{\infty}\Vert b_{FSR} \Vert_{1}}
		+ \sqrt{\frac{\varsigma}{\rho_{re}}}.
	\end{eqnarray}	
	
    \item[$K$-fold cross validation.] For the FSR estimate in $q$th round, the restricted eigenvalue value condition implies that $(1/n)\left\Vert X_{s}^q \Delta\right\Vert_{2}^{2}	\geqslant\rho_{re}^q\left\Vert \Delta\right\Vert _{2}^{2}$,	where $\rho_{re}^q$ is the minimal restricted eigenvalue for $(X_s^q)^{T}X_s^q$ in the $q$th round. Hence, if we define the minimum of all the minimal round-by-round eigenvalues from all $K$ rounds,
	   \[
		  \bar{\rho}_{re} = \min\{\rho_{re}^q \vert \forall q \in [1,K] \},
	   \]
	   then
	   \begin{eqnarray}
		  \frac{1}{K}\sum_{q=1}^K \bar{\rho}_{re} (\Vert \Delta^q \Vert_{2})^{2}
		  & \leqslant &
		  \frac{1}{K}\sum_{q=1}^K \frac{1}{n_s} (\Vert X_{s}^q \Delta^q \Vert _{2})^{2} \notag \\
		  & \leqslant & \left\vert \frac{1}{K}\sum_{q=1}^K \frac{\frac{1}{n_t} 
            (\Vert e_t^q \Vert_{2})^{2}}{1-\sqrt{\epsilon}}
		  - \frac{1}{K}\sum_{q=1}^K \frac{ (\Vert e_s^q \Vert_{2})^{2}}{n_s} \right\vert \notag \\
		  & & +\frac{1}{K}\sum_{q=1}^K \frac{4}{n_s} \Vert (e_s^q)^{T} X_{s}^q \Vert_{\infty}
		  \Vert b_{FSR}^q \Vert_{1}	+ \varsigma_{cv}.
	   \end{eqnarray}
	   Hence,
	   \begin{eqnarray}
		\frac{1}{K}\sum_{q=1}^K \left(\left\Vert b_{OLS}^q-b_{pen}^q\right\Vert _{2}\right)^2
		& \leqslant &
		\frac{1}{K}\sum_{q=1}^K \left|\frac{1}{\bar{\rho}_{re} n_t}
		\frac{\left\Vert e_{t}^q \right\Vert_{2}^{2}}
		{\left(1-\sqrt{\epsilon}\right)}
		-\frac{1}{K}\sum_{q=1}^K \frac{1}{\bar{\rho}_{re} n_s}
		\left\Vert e_{s}^q \right\Vert _{2}^{2}\right| \notag \\
		& &
		+\frac{1}{K}\sum_{q=1}^K \frac{4}{\bar{\rho}_{re} n_s}
        \left\Vert(e_{s}^q)^{T}X_{s} \right\Vert_{\infty}
		\left\Vert b_{FSR}^q\right\Vert _{1} +\frac{\varsigma}{\bar{\rho}_{re}}
	   \end{eqnarray}	
	\end{description}
\end{proof}


\end{document}